\documentclass{article}

\usepackage[accepted]{icml2026}

\usepackage[textsize=tiny]{todonotes}

\usepackage{amsmath,amsfonts,bm}

\def\eqref#1{equation~\ref{#1}}

\def\1{\bm{1}}

\DeclareMathAlphabet{\mathsfit}{\encodingdefault}{\sfdefault}{m}{sl}
\SetMathAlphabet{\mathsfit}{bold}{\encodingdefault}{\sfdefault}{bx}{n}

\DeclareMathOperator*{\argmax}{arg\,max}

\usepackage{mathtools}
\usepackage{amsmath}
\usepackage{amssymb}
\usepackage{amsthm}
\usepackage{amsfonts}
\usepackage{enumitem}
\usepackage{siunitx}
\usepackage{nicefrac}
\usepackage{booktabs}
\usepackage{microtype}
\usepackage{xspace}
\usepackage{array}
\usepackage{etoolbox}
\usepackage{float}
\usepackage{tabularx}
\usepackage{multirow}
\usepackage{threeparttable}
\usepackage{tcolorbox}
\usepackage{fancyvrb}
\usepackage{listings}
\usepackage{subcaption}
\usepackage{balance}
\usepackage{hyperref}
\usepackage[capitalise,noabbrev,nameinlink]{cleveref}
\usepackage{url}

\newcommand{\method}{Dream-MPC\xspace}

\Crefname{algorithm}{Alg.}{Algs.}
\Crefname{equation}{Eq.}{Eqs.}
\Crefname{figure}{Fig.}{Figs.}
\Crefname{tabular}{Tab.}{Tabs.}
\Crefname{table}{Tab.}{Tabs.}

\newcommand{\PreserveBackslash}[1]{\let\temp=\\#1\let\\=\temp}
\newcolumntype{C}[1]{>{\PreserveBackslash\centering}p{#1}}
\newcolumntype{R}[1]{>{\PreserveBackslash\raggedleft}p{#1}}
\newcolumntype{L}[1]{>{\PreserveBackslash\raggedright}p{#1}}
\newcolumntype{N}[1]{>{\centering\arraybackslash}m{#1}}

\definecolor{mygreen}{HTML}{7CB518}
\definecolor{myblue}{HTML}{3A86FF}
\definecolor{mypurple}{HTML}{8338EC}
\definecolor{mypink}{HTML}{FF006E}
\definecolor{myorange}{HTML}{FB5607}
\definecolor{myyellow}{HTML}{FFBE0B}

\definecolor{codegreen}{rgb}{0,0.5,0}
\definecolor{codered}{rgb}{0.7,0.1,0.1}
\definecolor{codegray}{rgb}{0.5,0.5,0.5}
\definecolor{codepurple}{rgb}{0.58,0,0.82}
\definecolor{backcolour}{rgb}{1,1,1}
\lstdefinestyle{python}{
    language=Python,
    backgroundcolor=\color{backcolour},   
    commentstyle=\color{codered}\textit,
    keywordstyle=\bfseries\color{codegreen},
    numberstyle=\tiny\color{codegray},
    stringstyle=\color{codepurple},
    basicstyle=\ttfamily\scriptsize,
    breakatwhitespace=false,         
    breaklines=true,                 
    captionpos=b,                    
    keepspaces=true,                 
    numbers=left,                    
    numbersep=4pt,                  
    showspaces=false,                
    showstringspaces=false,
    showtabs=false,                  
    tabsize=1,
    fancyvrb=true
}
\newenvironment{codesnippet}
  { \VerbatimEnvironment%
    \begin{Verbatim} }
  { \end{Verbatim}  }

\newtcolorbox{conclusionbox}{
  colback=blue!5,        %
  colframe=blue!75!black, %
  coltitle=black,        %
  fonttitle=\bfseries,   %
  boxrule=1pt,           %
  arc=1mm,               %
  left=2mm,              %
  right=2mm,             %
  top=1mm,               %
  bottom=1mm,            %
}

\theoremstyle{plain}

\theoremstyle{definition}

\theoremstyle{remark}

\icmltitlerunning{\method{}: Gradient-Based Model Predictive Control with Latent Imagination}

\begin{document}

\twocolumn[
  \icmltitle{\method{}: Gradient-Based Model Predictive Control with Latent Imagination}

  \icmlsetsymbol{equal}{*}

  \begin{icmlauthorlist}
    \icmlauthor{Jonathan Spieler}{ais}
    \icmlauthor{Sven Behnke}{ais}
  \end{icmlauthorlist}

  \icmlaffiliation{ais}{Autonomous Intelligent Systems, Computer Science Institute VI - Intelligent Systems and Robotics, Center for Robotics and the Lamarr Institute for Machine Learning and Artificial Intelligence, University of Bonn, Germany}

  \icmlcorrespondingauthor{Jonathan Spieler}{spieler@ais.uni-bonn.de}

  \icmlkeywords{Machine Learning, ICML, Reinforcement Learning, World Models, Model Predictive Control}

  \vskip 0.3in
]

\printAffiliationsAndNotice{}  

\begin{abstract}
  State-of-the-art model-based Reinforcement Learning (RL) approaches either use gradient-free, population-based methods for planning, learned policy networks, or a combination of policy networks and planning. Hybrid approaches that combine Model Predictive Control (MPC) with a learned model and a policy prior to leverage the advantages of both paradigms have shown promising results. However, these approaches typically rely on gradient-free optimization methods, which can be computationally expensive for high-dimensional control tasks. While gradient-based methods are a promising alternative, recent works have empirically shown that gradient-based methods often perform worse than their gradient-free counterparts. We propose \method, a novel approach that generates few candidate trajectories from a rolled-out policy and optimizes each trajectory by gradient ascent using a learned world model, uncertainty regularization and amortization of optimization iterations over time by reusing previously optimized actions. Our results on 24 continuous control tasks show that \method can significantly improve the performance of the underlying policy and can outperform gradient-free MPC and state-of-the-art baselines. Code and videos are available at \url{https://dream-mpc.github.io}.
\end{abstract}

\begin{figure}[htpb]
  \includegraphics[width=\linewidth, keepaspectratio]{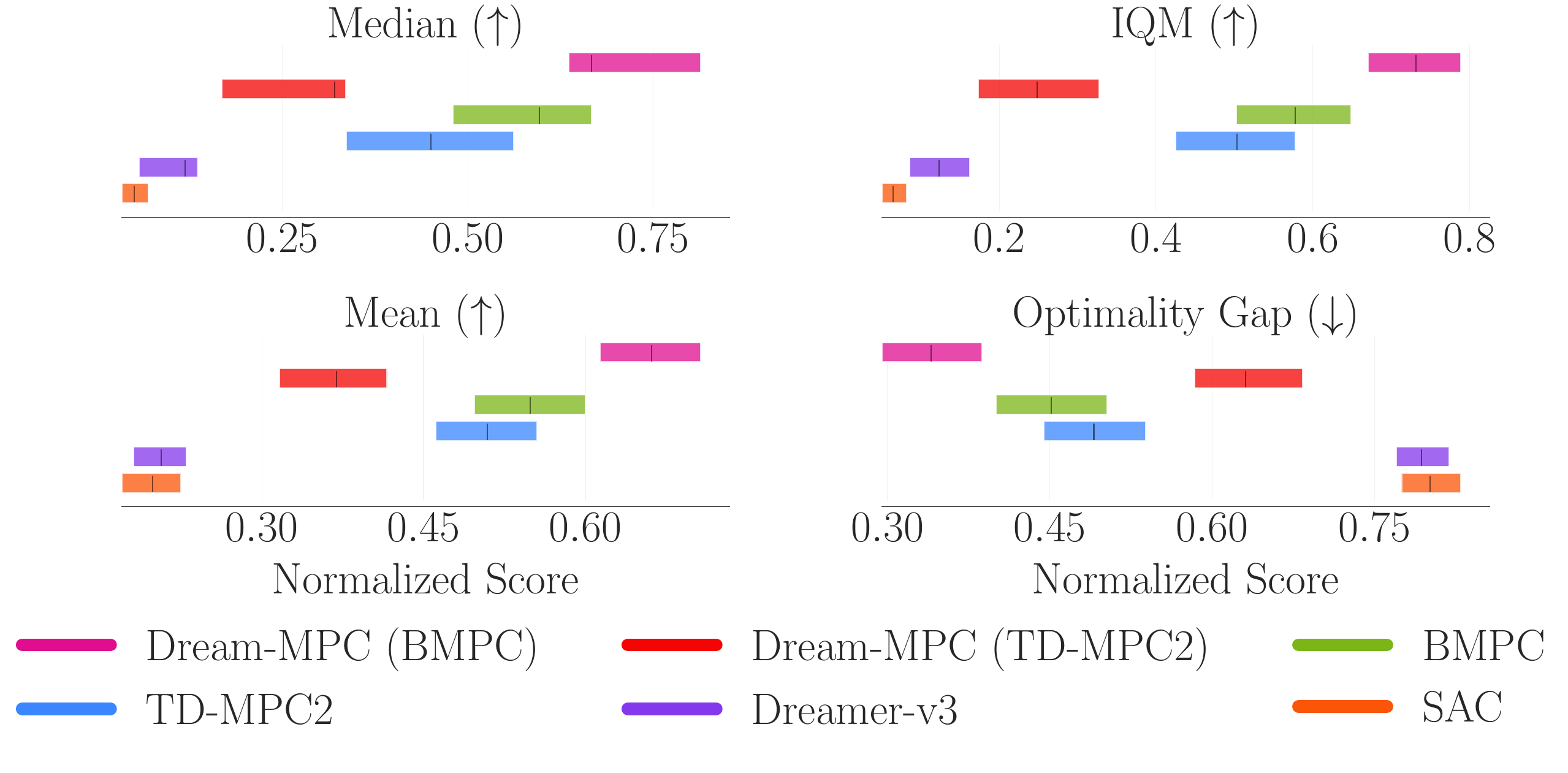}
  \caption{\textbf{Aggregate performance metrics.} Optimality gap, interquartile median (IQM), mean and median normalized scores with 95\% confidence intervals of different methods. Notably, \method with a strong policy achieves the best results.}
  \label{fig:teaser}
  \vspace*{-2ex}
\end{figure}

\section{Introduction}
\looseness=-1
Reinforcement Learning has achieved promising results in recent years and demonstrated its potential for robotics \citep{wu_2023,lancaster_2023,seo2025learningsimtorealhumanoidlocomotion}. However, model-free methods often struggle with sample efficiency and generalization, especially in complex and high-dimensional environments \citep{byravan_evaluating_2021}. Model-based RL, on the other hand, can be more sample-efficient and can generalize better, but requires an accurate model of the environment \citep{xiao_learning_nodate}. There has been growing interest in world models that are learned from data and can be used for decision-making \citep{sutton_dyna_1991, ha_world_2018}. Many recent works \citep{hafner_learning_2019, hansen_temporal_2022, hansen_td-mpc2_2024, srinivas_universal} use a learned world model for planning through imaginary rollouts with MPC \citep{richalet_model_1978,cutler_dynamic_1979} and rely on gradient-free, sampling-based methods such as the Cross Entropy Method (CEM) \citep{rubinstein_optimization_1997} or Model Predictive Path Integral (MPPI) \citep{williams_mppi_2015, williams_information_2017} for trajectory optimization. Although sampling-based MPC methods can be parallelized using Graphics Processing Units (GPUs), their implementation on embedded systems can be challenging due to the limited computational resources. In addition, planning with sampling-based methods is highly inefficient or even intractable in high-dimensional spaces, which might limit their applicability to real-world robotics tasks \citep{xie_latent_2021}. 

In contrast, fully amortized methods such as Dreamer \citep{hafner_dream_2020} learn a purely reactive policy via imaginary rollouts. Inference for the learned policy is computationally less expensive than the search procedure using CEM, but amortized policies often lack generalization \citep{byravan_evaluating_2021}. Since the learned world models are differentiable, it is natural to propose gradient-based methods for trajectory optimization because they can be more efficient than gradient-free, sampling-based methods. Instead of sampling many action sequences and evaluating them as done by CEM, gradients backpropagated through the model can be used to guide the optimization procedure \citep{bharadhwaj_model-predictive}. When the action dimension increases, there is an exponential growth in search space for CEM, while there is only a small increase in computational load for gradient descent, i.e., an additional gradient dimension \citep{bharadhwaj_model-predictive}. While few works propose to combine gradient-based optimization with world models, the empirical results observed are worse than for their gradient-free counterparts \citep{bharadhwaj_model-predictive,s_v_gradient-based_2023,zhou_dino-wm}. 

We propose \method, a novel method which combines gradient-based MPC with a learned policy network and world model that can significantly improve the performance of the policy and even outperform its gradient-free equivalent and state-of-the-art methods. Our method incorporates uncertainty directly into the optimization objective and amortizes optimization iterations over time to further improve performance and computational efficiency compared to previous approaches. We evaluate our method empirically on various tasks from different domains, including high-dimensional tasks and tasks with visual observations, as well as for different model-based RL algorithms with distinct types of world models and when using gradient-based MPC during training. 

In summary, our contributions are as follows:
\begin{itemize}
  \item We propose \method, a gradient-based MPC method, which uses a learned world model for predicting future states and costs and achieves state-of-the-art performance on continuous control tasks.
  \item \method can outperform gradient-free MPC methods, while showing improvements in computational efficiency compared to gradient-free, sampling-based MPC methods.
  \item We overcome limitations of prior gradient-based MPC methods by introducing uncertainty regularization and action reuse and evaluate our method extensively on different environments and settings.
\end{itemize}

\begin{figure*}[htpb]
  \centering
  \includegraphics[width=0.95\linewidth, keepaspectratio]{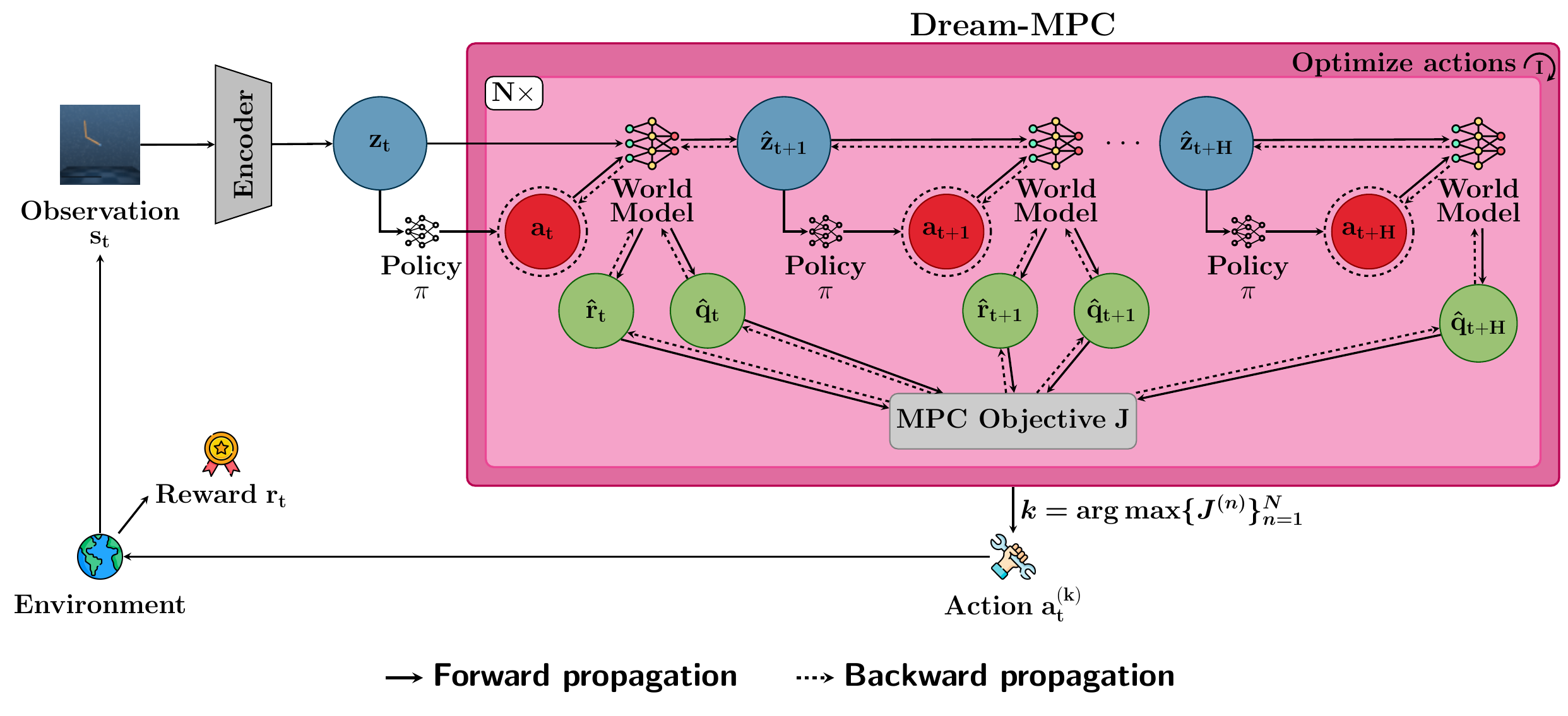}
  \caption{\textbf{Overview of the proposed approach.} \method optimizes action sequences rolled out from a policy network $\pi$ in latent space $z$ with gradient-based MPC. $N$ candidate trajectories are sampled from the policy prior and optimized for $I$ iterations using gradient ascent to maximize the objective $J$. The first action from the candidate trajectory with the highest value of $J$ is applied to the environment, and the procedure is repeated for the next time step. The policy network and world model are shared across candidates and time steps.}
  \label{fig:overview}
  \vspace*{-2ex}
\end{figure*}

\section{Related Work}
\textbf{Model-based RL.} Model-based RL tries to learn a model of the environment that can be used to predict the outcome of actions and plan accordingly \citep{sutton_dyna_1991}. World models are considered a central component of human thinking and decision-making processes \citep{sutton_dyna_1991,ha_world_2018,lecun_path_2022}. While some approaches to world modelling show promising results and are able to generalize to different domains, they are mostly focused on representation learning and not or only partially cover the planning aspect. The combination of elements of planning and search (especially Monte Carlo Tree Search) with deep reinforcement learning has shown remarkable successes in game domains \citep{silver_mastering_2016,silver_mastering_2017}. Most recent model-based RL approaches use the learned world model for planning through imaginary rollouts \citep{srinivas_universal,micheli_2023,hansen_td-mpc2_2024,hafner_mastering_2025,mosbach_2025}. However, the performance of these approaches depends heavily on the quality of the learned world model \citep{talvitie_model} and often suffers from the compounding error problem \citep{asadi_combating_2019}.

\textbf{MPC and RL.} State-of-the-art approaches, such as those from the Dreamer family \citep{hafner_dream_2020,hafner_mastering_2022,hafner_mastering_2025}, use a policy network to predict the actions directly. While policy networks have shown remarkable success for robotics applications, the world model and value function are typically only utilized during training, and the policy is then frozen during inference. This procedure leads to a reactive policy, which can be considered as offline planning and limits the generalization capabilities \citep{byravan_evaluating_2021}. To address this limitation, recent works such as TD-MPC \citep{hansen_temporal_2022,hansen_td-mpc2_2024}, POLO \citep{lowrey_plan_2019} or PlaNet \citep{hafner_learning_2019} combine model-based RL with online planning through MPC to leverage the benefits of both paradigms. Typically, MPC is performed using gradient-free, sampling-based methods such as CEM or MPPI. Although, the results obtained empirically are often good, for each time step, hundreds or thousands of different action alternatives are sampled and evaluated, which increases the computational effort and renders these approaches only partly suitable for real-time applications. 

\textbf{Gradient-based Planning.} The idea of gradient-based planning has been around for decades \citep{kelley_gradient_1960} and typically refers to backpropagating gradients of a cost function with respect to actions to iteratively optimize a sequence of actions by gradient descent. While early works relied on known analytic forms of environment dynamics, more recent works revisited the idea with learned approximate models of the environment \citep{srinivas_universal,silver_predictron_2017,henaff_model-based_2018}. However, there are only few works that have been able to successfully perform gradient-based planning and these approaches are usually limited since they either require expert demonstrations \citep{srinivas_universal} or cannot scale to more challenging robotics tasks \citep{henaff_model-based_2018}. Works such as \citep{bharadhwaj_model-predictive} and \citep{s_v_gradient-based_2023} use a Gaussian as a proposal distribution for gradient-based optimization. Typically, a more informative proposal is used for MPC to warm-start the optimization procedure, for example a policy network. Prior works which combine policy models and MPC mostly use the policy model to generate a trajectory, which is then optimized using gradient-free methods \citep{byravan_evaluating_2021,mansard_using_2018,hamrick_role_2021,argenson_model-based_2021,morgan_model_2021}. Since the learned world models are usually differentiable, also gradient-based methods have been proposed for optimizing the trajectory proposal from a policy model \citep{s_v_gradient-based_2023}. However, gradient-based optimization methods perform worse in their experiments compared to their gradient-free counterparts. The reasons are attributed to problems with the gradients, but are not analyzed in detail.

Note that while the general idea of combining policy networks with MPC itself is not new, previously proposed methods have only been applied to few and relatively simple tasks without systematically evaluating their performance. To the best of our knowledge, we are the first to achieve a gradient-based MPC method with a learned world model that can outperform its gradient-free equivalent and state-of-the-art baselines by introducing uncertainty regularization and reusing previously planned actions. We also evaluate the performance of gradient-based MPC for a broad variety of environments, including state- and image-based observations and different types of world models. In order to place our work in context with previous work, we provide a summary over the main differences between \method and Policy+Grad-MPC \citep{s_v_gradient-based_2023} in \cref{sec:differences-grad-mpc}.

\section{Preliminaries}

\textbf{Reinforcement Learning} can be formulated as an infinite-horizon Markov Decision Process (MDP) with continuous action and state spaces, which can be defined as a tuple $\langle \mathcal{S}, \mathcal{A}, \mathcal{T}, \mathcal{R}, \gamma \rangle$, where $\mathcal{S}$ and $\mathcal{A}$ are the state and action spaces, $\mathcal{T}: \mathcal{S} \times \mathcal{A} \rightarrow \mathcal{S}$ is the transition or dynamics function, $\mathcal{R}: \mathcal{S} \times \mathcal{A} \rightarrow \mathbb{R}$ is the reward function and $\gamma$ is a discount factor. The goal is to obtain a policy $\pi: \mathcal{S} \rightarrow \mathcal{A}$, which maximizes the expected discounted sum of rewards, i.e., the return $\mathbb{E}_{\pi} [\sum_{t=0}^{\infty} \gamma^t r_t]$, where $r_t = \mathcal{R}(s_t, \pi(s_t))$. Model-based RL learns a model of the environment, often referred to as world model, which is then used for selecting actions and deriving a policy by planning with the learned model.

\textbf{Model Predictive Control} is a well-known method for trajectory optimization, which minimizes a cost function over a finite horizon while taking the system dynamics and constraints into account. The optimization problem is solved at each time step, using the current state as initial condition and the predicted future states. The solution provides the optimal action sequence for the next few time steps with respect to the predicted costs. Thus, MPC generates a locally optimal sequence of actions up to the prediction horizon $H$ by solving the following optimization problem:
\begin{equation}\label{eq:mpc}
    \pi(s_t) = \argmax_{a_{t:t+H}} \mathbb{E} \left[\sum_{i = 0}^{H} \gamma^{t+i} \mathcal{R}(s_{t+i}, a_{t+i})\right].
\end{equation}
The learned model is used to estimate the return of a candidate trajectory \citep{negenborn_2005}. Since solving \cref{eq:mpc} leads to a locally optimal solution and is not guaranteed to solve the general RL problem outlined before, most state-of-the-art methods learn value functions to bootstrap return estimates beyond the horizon $H$.

\section{\method: Gradient-Based Model Predictive Control}\label{sec:dream-mpc}
We propose \method, which uses gradient ascent to optimize action sequences sampled from a policy network in an MPC-like manner. The idea is shown in \cref{fig:overview}. Since gradient ascent is prone to getting stuck at local optima, we propose to generate few candidate trajectories by sampling from a stochastic policy network. Instead of sampling thousands of trajectories from a Gaussian distribution like CEM, we only consider few trajectories based on the policy. Namely, for each time step $t$, the algorithm creates $N$ initial action sequences by performing an imaginary rollout of a stochastic policy $\pi_{\theta}$ in latent space $z$ using a learned latent dynamics model $d$: 
\begin{align}\label{eq:trajectories}
  \begin{aligned}
    &\hat{a}_{\tau}^{(n)} \sim \pi_{\theta}(\cdot | z_{\tau}^{(n)}), \quad z_{\tau+1}^{(n)} = d(z_{\tau}^{(n)}, \hat{a}_{\tau}^{(n)}),\\
    &\quad \text{with} \quad \tau=t, ..., t+H, \quad n=1,..., N.
  \end{aligned}
\end{align}
In case of a deterministic policy we add small perturbations to the initial action sequence sampled from the policy to generate $N$ candidate trajectories. The learned world model predicts the following latent states as well as the rewards $\hat{r}$ for each state and the terminal values $\hat{q}$. Each trajectory is then refined using gradient ascent with step size $\alpha$ to maximize the respective expected return, which is estimated using the predictions from the world model. The first action of the candidate trajectory with the highest expected return is applied, and the planning procedure is repeated in the next time step. Sampling from a policy provides a warm-start through proposing a decent initial solution for the optimization, which has been shown to be essential for the performance of gradient-free \citep{hansen_temporal_2022} and gradient-based optimization methods \citep{parmas_pipps_2018}. Our method allows for combining the benefits of both, fully amortized methods using reactive policies and fully online planning, namely improved generalization while reducing computational costs. In contrast to naively sampling random action sequences, which do not leverage any knowledge of the optimization problem, our approach uses gradients backpropagated through the learned world model to efficiently guide the optimization.

Since we optimize actions over a receding horizon, but only apply the first action at each time step, we propose to amortize optimization iterations over time by reusing corresponding optimized actions from previous time steps to initialize actions as a mixture of previously optimized action $\tilde{a}$ and policy actions $\hat{a}$:
\begin{equation}\label{eq:action-reuse}
  a_{\tau}^{(n)} = \rho \cdot \tilde{a}_{\tau-1}^{(n)} + (1 - \rho) \cdot \hat{a}_{\tau}^{(n)}, \quad n=1, ..., N,
\end{equation}
where $\rho$ is the reuse coefficient, which controls the influence of the previously optimized actions. For the action at time step $t+H$, there is no previously planned action. Thus, we initialize the planned action by the same value as the planned action of the previous time step.

For our experiments, we integrate our method into TD-MPC2 \citep{hansen_td-mpc2_2024}, a model-based RL algorithm, which performs local trajectory optimization using MPPI in the latent space of a learned world model. Instead of learning a dynamics model using a reconstruction objective, TD-MPC2 implicitly learns a control-centric world model from environment interactions using a combination of joint-embedding prediction, reward prediction, and TD-learning without decoding observations. The TD-MPC2 architecture consists of following five learned components:
\begin{equation*}
    \label{eq:tdmpc2-components}
    \begin{array}{lll}
        \text{Encoder:} && z_t = h(s_t),\\
        \text{Latent dynamics:} && z_{t+1} = d(z_t, a_t),\\
        \text{Reward:} && \hat{r}_t = R(z_t, a_t),\\
        \text{Terminal value:} && \hat{q}_t = Q(z_t, a_t),\\
        \text{Policy prior:} && \hat{a}_t \sim \pi_{\theta}(z_t),
    \end{array}
\end{equation*}
where $s$ and $a$ are the states and actions, and $z$ is the latent representation. Since we only consider single-task experiments in this work, we omit the learnable task embedding used for multi-task world models.

The policy prior $\pi_{\theta}$ serves to guide the sampling-based MPPI trajectory optimizer in TD-MPC2 as well as our gradient-based method. TD-MPC2 maintains a replay buffer $\mathcal{B}$ during online interaction, which is used to iteratively update the world model and collect new environment data by planning with the learned model. Please refer to \cref{sec:implementation-details} for details on the model training, architecture and MPPI planning procedure. We replace the MPPI planner by our gradient-based MPC method. Our gradient-based MPC algorithm is summarized in \cref{algo:dream-mpc}. The MPC procedure requires $N \times I \times H$ evaluations of the world model at each time step, which equals $512 \times 6 \times 3 = \num{9216}$ for MPPI while our method uses significantly less model evaluations, i.e., only $5 \times 1 \times 3 = 15$. Note that while we use TD-MPC2 for our experiments, our method can also be integrated into other model-based reinforcement learning approaches such as Dreamer \citep{hafner_dream_2020} or DINO-WM \citep{zhou_dino-wm}. We provide additional results for \method with Dreamer in \cref{sec:dreamer-results}. 

{\scriptsize
\begin{algorithm}[!htp]
    \small
    \caption{{\bf \small \method}
    \label{algo:dream-mpc}}
    \begin{algorithmic}
        \STATE {\bfseries Input:} Encoder $h(s)$, dynamics model $d(z, a)$, reward model $R(z, a)$, value function model $Q(z, a)$, policy prior $\pi_{\theta}(z)$, current state $s_t$, planning horizon $H$, optimization iterations $I$, candidates per iteration $N$, action optimization rate $\alpha$
        \newline
        \STATE Encode state into latent representation $z_t \leftarrow h(s_t)$. 
        \STATE Sample $N$ action sequences by rolling out the policy $\pi_{\theta}$ with the latent dynamics model $d$.
        \STATE Initialize candidate action sequences $a_{t:t+H}$ via \cref{eq:action-reuse}. 
        \FOR{optimization iteration $i=1, 2, \dots I$}
            \FOR{candidate action sequence $n=1, 2, \dots N$}
                \FOR{rollout step $\tau=t \dots t+H-1$}
                    \STATE Predict reward $\hat{r}_{\tau}^{(n)} = R(z_{\tau}, a_{\tau})$.
                    \STATE Predict uncertainty $u_{\tau}^{(n)}$ via \cref{eq:uncertainty}.
                    \STATE Predict next latent state $z_{\tau+1}^{(n)} \leftarrow d(z_{\tau}, a_{\tau})$.
                \ENDFOR
                \STATE Predict terminal value $\hat{q}_{t+H}^{(n)} = Q(z_{t+H}, a_{t+H})$. 
                \STATE Compute MPC objective $J^{(n)}$ using $\hat{r}$, $\hat{q}$ and $u$ via \cref{eq:optimization-obj}.
                \STATE Optimize action sequence via $a^{(n)}_{t:t+H} \leftarrow a_{t:t+H} + \alpha \nabla_{a} J^{(n)}$. 
            \ENDFOR
        \ENDFOR
        \STATE \textbf{return} First optimized action $a_t^{(k)}$ with $k = \argmax_{n} \{J^{(n)}\}_{n=1}^N$.
    \end{algorithmic}
\end{algorithm}
}

We further integrate our method into BMPC \citep{wang_2025}, which builds on TD-MPC2 and learns a policy $\pi_{\theta}$ by imitating an MPC expert $\pi_{\text{MPC}}$ and at the same time uses the policy to guide the MPC optimization process. The results of \citet{wang_2025} show that this bootstrapping approach can improve sample efficiency and asymptotic performance, especially for high-dimensional tasks. We use BMPC since it provides a higher quality policy compared to TD-MPC2, where the performance gap between the policy network and the MPC procedure is quite large as shown in \cref{sec:detailed-results}. For more details on BMPC, please refer to \cref{sec:bmpc}. 

The world model may be queried on unseen state-action pairs during planning, i.e., when estimating returns of candidate trajectories. This can lead to extrapolation errors even if we have learned a good value function \citep{feng_2023}. Recent work \citep{lin2025tdmpc2improvingtemporaldifference} has shown that TD-MPC2 suffers from value overestimation. To address this, we propose to regularize the planning procedure by penalizing trajectories with a large uncertainty and by balancing estimated returns and (epistemic) model uncertainty when evaluating candidate trajectories. Therefore, we estimate the (epistemic) uncertainty of a trajectory as proposed by \citet{hansen_td-mpc2_2024} for offline RL and multi-task world models:
\begin{equation}\label{eq:uncertainty}
    u_t = \text{avg}([\hat{q}_1,\hat{q}_2, \dots, \hat{q}_M]) \cdot \text{std}([\hat{q}_1, \hat{q}_2, \dots, \hat{q}_M]),
\end{equation}
where $\hat{q}_m$ is the predicted value from Q-function $m$ from an ensemble of $M$ Q-functions. The regularization strength at each time step is scaled based on the magnitude of the mean value predictions for a given latent state to account for different tasks without requiring task-specific coefficients. 

The planning objective is then redefined as:
\begin{align}\label{eq:optimization-obj}
  \begin{aligned}
    J = &\sum_{h=t}^{H-1} \left(\gamma^h \cdot R(z_h, a_h) - \lambda_{\text{unc}} \cdot u_h\right)\\
    & + \gamma^H \cdot Q(z_{t+H}, a_{t+H}) - \lambda_{\text{unc}} \cdot u_{t+H},
  \end{aligned}
\end{align}
where $\lambda_{\text{unc}}$ is a task-agnostic coefficient that balances return maximization and uncertainty minimization. While this requires to specify a coefficient that weighs both aspects, we found it sufficient in our experiments to set $\lambda_{\text{unc}} = 0.01$. We use the same hyperparameters as the underlying base methods and also the same set of hyperparameters across all tasks. The parameters specific to our MPC method are listed in \cref{tab:dream-mpc-hparams}. We also conduct experiments in which we use this uncertainty regularization for TD-MPC2 and BMPC and include the results in \cref{sec:detailed-results}. 

\section{Experiments}
We evaluate our method on a set of 24 diverse continuous control tasks from the DeepMind Control Suite \citep{tassa_dm_control_2020}, HumanoidBench \citep{sferrazza_humanoidbench_2024} and Meta-World \citep{metaworld_2019} covering a wide range of task difficulties including high-dimensional state and action spaces, sparse rewards, complex locomotion, and manipulation. Additionally, we also include results for six DeepMind Control tasks with visual observations. For details on the environments, please refer to \cref{sec:environment-details}.

\subsection{Comparison to Baselines}
We compare our method to following state-of-the-art baselines commonly used for continuous control tasks:
\begin{itemize}
  \item Soft-Actor-Critic (SAC) \citep{haarnoja_soft_2018}, a model-free RL method which uses a maximum entropy objective for policy learning,
  \item Dreamer-v3 \citep{hafner_mastering_2025}, a model-based RL method which learns a policy network using rollouts from a generative world model,
  \item TD-MPC2 \citep{hansen_td-mpc2_2024}, a model-based RL method which uses policy-guided MPPI for action selection, and
  \item BMPC \citep{wang_2025}, an extension of TD-MPC2 which uses imitation learning of the MPC planner for policy learning.
\end{itemize}

\begin{figure}[htpb]
    \centering
    \includegraphics[width=\linewidth]{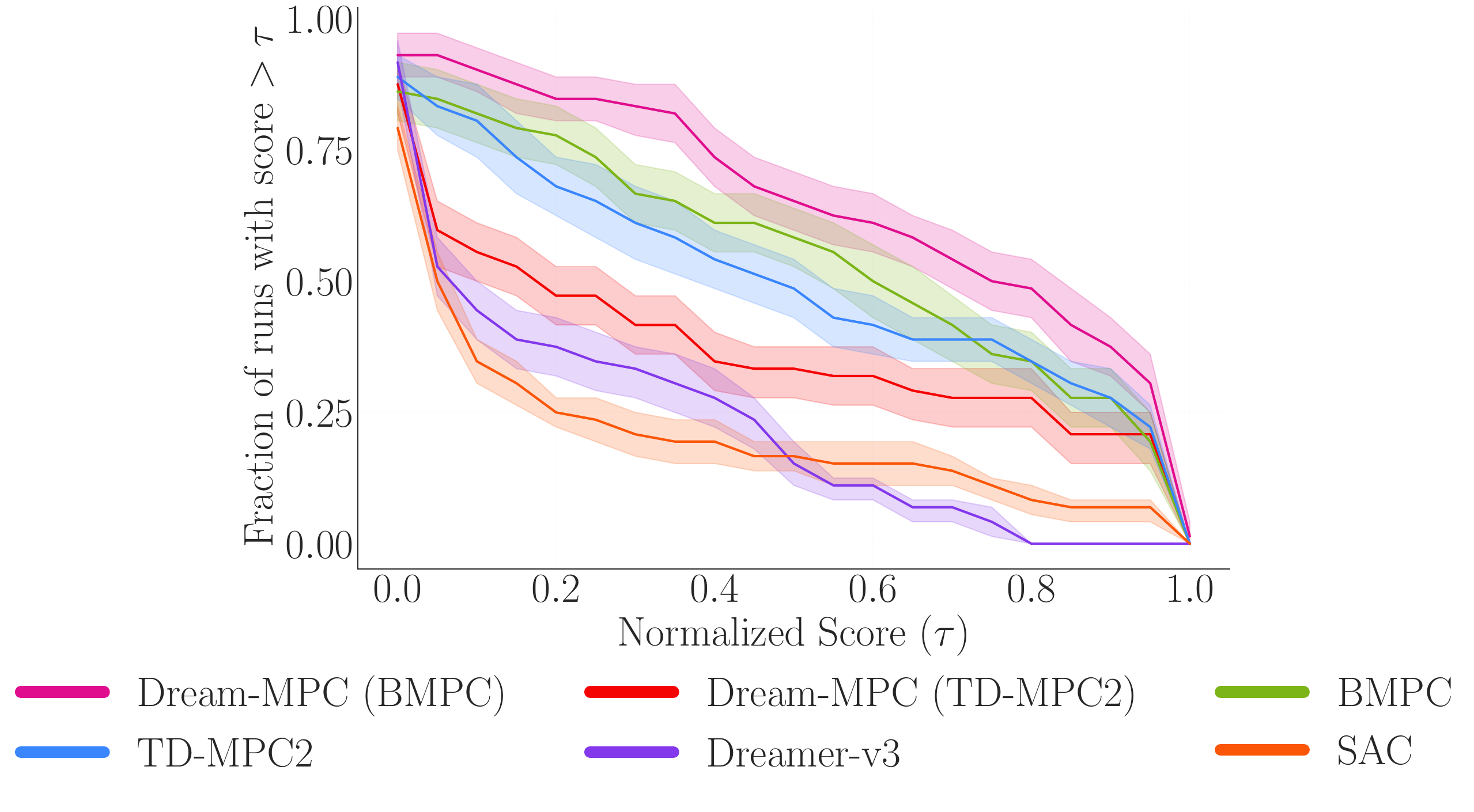}
    \caption{\textbf{Performance profiles.} Score distributions across all tasks, which provides insights into the variance of the performance. Detailed results are included in \cref{tab:eval-results-dmc,tab:eval-results-mw,tab:eval-results-humanoidbench}.}
    \label{fig:metrics}
\end{figure}

We first evaluate the performance of \method when replacing the MPPI planner by our proposed gradient-based MPC planner at test time using (pre-)trained TD-MPC2 and BMPC models, respectively. For more details on the baselines refer to \cref{sec:implementation-details}. We report performance metrics across all 24 tasks using the \textit{rliable} package provided by \citet{agarwal_2021} to evaluate the performance of our method. Specifically, we report the optimality gap, median, interquartile median (IQM), and mean normalized scores as well as the performance profile curves with 95\% confidence intervals based on the evaluation scores of trained BMPC agents in \cref{fig:teaser,fig:metrics}. Confidence intervals are estimated using the percentile bootstrap with stratified sampling as recommended by \citet{agarwal_2021}. For a comparison across different score scales of all tasks, we normalize DeepMind Control Suite scores by dividing by $1000$, and HumanoidBench scores as proposed in \citet{lee_2025}:
\begin{equation}
  \displaystyle %
  \text{\small Normalized-Score}(x) = \frac{x - \text{\small random score}}{\text{\small target score} - \text{\small random score}},
\end{equation}
where we use the random and target success scores provided by the authors. Please refer to \citet{lee_2025} for more details. Meta-World scores are left as they are since the success rates are already values between zero and one. The detailed evaluation results for all environments are shown in \cref{tab:eval-results-dmc,tab:eval-results-humanoidbench,tab:eval-results-mw}. Our gradient-based MPC method can improve the performance of the policy network and outperforms MPPI when using BMPC as a basis. Combined with BMPC, Dream-MPC improves the IQM by 26.7\% and the mean by 20.5\% compared to BMPC, outperforming all baselines.

While \method can also significantly improve the performance of the underlying policy for TD-MPC2 by 144.7\% (IQM) and 43.4\% (mean), it cannot consistently match the performance of MPPI because for TD-MPC2 there is a large gap between the performance of the policy only and with MPPI as shown in \cref{tab:tdmpc2-avg-scores-per-domain}. The detailed results for each environment are shown in \cref{tab:tdmpc2-eval-results-dmc,tab:tdmpc2-eval-results-mw,tab:tdmpc2-eval-results-humanoidbench}. We find that having a good policy is important because it leads to better value estimates, which are crucial for gradient-based MPC. This fact favours MPPI, which has a higher diversity of the initial solutions due to the sampling procedure. While we can further improve the performance of \method with TD-MPC2 as a basis, for example by increasing the number of optimization iterations, this also increases computational costs. It therefore highlights the importance of a good initial solution to warm-start the MPC optimization process, especially for high-dimensional problems.
\begin{scriptsize}
    \begin{table}[htbp]
    \centering
    \parbox{\linewidth}{
    \centering
    \setlength\tabcolsep{1pt}
      \caption{\textbf{Mean scores per domain.} Performance comparison of different TD-MPC2 variants on different benchmarks.}
      \scriptsize %
      \renewcommand{\arraystretch}{1.15}
      \resizebox{\linewidth}{!}{\begin{tabular}{@{}lN{1.75cm}N{1.75cm}N{2.25cm}@{}}
        \toprule
          \textbf{Benchmark}             & \textbf{TD-MPC2}                  & \textbf{TD-MPC2 (policy only)}   & \textbf{\method (TD-MPC2)}   \\
          \hline
            DeepMind Control         & 657 $\pm$ 225           & 367 $\pm$ 247  & 433 $\pm$ 259                               \\
            HumanoidBench         & 761 $\pm$ 1617             & 224 $\pm$ 505  & 379 $\pm$ 897                               \\
            Meta-World         & 0.67 $\pm$ 0.33             & 0.38 $\pm$ 0.35  & 0.62 $\pm$ 0.31                                \\
          \bottomrule
      \end{tabular}}
      \label{tab:tdmpc2-avg-scores-per-domain} %
    }
    \vspace*{-1ex}
    \end{table}
\end{scriptsize}

\subsection{Visual Observations}
Additionally, we evaluate the performance of our method using image-based observations to demonstrate that our method also works well in these settings. The results are shown in \cref{tab:bmpc-eval-results-visual}. We find that our method can also improve the performance of the underlying policy and even outperforms MPPI for visual observations.

\begin{scriptsize}
    \begin{table}[hbpt]
    \centering
    \resizebox{\linewidth}{!}{\begin{threeparttable}
    \setlength\tabcolsep{1pt}
      \caption{\textbf{Visual observations.} Performance comparison of different BMPC variants on tasks from the DeepMind Control Suite using image-based observations.}
      \scriptsize %
      \renewcommand{\arraystretch}{1.15}
      \begin{tabular}{@{}lN{2cm}N{1.5cm}N{1.75cm}@{}}
        \toprule
          \textbf{Environment}             & \textbf{BMPC}                  & \textbf{  BMPC} \newline \textbf{(policy only)}   & \textbf{\method (BMPC)}   \\
          \hline
            Acrobot Swingup         & 287 $\pm$ 45             & \textbf{292 $\pm$ 18}   & \underline{288 $\pm$ 31}                                     \\
            Cartpole Swingup Sparse & \underline{709 $\pm$ 120}            & 625 $\pm$ 283           & \textbf{725 $\pm$ 141}                                       \\
            Cheetah Run             & \underline{609 $\pm$ 23} & 597 $\pm$ 45            & \textbf{643 $\pm$ 9}                                         \\
            Hopper Hop              & 253 $\pm$ 11             & \underline{264 $\pm$ 6} & \textbf{275 $\pm$ 3}                                         \\
            Quadruped Walk          & \underline{427 $\pm$ 78} & 402 $\pm$ 44            & \textbf{435 $\pm$ 76}                                        \\
            Walker Run              & 740 $\pm$ 15             & \underline{740 $\pm$ 6} & \textbf{762 $\pm$ 6}                                         \\
          \bottomrule
      \end{tabular}
      \label{tab:bmpc-eval-results-visual}  
      \footnotesize{The results are the mean episode returns and standard deviations for three random seeds and ten test episodes. \textbf{Best} and \underline{second best} results are highlighted.}
    \end{threeparttable}%
    }
    \vspace*{-1ex}
    \end{table}
\end{scriptsize}

\begin{figure*}[htpb]
    \centering
    \includegraphics[width=0.9\linewidth, keepaspectratio]{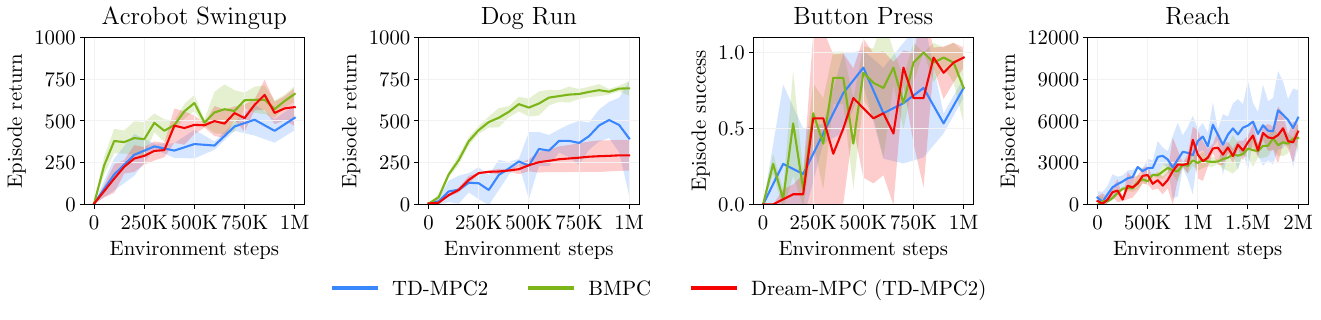}
    \caption{\textbf{Learning curves for four tasks from the DeepMind Control Suite, Meta-World and HumanoidBench.} The line represents the mean episodic return and the shaded area the 95\% confidence interval across 3 seeds.}
    \label{fig:learning-curves-dream-mpc-tdmpc2}
\end{figure*}

\subsection{Training with Gradient-based MPC}
In addition to analyzing our gradient-based MPC method only during inference, we also evaluate its performance when it is already being used during training. Therefore, we use TD-MPC2 as a basis without imitation learning because we hypothesize that the bootstrapping approach of BMPC might lead to unstable training and premature convergence, especially since we have only few candidate trajectories. While combining gradient-based MPC with imitation learning is an interesting research direction, we leave this for future work. \cref{fig:learning-curves-dream-mpc-tdmpc2} shows the learning curves of BMPC, TD-MPC2 and of \method for four different environments. Overall, our gradient-based MPC planner can match the performance of TD-MPC2's MPPI planner. While for simpler control problems \method can even outperform TD-MPC2 and match BMPC, we find that for high-dimensional problems our method performs slightly worse. The lower asymptotic performance of gradient-based MPC results from premature convergence due to less exploration compared to MPPI, which is used by TD-MPC2 and BMPC. We discuss this in more detail in \cref{sec:analysis-dog-run}.

\subsection{Integration into Dreamer}\label{sec:dreamer-results}
We also find improvements in sample-efficiency and asymptotic performance when integrating our method into Dreamer, which is shown in \cref{fig:learning-curves-dreamer}. Our evaluation results shown in \cref{tab:eval-results-dreamer} further demonstrate that \method outperforms all baselines including its gradient-free equivalents and both previously proposed Grad-MPC variants. Implementation details and additional experiments are included in \cref{sec:dreamer}.
\begin{scriptsize}
    \begin{table}[htbp]
    \centering
    \resizebox{\linewidth}{!}{
    \begin{threeparttable}
    \setlength\tabcolsep{1pt}
      \caption{\textbf{Performance comparison of different algorithms.}}
      \scriptsize %
      \renewcommand{\arraystretch}{1.15}
      \begin{tabular}{@{}lN{1.5cm}N{1.25cm}N{1.25cm}N{1.25cm}@{}}
        \toprule
        \textbf{Method} & \textbf{Acrobot Swingup} & \textbf{Cheetah Run} & \textbf{Hopper Hop} & \textbf{Walker Run}  \\
        \midrule
        SAC+AE  & 7 $\pm$ 19 & 495 $\pm$ 100 & 86 $\pm$ 75 & 453 $\pm$ 69 \\ 
        PlaNet  & 7 $\pm$ 18 & 535 $\pm$ 70 & 1 $\pm$ 4 & 228 $\pm$ 149 \\ 
        Dreamer & 134 $\pm$ 91 & \underline{751 $\pm$ 111} & \underline{182 $\pm$ 43} & 575 $\pm$ 33 \\ 
        Grad-MPC & 7 $\pm$ 18 & 438 $\pm$ 81 & 3 $\pm$ 5 & 382 $\pm$ 35  \\
        Policy+Grad-MPC & \underline{144 $\pm$ 7} & 591 $\pm$ 131 & 158 $\pm$ 47 & 556 $\pm$ 33 \\
        CEM+policy & 12 $\pm$ 26 & 674 $\pm$ 20 & 43 $\pm$ 42 & \textbf{638 $\pm$ 21} \\ 
        \method (Dreamer) & \textbf{147 $\pm$ 101} & \textbf{836 $\pm$ 49} & \textbf{298 $\pm$ 86} & \underline{632 $\pm$ 52}  \\ 
        \bottomrule
      \end{tabular}
      \label{tab:eval-results-dreamer}  
      \footnotesize{The results are the mean episode returns and standard deviations for three random seeds and ten test episodes. \textbf{Best} and \underline{second best} results are highlighted.}
      \end{threeparttable}%
      }
      \vspace*{-1ex}
    \end{table}
\end{scriptsize}

\subsection{Ablation Study}
We perform ablations to evaluate our design choices and provide insights into which components are crucial to successfully perform gradient-based MPC. Using a high-quality policy prior to warm-start the MPC optimization is particularly important for high-dimensional problems, as shown in \cref{tab:ablations}. Together with reusing previously optimized actions, warm-starting reduces computational costs. 
\begin{scriptsize}
    \begin{table}[htbp]
    \centering
    \resizebox{\linewidth}{!}{
    \begin{threeparttable}
    \setlength\tabcolsep{1pt}
      \caption{\textbf{\method ablations.} We compare the performance of different variants using trained BMPC models.}
      \scriptsize %
      \renewcommand{\arraystretch}{1.15}
      \begin{tabular}{@{}lN{1cm}N{1.35cm}N{1.5cm}c@{}}
        \toprule
        \textbf{Method} & \textbf{Acrobot Swingup} & \textbf{Humanoid Run} & \textbf{Button Press} & \textbf{Reach} \\
        \midrule
        \method (BMPC) & \textbf{596 $\pm$ 50} & \textbf{531 $\pm$ 38}  & 0.67 $\pm$ 0.47 & \textbf{4348 $\pm$ 215} \\ 
        \quad w/o MPC (policy-only) & 564 $\pm$ 52 & 458 $\pm$ 15 & \textbf{1.00 $\pm$ 0.00} & 2117 $\pm$ 309 \\ 
        \quad w/o policy prior & 554 $\pm$ 21 & 7 $\pm$ 4 & 0.70 $\pm$ 0.22  & 842 $\pm$ 239 \\ 
        \quad w/o gradient ascent & 579 $\pm$ 43 & 496 $\pm$ 25 & \underline{0.97 $\pm$ 0.05} & 2362 $\pm$ 323 \\ 
        \quad w/o uncertainty reg. & \underline{583 $\pm$ 58} & \underline{524 $\pm$ 39} & \textbf{1.00 $\pm$ 0.00}  & \underline{4239 $\pm$ 232} \\ 
        \bottomrule
      \end{tabular}
      \label{tab:ablations}  
      \footnotesize{The results are the mean episode returns and standard deviations for three random seeds and ten test episodes. \textbf{Best} and \underline{second best} results are highlighted.}
      \end{threeparttable}%
    }
    \vspace*{-1ex}
    \end{table}
\end{scriptsize}

\begin{figure*}[!ht]
  \setlength{\belowcaptionskip}{5pt}
  \centering
  \begin{subfigure}[b]{0.7\linewidth}
    \includegraphics[width=\linewidth]{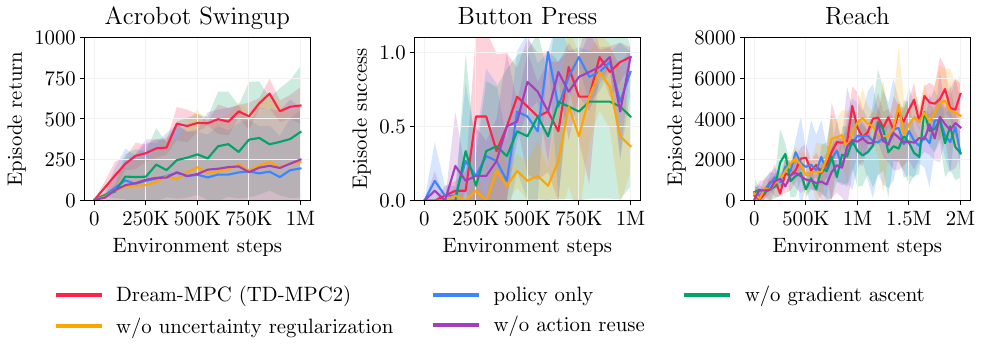}
    \caption{Ablation of design choices}\label{fig:ablations-design-choices}
  \end{subfigure}
  \begin{subfigure}[b]{0.765\linewidth}
    \includegraphics[width=\linewidth]{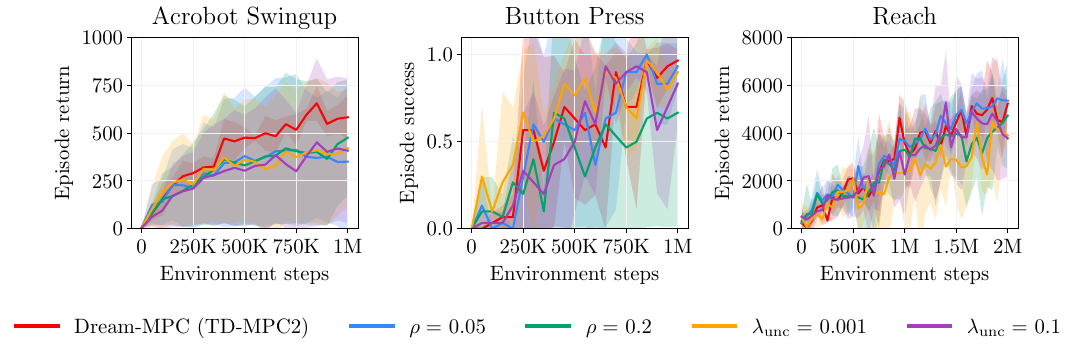}
    \caption{Sensitivity analysis of coefficients}\label{fig:ablations-sensitivity-analysis}
  \end{subfigure}
  \caption{
      \textbf{Ablations.} (a) Performance of different \method (TD-MPC2) variants demonstrating the importance of each design choice. (b)  Performance of \method (TD-MPC2) with different uncertainty regularization and action reuse coefficients. The line represents the mean episodic return and the shaded area the 95\% confidence interval across 3 seeds.
  }
  \label{fig:ablations}
\end{figure*}

\looseness=-1
We replace the policy prior by a Gaussian distribution to highlight the importance of a good initial proposal distribution to warm-start the MPC process and use the same number of candidate trajectories as MPPI, i.e., 512. For a fair comparison, we compensate for the less informative prior by increasing the number of optimization iterations to five, which, depending on the environment, leads to an increase in inference time by a factor of about five to ten compared to \method. We further find that uncertainty regularization and amortization of optimization iterations by reuse of previous planned actions are especially important when using gradient-based MPC during training, as illustrated in \cref{fig:ablations-design-choices}. \cref{fig:ablations-sensitivity-analysis} shows a sensitivity analysis of the uncertainty regularization and action reuse coefficients, emphasizing that \method is quite robust to the choice of these parameters. We also conduct experiments in which we use this uncertainty regularization for TD-MPC2 and BMPC and include the results in \cref{sec:detailed-results}. The results indicate that for BMPC, the performance slightly improves -- except for HumanoidBench -- while for TD-MPC2, the uncertainty regularization leads to a performance decrease for all three domains. The smaller benefits of our proposed uncertainty regularization reported in \cref{tab:ablations} result from smaller value approximation errors for BMPC as discussed in \cref{sec:analysis-dog-run}. Additionally, we provide an analysis of the planner gradients when integrating our method into Dreamer in \cref{sec:gradient-analysis}, which suggests that \method is more robust compared to the previously proposed Grad-MPC.

We further evaluate the performance of fully trained BMPC agents with gradient-based MPC when varying the number of candidates, the number of optimization iterations, and the planning horizon. The results for Acrobot Swingup, Humanoid Run and Slide are shown in \cref{fig:planning-params}. All other hyperparameters are fixed to their default value when varying one. While we use a single set of hyperparameters across all environments, algorithms, and for state-based and visual observations, we find that dynamically adjusting the planning parameters can help to further improve performance. The parameter sweep also shows that increasing the horizon and the number of optimization iterations does not necessarily always increase the performance further, but can also impair the performance for some environments, which may result from an inaccurate model, especially when using a longer prediction horizon than the one used for training the model.

\subsection{Inference Time}
We benchmark inference times of the different methods on a single Nvidia GeForce RTX 4090 GPU. The results in \cref{tab:times} show that \method is about as fast as MPPI for lower dimensional problems, potentially enabling its usage for real-world robotics applications, which require high control frequencies. Although the inference time increases for high-dimensional problems, our method is still significantly faster than Grad-MPC \citep{s_v_gradient-based_2023} for example, which samples hundreds of action sequences from a Gaussian and optimizes each candidate solution for multiple iterations by using gradient ascent. The corresponding inference times are shown in \cref{tab:dreamer-times}.

\begin{scriptsize}
    \begin{table}[htpb]
    \centering
      \caption{\textbf{Inference times of different methods for Acrobot Swingup.} Mean and standard deviation for three random seeds and ten test episodes per seed.}
      \renewcommand{\arraystretch}{1.25}
      \scriptsize %
      \resizebox{0.6\linewidth}{!}{
        \centering
        \begin{tabular}{@{}lc@{}}
          \toprule
          \textbf{Method} & \textbf{Inference time} \\
          \midrule
          BMPC & 18.77 $\pm$ 0.11 \unit{\milli\second} \\ 
          \method (BMPC) & \textbf{18.15 $\pm$ 0.12 \unit{\milli\second}} \\
          TD-MPC2 & 20.83 $\pm$ 0.14 \unit{\milli\second} \\ 
          \bottomrule
        \end{tabular}
      }
      \label{tab:times}  
      \vspace*{-1ex}
    \end{table}
\end{scriptsize}

\section{Conclusion}
We propose \method, a novel method for gradient-based planning with a learned policy network and world model, which incorporates amortization of optimization iterations over time and uncertainty to overcome the limitations of previously proposed gradient-based MPC methods, namely worse performance compared to their gradient-free equivalents and high computational costs while also achieving state-of-the-art performance. We empirically evaluate our method on a broad set of diverse tasks from different domains, including visual observations, to demonstrate its effectiveness. Overall, our results highlight that gradient-based trajectory optimization with a learned world model has the potential to significantly improve the performance of model-based RL algorithms. 

\section*{Acknowledgements}
This work was funded by German Research Foundation (DFG) grant BE 2556/16-2 (Research Unit FOR 2535 Anticipating Human Behavior) and by the Federal Ministry of Research, Technology and Space of Germany (BMFTR) within the WestAI - AI Service Center West, grant no. 16IS22094A and within the Robotics Institute Germany, grant no. 16ME0999. Computational resources were provided by the German AI Service Center WestAI.

\section*{Impact Statement}
This paper presents work whose goal is to advance the field of machine learning by introducing \method, a gradient-based MPC method for planning with world models. This advancement is particularly relevant for domains such as robotics or autonomous systems, where both prediction and planning are important for safe interaction with the environment. Thus, the deployment of such a technology to real systems requires careful consideration of ethical and safety-relevant aspects such as out-of-distribution generalization or a lack of transparency in the decision-making process.

\bibliography{icml2026_conference}
\bibliographystyle{icml2026}

\newpage
\appendix
\onecolumn
\section{Limitations and Future Work}
\textbf{Fixed optimization parameters.} Our experiments suggest that it may be beneficial to dynamically adapt the optimization parameters such as the action optimization step size and number of iterations to further improve the performance, especially for high-dimensional problems. 

\textbf{Policy prior.} We find that gradient-based MPC requires a high-quality policy prior for efficient planning, which may not always be available. This can potentially be addressed by more sophisticated planning architectures.

\textbf{Future work.} As our current approach is applied to single-task problems, it would also be interesting to extend it to multi-task world models to evaluate its potential in this setting.

\section{Environment Details}\label{sec:environment-details}
We evaluate our method on a total of 24 continuous control tasks from three different domains: eight environments from the Deep Mind Control suite, including four high-dimensional locomotion tasks, eight environments from HumanoidBench, and eight environments from Meta-World. All three domains are infinite-horizon continuous control environments for which we use a fixed episode length, an action repeat of 2 for the DeepMind Control Suite and Meta-World and 1 for HumanoidBench, and no termination conditions. We follow the success definition of \citet{hansen_td-mpc2_2024}. This section provides an overview and details for all tasks considered, including their observation and action dimensions.

\begin{figure}[htpb]
    \centering
    \captionsetup[subfigure]{labelformat=empty}
    \begin{subfigure}[ht]{0.2\textwidth}
        \centering
        \includegraphics[width=\textwidth]{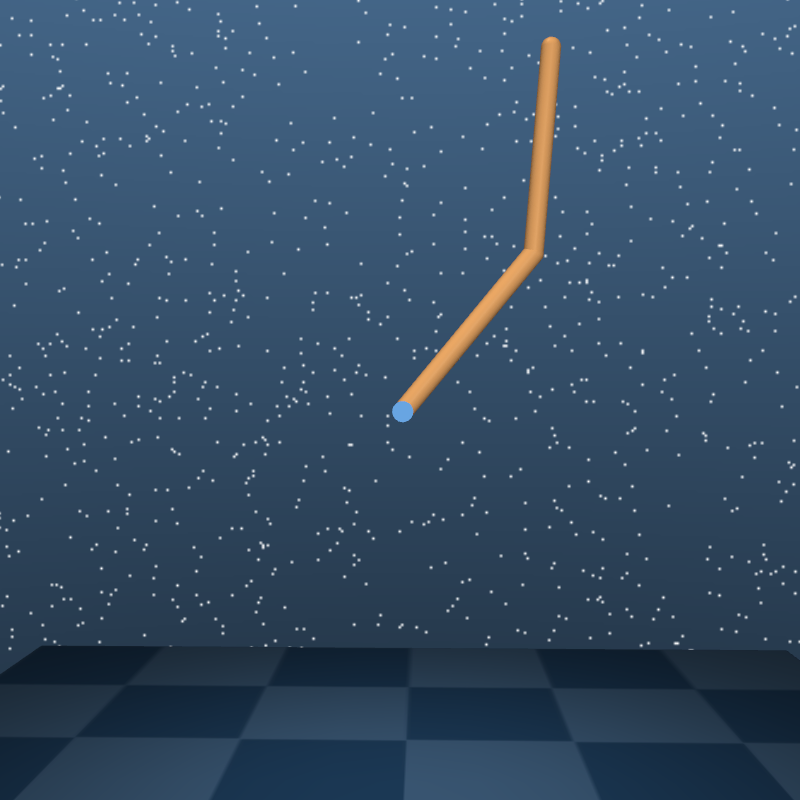}
        \vspace{-1.5em}
        \caption{Acrobot}
    \end{subfigure}
    \hfill
    \begin{subfigure}[ht]{0.2\textwidth}
        \centering
        \includegraphics[width=\textwidth]{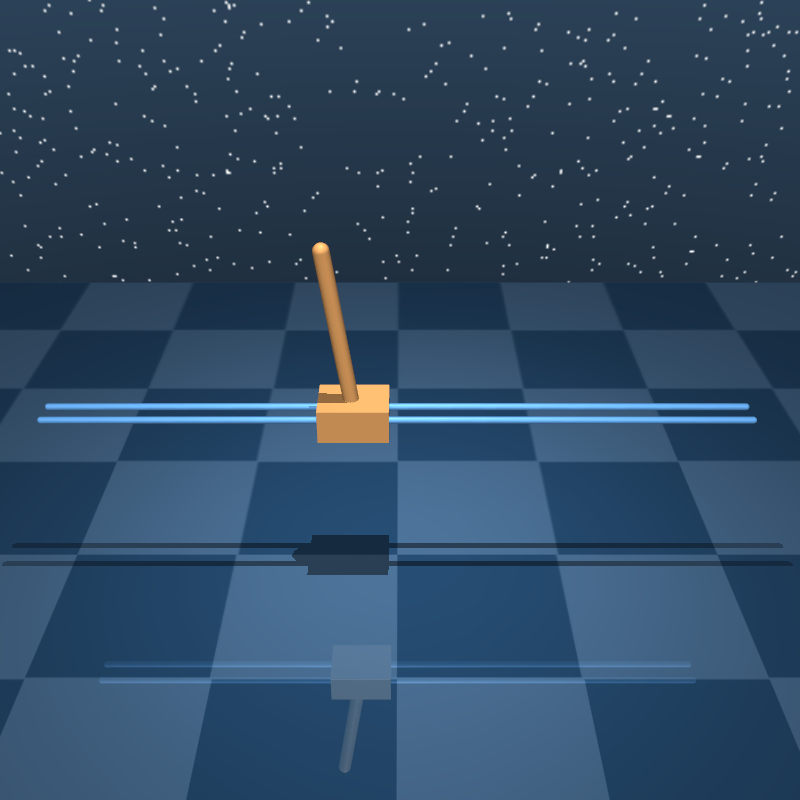}
        \vspace{-1.5em}
        \caption{Cartpole}
    \end{subfigure}
    \hfill
    \begin{subfigure}[ht]{0.2\textwidth}
        \centering
        \includegraphics[width=\textwidth]{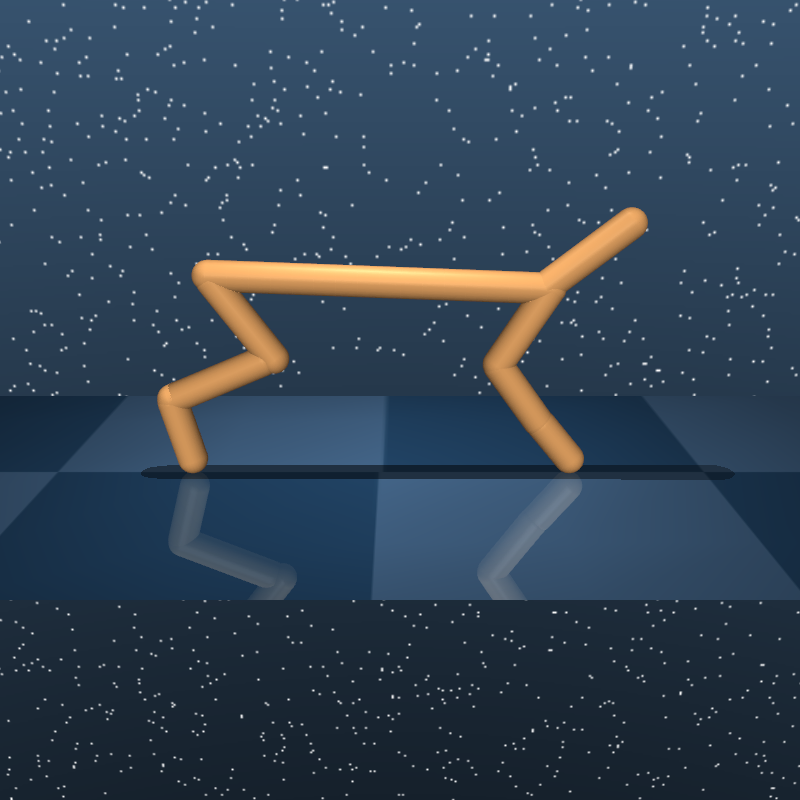}
        \vspace{-1.5em}
        \caption{Cheetah}
    \end{subfigure}
    \hfill
    \begin{subfigure}[ht]{0.2\textwidth}
        \centering
        \includegraphics[width=\textwidth]{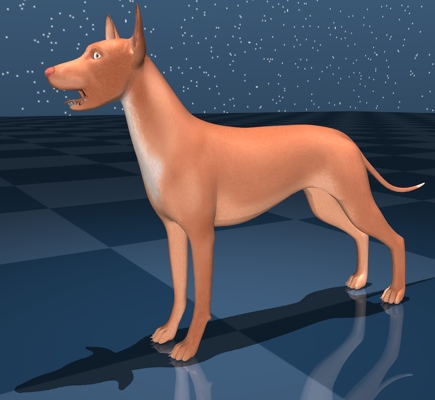}
        \vspace{-1.5em}
        \caption{Dog}
    \end{subfigure}
    \\
    \begin{subfigure}[ht]{0.2\textwidth}
        \centering
        \includegraphics[width=\textwidth]{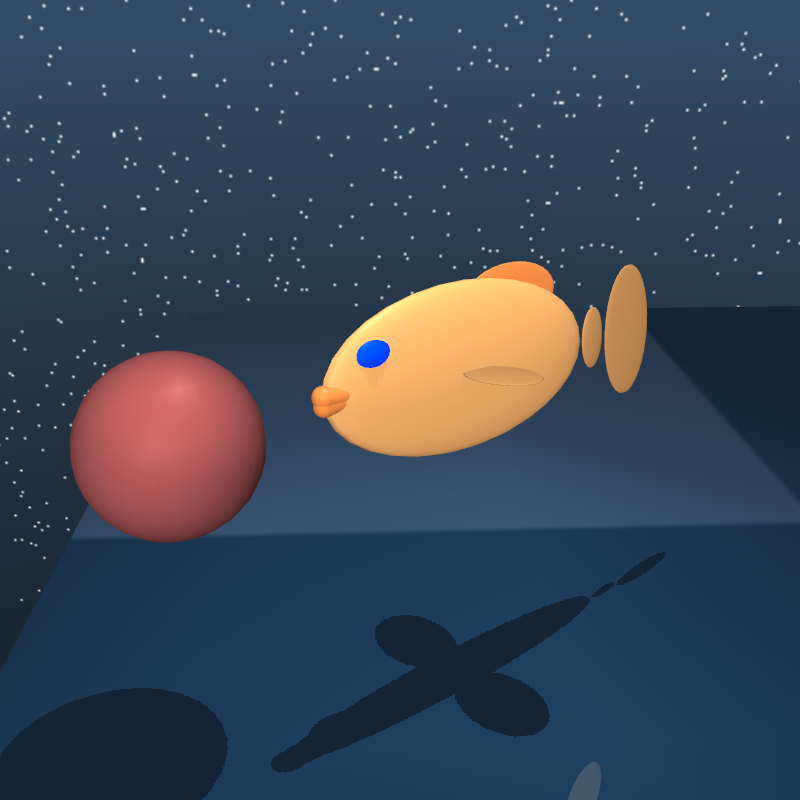}
        \vspace{-1.5em}
        \caption{Fish}
    \end{subfigure}
    \hfill
    \begin{subfigure}[ht]{0.2\textwidth}
        \centering
        \includegraphics[width=\textwidth]{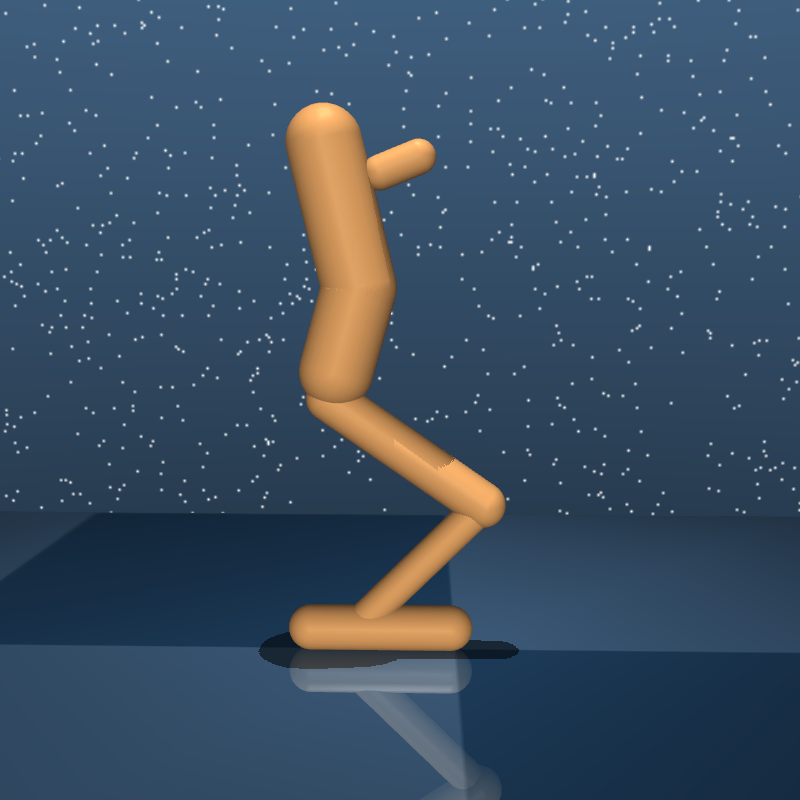}
        \vspace{-1.5em}
        \caption{Hopper}
    \end{subfigure}
    \hfill
    \begin{subfigure}[ht]{0.2\textwidth}
        \centering
        \includegraphics[width=\textwidth]{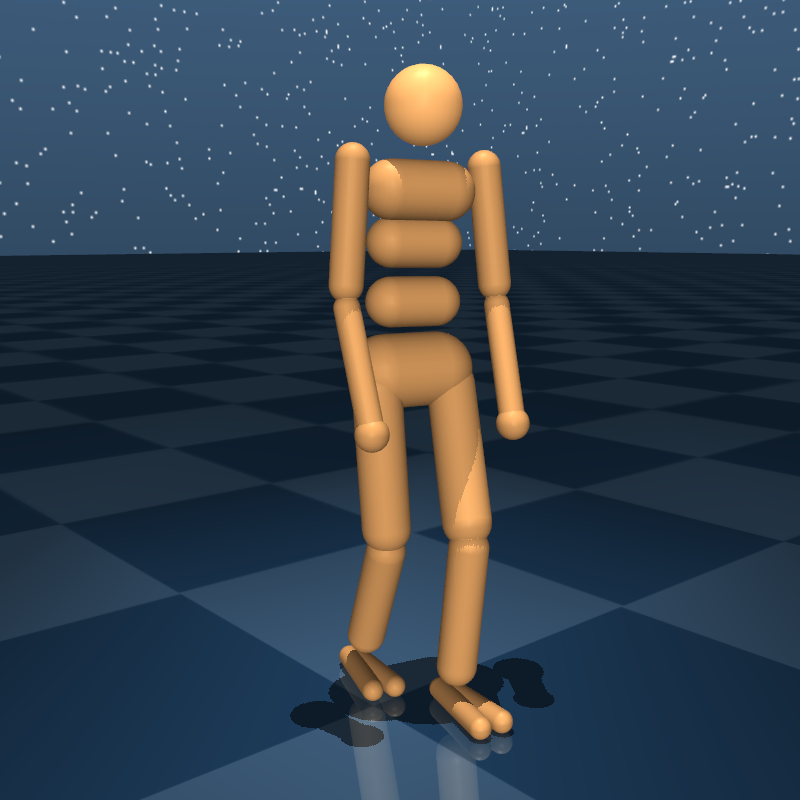}
        \vspace{-1.5em}
        \caption{Humanoid}
    \end{subfigure}
    \hfill
    \begin{subfigure}[ht]{0.2\textwidth}
        \centering
        \includegraphics[width=\textwidth]{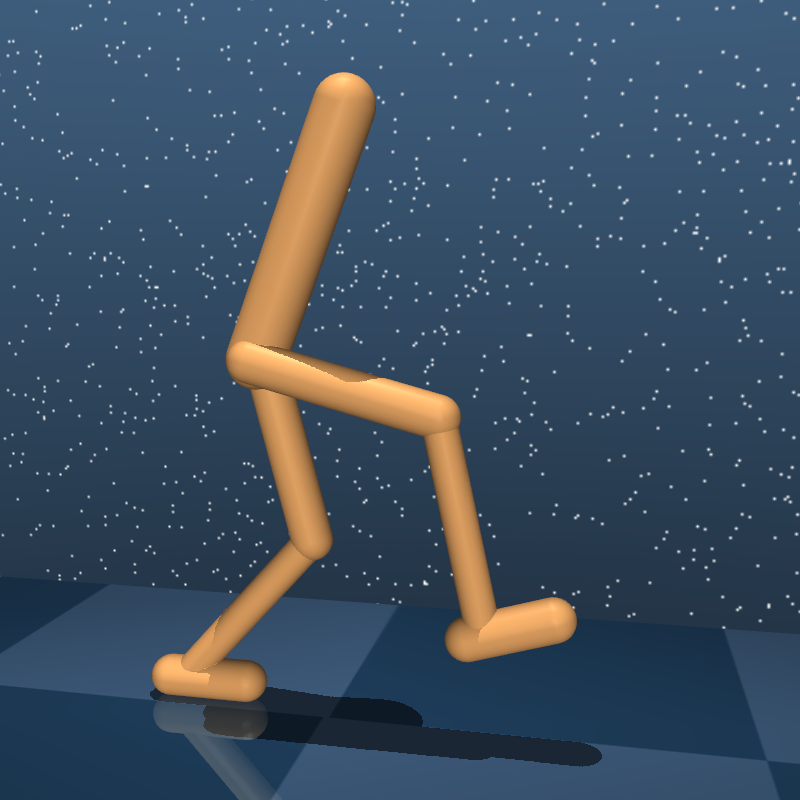}
        \vspace{-1.5em}
        \caption{Walker}
    \end{subfigure}
    \caption{
        \textbf{DeepMind Control Suite benchmarking domains \citep{tassa_dm_control_2020}.}
    }
    \label{fig:dmc-domains}
\end{figure}

\begin{table}[H]
  \centering
  \caption{\textbf{Overview of DeepMind Control Suite tasks.} Classification is based on \citet{hubert_2021,yarats_2022}}
  \begin{tabular}{lllcc}
    \toprule
    \textbf{Task} & \textbf{Difficulty} & \textbf{Reward} & \textbf{dim($\mathcal{S}$)} & \textbf{dim($\mathcal{A}$)} \\
    \midrule
    Acrobot Swingup & hard & dense & 6 & 1 \\
    Cartpole Swingup Sparse & easy & sparse & 5 & 1 \\
    Dog Run & hard & dense & 223 & 38 \\
    Dog Walk & hard & dense & 223 & 38 \\
    Fish Swim & medium & dense & 24 & 5 \\
    Hopper Hop & medium & dense & 15 & 4 \\ 
    Humanoid Run & hard & dense & 67 & 24 \\
    Humanoid Walk & hard & dense & 67 & 24 \\
    \bottomrule
  \end{tabular}
  \label{tab:dmc}  
\end{table}

We consider following eight tasks from Meta-World:
\begin{itemize}
  \item Assembly: Pick up a nut and place it onto a peg (peg and nut positions are randomized),
  \item Button Press: Press a button (button positions are randomized),
  \item Disassemble: Remove a nut from a peg (peg and nut positions are randomized),
  \item Lever Pull: Pull a lever down 90 degrees (lever positions are randomized),
  \item Pick Place Wall: Pick a puck, bypass a wall and place the puck (puck and goal positions are randomized),
  \item Push Back: Push the puck to a goal (puck and goal positions are randomized),
  \item Shelf Place: Pick and place a puck onto a shelf (puck and shelf positions are randomized),
  \item Window Open: Push and open a window (window positions are randomized).
\end{itemize}
All tasks from Meta-World share the same embodiment, observation space (dim($\mathcal{S}$) = 39) and action space (dim($\mathcal{A}$) = 4). Please refer to \citet{metaworld_2019} for the definitions of the reward functions and success metrics used in the Meta-World tasks.

\begin{figure}[htpb]
    \centering
    \captionsetup[subfigure]{labelformat=empty}
    \begin{subfigure}[ht]{0.2\textwidth}
        \centering
        \includegraphics[width=\textwidth]{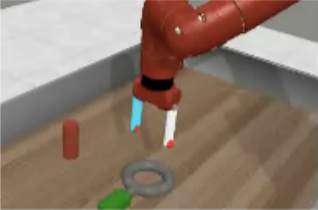}
        \vspace{-1.5em}
        \caption{Assembly}
    \end{subfigure}
    \hfill
    \begin{subfigure}[ht]{0.2\textwidth}
        \centering
        \includegraphics[width=\textwidth]{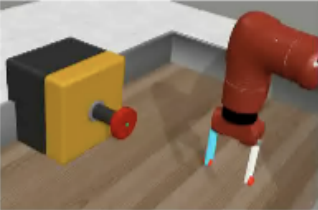}
        \vspace{-1.5em}
        \caption{Button Press}
    \end{subfigure}
    \hfill
    \begin{subfigure}[ht]{0.2\textwidth}
        \centering
        \includegraphics[width=\textwidth]{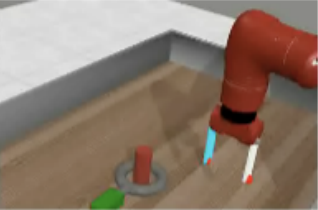}
        \vspace{-1.5em}
        \caption{Disassemble}
    \end{subfigure}
    \hfill
    \begin{subfigure}[ht]{0.2\textwidth}
        \centering
        \includegraphics[width=\textwidth]{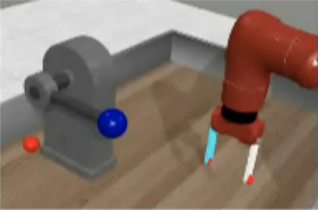}
        \vspace{-1.5em}
        \caption{Lever Pull}
    \end{subfigure}
    \\
    \begin{subfigure}[ht]{0.2\textwidth}
        \centering
        \includegraphics[width=\textwidth]{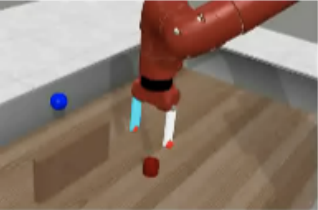}
        \vspace{-1.5em}
        \caption{Pick Place Wall}
    \end{subfigure}
    \hfill
    \begin{subfigure}[ht]{0.2\textwidth}
        \centering
        \includegraphics[width=\textwidth]{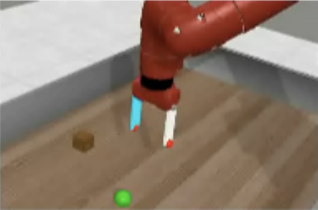}
        \vspace{-1.5em}
        \caption{Push Back}
    \end{subfigure}
    \hfill
    \begin{subfigure}[ht]{0.2\textwidth}
        \centering
        \includegraphics[width=\textwidth]{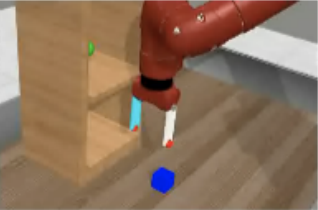}
        \vspace{-1.5em}
        \caption{Shelf Place}
    \end{subfigure}
    \hfill
    \begin{subfigure}[ht]{0.2\textwidth}
        \centering
        \includegraphics[width=\textwidth]{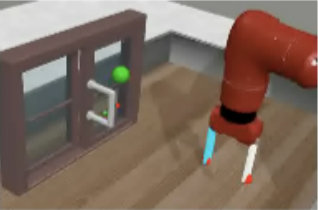}
        \vspace{-1.5em}
        \caption{Window Open}
    \end{subfigure}
    \caption{
        \textbf{Meta-World manipulation tasks.} We consider eight different tasks from the Meta-World Benchmark.
    }
    \label{fig:mw-domains}
\end{figure}

\begin{figure}[htpb]
    \centering
    \captionsetup[subfigure]{labelformat=empty}
    \begin{subfigure}[ht]{0.2\textwidth}
        \centering
        \includegraphics[width=\textwidth]{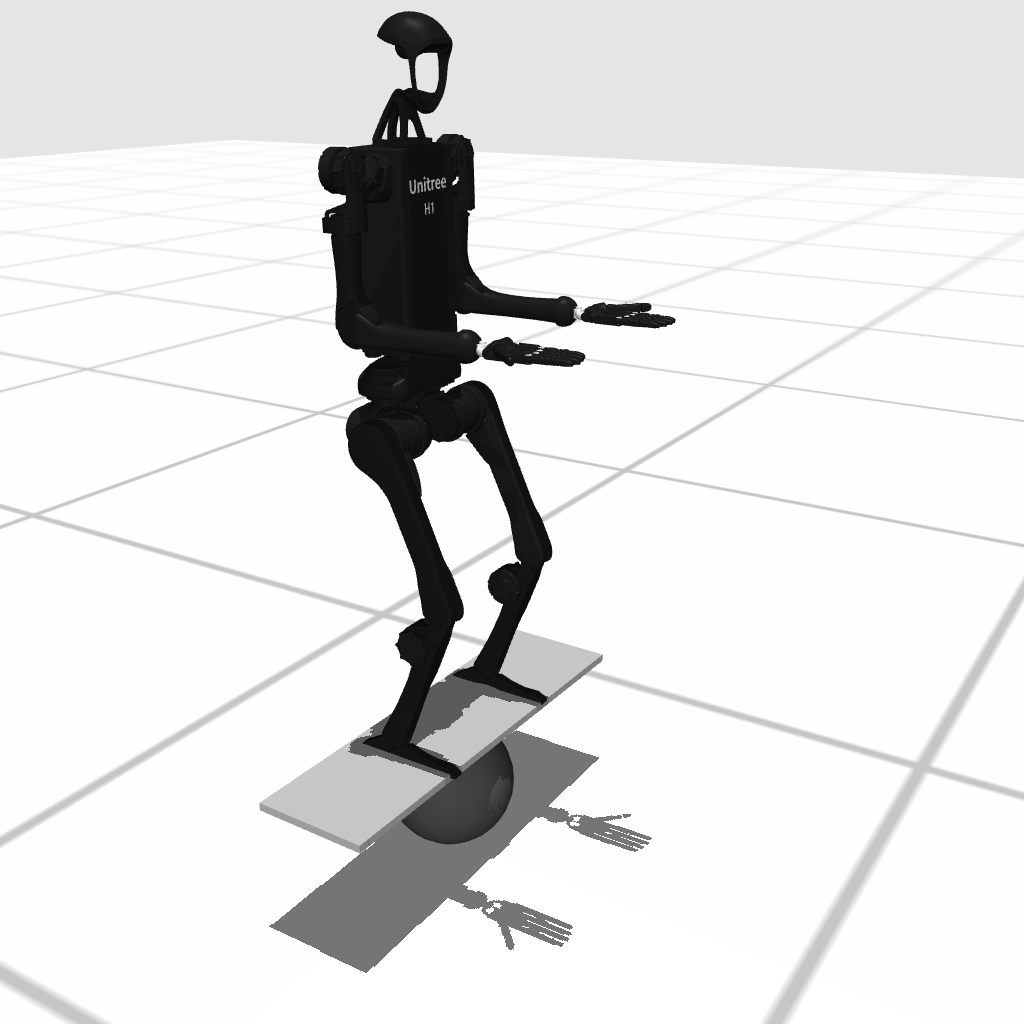}
        \vspace{-1.5em}
        \caption{Balance}
    \end{subfigure}
    \hfill
    \begin{subfigure}[ht]{0.2\textwidth}
        \centering
        \includegraphics[width=\textwidth]{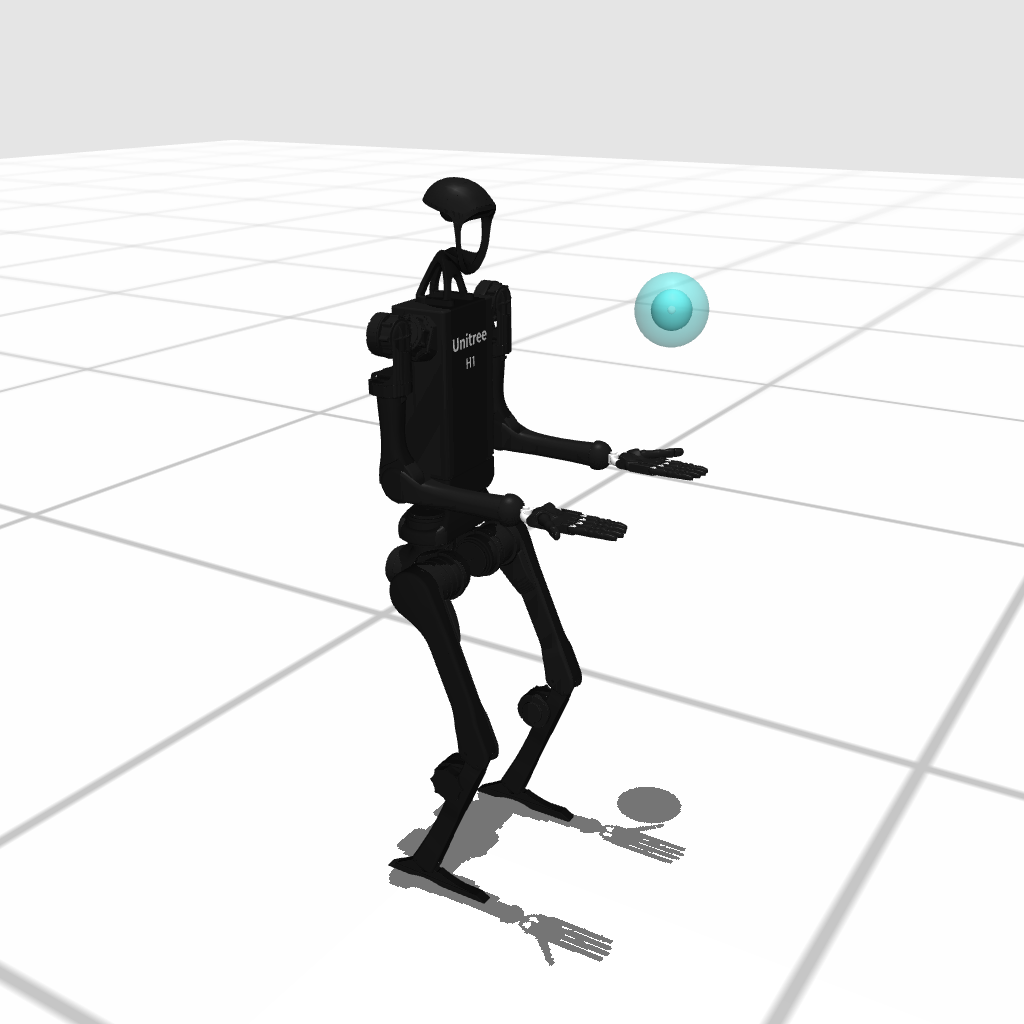}
        \vspace{-1.5em}
        \caption{Reach}
    \end{subfigure}
    \hfill
    \begin{subfigure}[ht]{0.2\textwidth}
        \centering
        \includegraphics[width=\textwidth]{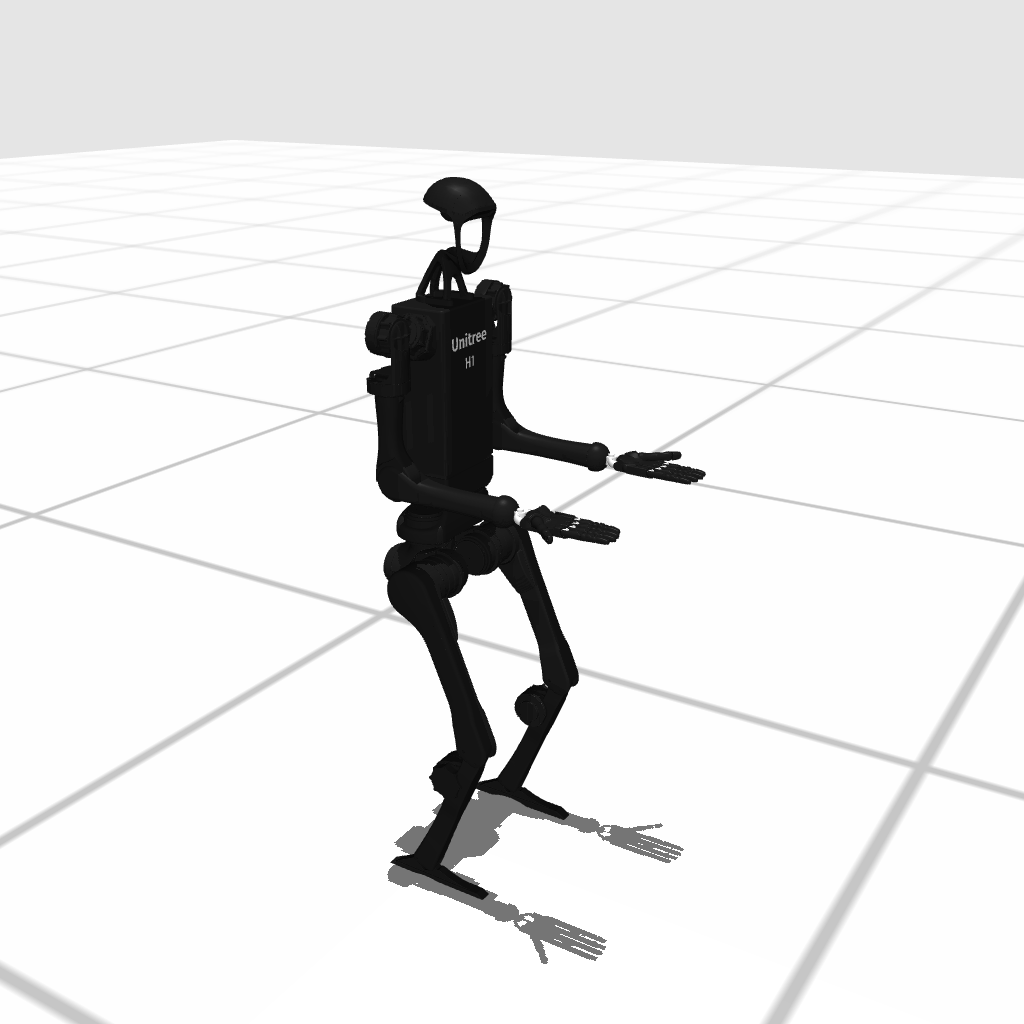}
        \vspace{-1.5em}
        \caption{Run}
    \end{subfigure}
    \hfill
    \begin{subfigure}[ht]{0.2\textwidth}
        \centering
        \includegraphics[width=\textwidth]{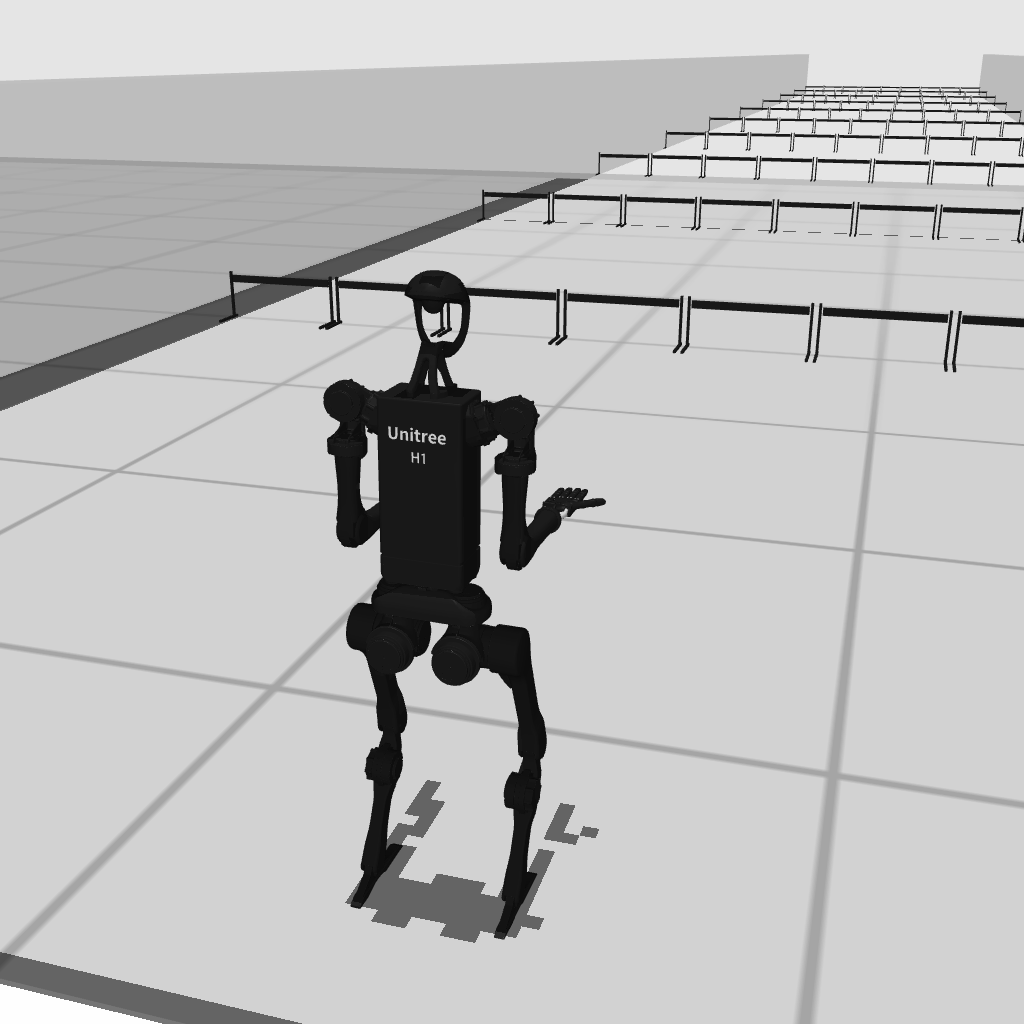}
        \vspace{-1.5em}
        \caption{Hurdle}
    \end{subfigure}
    \\
    \begin{subfigure}[ht]{0.267\textwidth}
        \centering
        \includegraphics[width=\textwidth]{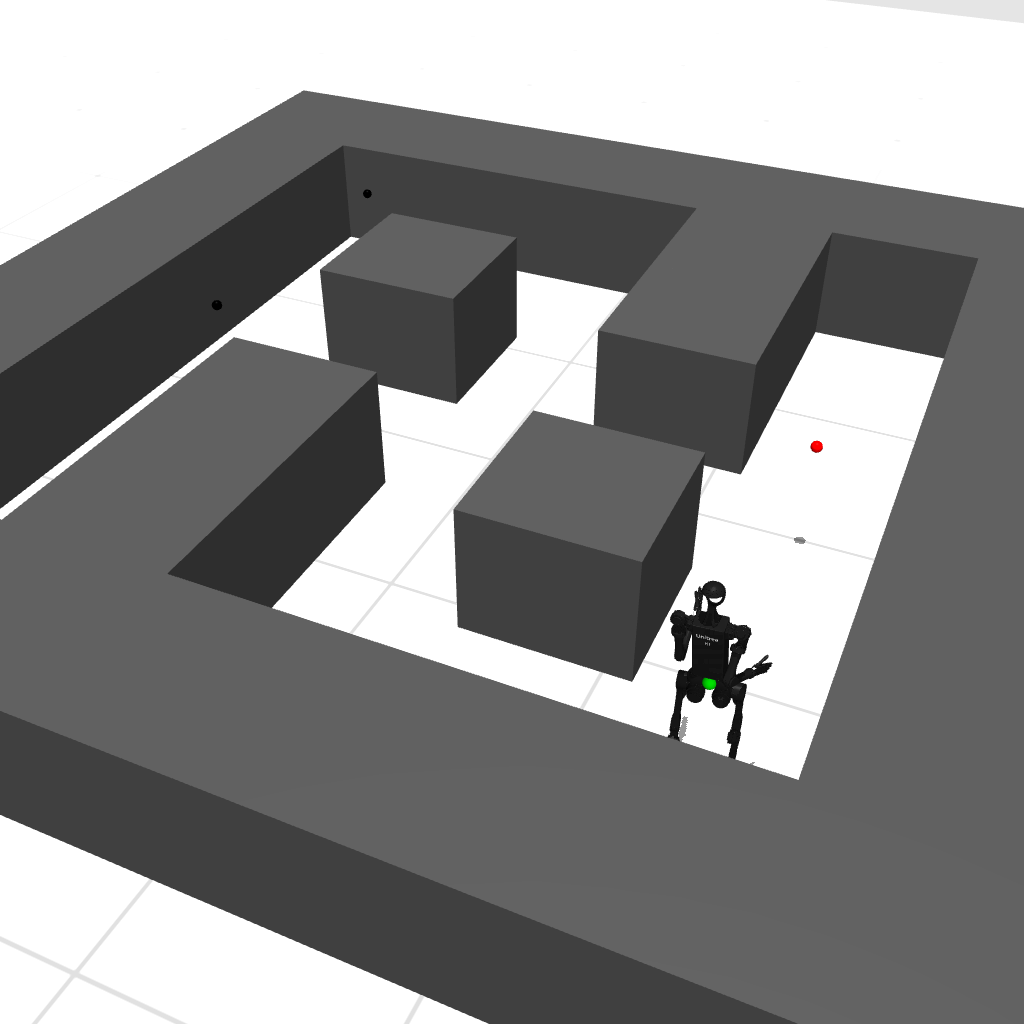}
        \vspace{-1.5em}
        \caption{Maze}
    \end{subfigure}
    \vspace{0.6em}
    \hfill
    \begin{subfigure}[ht]{0.267\textwidth}
        \raggedleft
        \includegraphics[width=\textwidth]{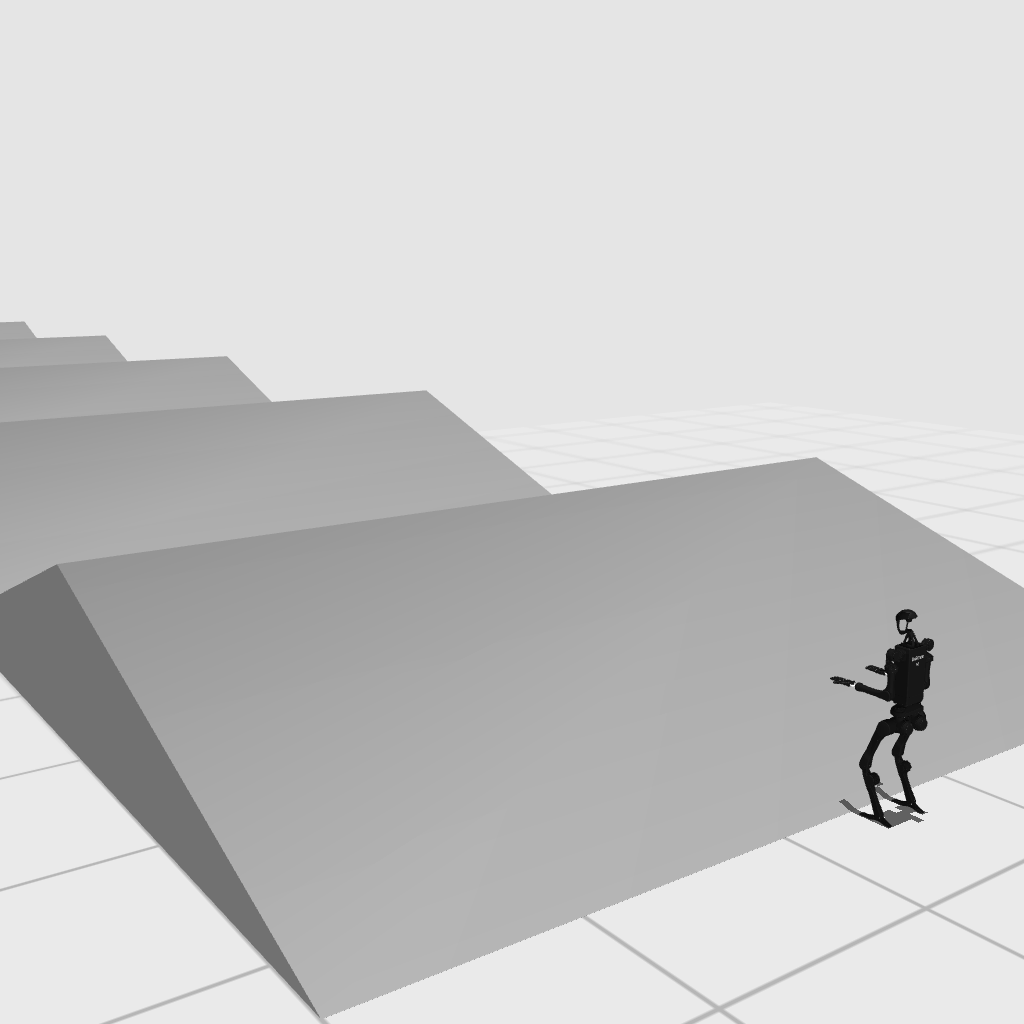}
        \vspace{-1.5em}
        \caption{Slide}
    \end{subfigure}
    \hfill
    \begin{subfigure}[ht]{0.267\textwidth}
        \centering
        \includegraphics[width=\textwidth]{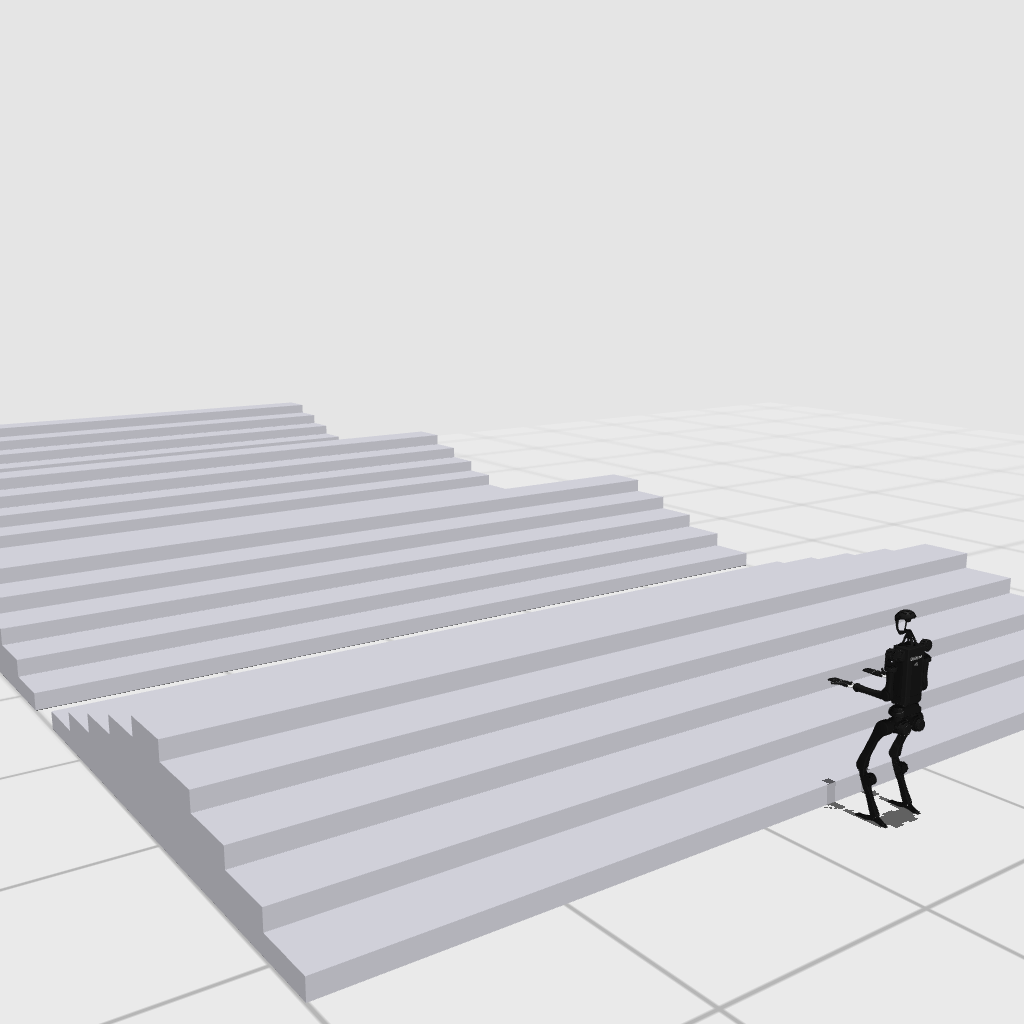}
        \vspace{-1.5em}
        \caption{Stair}
    \end{subfigure}
    \caption{
        \textbf{HumanoidBench locomotion tasks.} We consider eight tasks from the HumanoidBench locomotion benchmark that cover a wide variety of interactions and difficulties. This figure illustrates an initial state for each task.
    }
    \label{fig:locomotion_task_suite}
\end{figure}

We further consider following eight tasks from the twelve benchmarking locomotion tasks of HumanoidBench: 
\begin{itemize}
  \item Balance Hard: Balance on the unstable board while the spherical pivot beneath the board does move,
  \item Balance Simple: Balance on the unstable board while the spherical pivot beneath the board does not move,
  \item Hurdle: Keep forward velocity close to 5 m/s while successfully overcoming hurdles,
  \item Maze: Reach the goal position in a maze by taking multiple turns at the intersections,
  \item Reach: Reach a randomly initialized 3D point with the left hand,
  \item Run: Run forward at a speed of 5 m/s,
  \item Slide: Walk over an iterating sequence of upward and downward slides at 1 m/s,
  \item Stair: Traverse an iterating sequence of upward and downward stairs at 1 m/s.
\end{itemize}
Visualizations of the tasks are shown in \cref{fig:locomotion_task_suite}.

The benchmark uses the Unitree H1 with two dexterous hands. The observation and action spaces, and degrees of freedom of the robot system with the dexterous hands are summarized in \cref{tab:humanoid-spec}.
  \begin{table}[H]
    \centering
    \caption{\textbf{Humanoid robot specifications with two hands.}}
    \begin{tabular}{lc}
      \toprule
      \textbf{Parameter} & \textbf{Value} \\
      \midrule
      Observation space & 151 \\
      Action space & 61 \\
      DoF (body) & 25 \\
      DoF (hands) & 50 \\
      \bottomrule
    \end{tabular}
    \label{tab:humanoid-spec}  
  \end{table}

\section{Implementation Details}\label{sec:implementation-details}

\textbf{TD-MPC2 implementation.} We use the official implementation of TD-MPC2 available at \url{https://github.com/nicklashansen/tdmpc2}, and use the default hyperparameters suggested by the authors. A complete list of hyperparameters is provided in \cref{tab:tdmpc2-hparams}. Details on TD-MPC2 can be found in \cref{sec:tdmpc2}. For the experiments with (pre-)trained models, we use the models provided by \citet{hansen_td-mpc2_2024} for the DeepMind Control Suite and Meta-World, except for Cartpole Swingup Sparse, Dog Run, Dog Walk, Humanoid Run and Humanoid Walk because some checkpoints cannot be loaded after code restructuring\footnote{cf. \url{https://github.com/nicklashansen/tdmpc2/issues/23}}. Thus, we train new models from scratch for these tasks as well as for HumanoidBench.

\textbf{BMPC implementation.} We use the official implementation of BMPC from \url{https://github.com/wertyuilife2/bmpc}, and use the default hyperparameters suggested by the authors. Since the code is based on the official TD-MPC2 codebase and incorporates both algorithms, we use this implementation as a basis for our method. Details on BMPC are provided in \cref{sec:bmpc}. We train all BMPC models from scratch.

\textbf{Dreamer-v3 baseline implementation.} We use the official implementation of Dreamer-v3 available at \url{https://github.com/danijar/dreamerv3}. We follow the decision of \citet{hansen_td-mpc2_2024} and use the authors' suggested hyperparameters for proprioceptive control (DeepMind Control Suite). Please refer to \citet{hafner_mastering_2025} and \citet{hansen_td-mpc2_2024} for a complete list of hyperparameters and implementation details.

\textbf{SAC baseline implementation.} We use the SAC implementation from \url{https://github.com/denisyarats/pytorch_sac} as in the TD-MPC \citep{hansen_temporal_2022} paper, and use the hyperparameters suggested by the authors. Please refer to their paper for a complete list of hyperparameters.

\subsection{TD-MPC2}\label{sec:tdmpc2}

\textbf{Architectural details.} All components of TD-MPC2 are implemented as multi-layer perceptrons (MLPs). The encoder $h$ contains a variable number of layers ($2-5$), depending on the architecture size; all other components are 3-layer MLPs. Intermediate layers consist of a linear layer followed by LayerNorm and a Mish activation function. The latent representation is normalized as a simplicial embedding. $Q$-functions additionally use Dropout. We summarize the TD-MPC2 architecture for the $5$M parameter \texttt{base} (default for online RL) model size using PyTorch-like notation:
\begin{codesnippet}
Encoder parameters: 167,936                                                                                                                                                                                                                                            
Dynamics parameters: 843,264                                                                                                                                                                                                                                           
Reward parameters: 631,397                                                                                                                                                                                                                                             
Policy parameters: 582,668                                                                                                                                                                                                                                             
Q parameters: 3,156,985                                                                                                                                                                                                                                                
Task parameters: 7,680                                                                                                                                                                                                                                                 
Total parameters: 5,389,930

Architecture: TD-MPC2 base 5M(                                                                                                                                                                                                                                                    
  (task_embedding): Embedding(T, 96, max_norm=1)                                                                                                                                                                                                                           
  (encoder): ModuleDict(                                                                                                                                                                                                                                              
    (state): Sequential(                                                                                                                                                                                                                                               
      (0): NormedLinear(in_features=S+T, out_features=256, act=Mish)                                                                                                                                                                                        
      (1): NormedLinear(in_features=256, out_features=512, act=SimNorm)                                                                                                                                                                                     
    )                                                                                                                                                                                                                                                                  
  )                                                                                                                                                                                                                                                                    
  (dynamics): Sequential(                                                                                                                                                                                                                                             
    (0): NormedLinear(in_features=512+T+A, out_features=512, act=Mish)                                                                                                                                                                                          
    (1): NormedLinear(in_features=512, out_features=512, act=Mish)
    (2): NormedLinear(in_features=512, out_features=512, act=SimNorm)
  )
  (reward): Sequential(
    (0): NormedLinear(in_features=512+T+A, out_features=512, act=Mish)
    (1): NormedLinear(in_features=512, out_features=512, act=Mish)
    (2): Linear(in_features=512, out_features=101,)
  )
  (pi): Sequential(
    (0): NormedLinear(in_features=512+T, out_features=512, act=Mish)
    (1): NormedLinear(in_features=512, out_features=512, act=Mish)
    (2): Linear(in_features=512, out_features=2A, bias=True)
  )
  (Qs): Vectorized ModuleList(
    (0-4): 5 x Sequential(
      (0): NormedLinear(in_features=512+T+A, out_features=512, dropout=0.01, act=Mish)
      (1): NormedLinear(in_features=512, out_features=512, act=Mish)
      (2): Linear(in_features=512, out_features=101, bias=True)
    )
  )
)
\end{codesnippet}
where \texttt{S} is the input dimensionality, \texttt{T} is the number of tasks, and \texttt{A} is the action space. We exclude the task embedding \texttt{T} from single-task experiments. The exact parameter counts listed above are for \texttt{S}$=39$, \texttt{T}$=80$, and \texttt{A}$=6$. Since we only perform single-task experiments in this work, all models contain around 5M parameters for TD-MPC2.

\textbf{Policy-guided MPC.} TD-MPC2 uses Model Predictive Path Integral (MPPI) \citep{williams_mppi_2015, williams_information_2017} for local trajectory optimization, which is a gradient-free, sampling-based MPC method. MPPI iteratively samples action sequences $(a_{t}, a_{t+1}, \dots, a_{t+H})$ of length $H$ from $\mathcal{N}(\mu,\sigma^{2})$, evaluates their expected return by rolling out latent trajectories with the model, and updates the parameters $\mu,\sigma$ of a time-dependent multivariate Gaussian with diagonal covariance based on a weighted average such that the expected return is maximized. This iterative optimization procedure is repeated for a fixed number of iterations and the first action $a_{t} \sim \mathcal{N}(\mu^{*}_{t}, \sigma^{*}_{t})$ is applied to the environment. TD-MPC2 augments the sampling procedure with samples from the policy prior $\pi_{\theta}$ and warm-starts the optimization procedure by initializing $(\mu,\sigma)$ as the solution of the previous step shifted by one to improve performance. Please refer to \citet{hansen_temporal_2022} for more details.

\vspace{0.4in}
\begin{center}
    \textcolor{gray}{\textbf{\textemdash~Appendices continue on next page~\textemdash}}
\end{center}

\clearpage

\begin{table}[h!tpb]
\centering
\parbox{\textwidth}{
\caption{\textbf{TD-MPC2 hyperparameters.} We use the same hyperparameters across all tasks. Certain hyperparameters are set automatically using heuristics.}
\label{tab:tdmpc2-hparams}
\centering
\begin{tabular}{@{}ll@{}}
\toprule
\textbf{Hyperparameter}         & \textbf{Value} \\ \midrule
\textbf{\textcolor{myblue}{\underline{\smash{Planning}}}} \\
Horizon ($H$)                   & $3$ \\
Iterations                      & $6~(+ 2~\textnormal{if}~\|\mathcal{A}\| \ge 20)$ \\
Population size                 & $512$ \\
Policy prior samples            & $24$ \\
Number of elites                & $64$ \\
Minimum std.                    & $0.05$ \\
Maximum std.                    & $2$ \\
Temperature                     & $0.5$ \\
Momentum                        & No \\
\\
\textbf{\textcolor{myblue}{\underline{\smash{Policy prior}}}} \\
Log std. min.                   & $-10$ \\
Log std. max.                   & $2$ \\
\\
\textbf{\textcolor{myblue}{\underline{\smash{Replay buffer}}}} \\
Capacity                        & $1,000,000$ \\
Sampling                        & Uniform \\
\\
\textbf{\textcolor{myblue}{\underline{\smash{Architecture (5M)}}}} \\
Encoder dim                     & $256$ \\
MLP dim                         & $512$ \\
Latent state dim                & $512$ \\
Task embedding dim              & $96$ \\
Task embedding norm             & $1$ \\
Activation                      & LayerNorm + Mish \\
$Q$-function dropout rate       & $1\%$ \\
Number of $Q$-functions         & $5$ \\
Number of reward/value bins     & $101$ \\
SimNorm dim ($V$)               & $8$ \\
SimNorm temperature ($\tau$)    & $1$ \\
\\
\textbf{\textcolor{myblue}{\underline{\smash{Optimization}}}} \\
Update-to-data ratio            & $1$ \\
Batch size                      & $256$ \\
Joint-embedding coef.           & $20$ \\
Reward prediction coef.         & $0.1$ \\
Value prediction coef.          & $0.1$ \\
Temporal coef. ($\lambda$)      & $0.5$ \\
$Q$-fn. momentum coef.          & $0.99$ \\
Policy prior entropy coef.      & $1\times10^{-4}$ \\
Policy prior loss norm.         & Moving $(5\%, 95\%)$ percentiles \\
Optimizer                       & Adam \citep{kingma_adam_2017} \\
Learning rate                   & $3\times10^{-4}$ \\
Encoder learning rate           & $1\times10^{-4}$ \\
Gradient clip norm              & $20$ \\
Discount factor                 & Heuristic \\
Seed steps                      & Heuristic \\ \bottomrule
\end{tabular}%
}
\end{table}

\subsection{BMPC}\label{sec:bmpc}
\textbf{Architectural details.} The main architectural difference of BMPC to TD-MPC2 is that it uses two $V$-functions instead of five $Q$-functions:
\begin{codesnippet}
V parameters: 1,256,650

Total parameters: 3,489,595

Architecture: Difference BMPC to TD-MPC2
  (
    (Vs): Vectorized ModuleList(
      (0-1): 2 x Sequential(
        (0): NormedLinear(in_features=512+T, out_features=512, dropout=0.01, act=Mish)
        (1): NormedLinear(in_features=512, out_features=512, act=Mish)
        (2): Linear(in_features=512, out_features=101, bias=True)
      )
    )
  )
\end{codesnippet}

\textbf{Policy learning.} The policy is learned using the following objective:
\begin{equation}
\label{eq:bmpc-policy-obj}
\begin{aligned}
\mathcal{L}_{\pi}(\theta) 
&\doteq \underset{(\mathbf{s},\mathbf{a})_{0:H} \sim \mathcal{B}}{\mathbb{E}} \left[ \sum_{t=0}^{H} \lambda^{t} \left[ \mathrm{KL}(\pi_{\text{MPC}}(\cdot \vert h(\mathbf{s}_{t}),\pi_{\theta}),\pi_{\theta}(\cdot \vert \mathbf{z}_{t}))/\mathrm{max}(1,S)-\beta \mathcal{H}(\pi_{\theta}(\cdot \vert \mathbf{z}_{t})) \right] \right], \\
\mathbf{z}_{0}&=h(\mathbf{s}_{0}), \ \mathbf{z}_{t+1}=d(\mathbf{z}_{t}, \mathbf{a}_{t}), \\ 
S &\doteq \mathrm{EMA}(\mathrm{Per}(\mathrm{KL}(\pi_{\text{MPC}},\pi_{\theta}),95)-\mathrm{Per}(\mathrm{KL}(\pi_{\text{MPC}},\pi_{\theta}),5), 0.99),
\end{aligned}
\end{equation}
where $\mathcal{H}$ is the entropy, $\mathrm{KL}$ is the Kullback-Leibler divergence, $\mathbf{z}_{0:H}$ are latent vectors rolled out using the models $h$ and $d$, and $\beta$ and $\lambda$ are hyperparameters for loss balancing and temporal weighting, respectively. The $\mathrm{KL}$ loss is normalized using moving percentiles $S$, which are commonly used to stabilize training. 

\textbf{Model-based TD-learning.} Since BMPC does not use a SAC-style max-Q approach for policy improvement, the authors decide to learn a state value function $V_{\phi}$ instead of a state-action value function $Q_{\phi}$. The value network is learned by minimizing the cross-entropy loss with respect to the discretized n-step TD-target $\hat{V}$ computed by using the latest model, policy, and target value network:
\begin{equation}
\label{eq:bmpc-value-obj}
\begin{aligned}
\mathcal{L}_{V}(\phi) &\doteq \underset{(\mathbf{s},\mathbf{a})_{0:H} \sim \mathcal{B}}{\mathbb{E}} \left[ \sum_{t=0}^{H} \lambda^{t} \left[ \mathrm{CE}(V_{\phi}(\mathbf{z}_{t}),\hat{V}(h(\mathbf{s}_{t}))) \right] \right], \, \mathbf{z}_{0}=h(\mathbf{s}_{0}), \, \mathbf{z}_{t+1}=d(\mathbf{z}_{t},\mathbf{a}_{t}) \\
\hat{V}(\mathbf{z}_{t}^{\prime}) &\doteq \gamma^{N}V_{\phi^{-}}(\mathbf{z}_{t+N}^{\prime}) + \sum_{k=0}^{N-1} \gamma^{k}R(\mathbf{z}_{t+k}^{\prime},\pi_{\theta}(\mathbf{z}_{t+k}^{\prime})), \, \mathbf{z}_{t+1}^{\prime}=d(\mathbf{z}_{t}^{\prime}, \pi_{\theta}(\mathbf{z}_{t}^{\prime}))
\end{aligned}
\end{equation}
where $N$ is the TD horizon, $\mathbf{z}_{0:H}$ are latent vectors rolled out through the models $h$ and $d$. $\hat{V}$ is the TD-target computed using the model $d,R$ and the policy $\pi_{\theta}$ in an on-policy manner. The authors use a fixed value of $N=1$ to keep compounding errors small.

\textbf{Lazy reanalyze.} BMPC stores imitation targets in the replay buffer and uses lazy reanalyze to avoid costly replanning for all samples during every update to compute the policy objective. For every $k$-th network update, $b$ samples are drawn from the batch and used to get new imitation targets, i.e., the mean and standard deviation of the action distribution $\pi_{t}=\pi_{\text{MPC}}(\cdot \vert h(\mathbf{s}_{t}),\pi_{\theta})$ by replanning. These targets $\pi_{t}$ are then placed back into the replay buffer. Since the replanning is performed independently of the training process, the replay buffer can be approximately seen as an expert dataset and used to sample state-action pairs from it for supervised learning. During replanning, additional noise is added to the policy prior to increase exploration in MPC planning. Thus, the resulting surrogate policy objective with lazy reanalyze can be defined as:
\begin{equation}
\label{eq:policy-obj-reanalyze}
\mathcal{L}_{\pi}^{\text{lazy}}(\theta) \doteq \underset{(\mathbf{s},\mathbf{a},\pi)_{0:H} \sim B}{\mathbb{E}} \left[ \sum_{t=0}^{H} \lambda^{t} \left[ \mathrm{KL}(\pi_{t},\pi_{\theta}(\cdot \vert \mathbf{z}_{t}))/\mathrm{max}(1,S)-\beta \mathcal{H}(\pi_{\theta}(\cdot \vert \mathbf{z}_{t})) \right] \right]
\end{equation}
where $\pi_{t}$ is the expert action distribution from the replay buffer.

\begin{table}[h]
\centering
\parbox{\textwidth}{
\caption{\textbf{BMPC hyperparameters.} We use the same hyperparameters for all tasks. All other hyperparameters are the default TD-MPC2 values.}
\label{tab:bmpc-hparams}
\centering
\begin{tabular}{@{}ll@{}}
\toprule
\textbf{Hyperparameter}         & \textbf{Value} \\ \midrule
\textbf{\textcolor{mygreen}{\underline{\smash{Planning}}}} \\
Horizon                             & $3$                   \\
Replanning horizon                 & $3$                   \\
Lazy reanalyze interval ($k$)       & $10$                  \\
Lazy reanalyze batch size ($b$)     & $20$                  \\ 
\\
\textbf{\textcolor{mygreen}{\underline{\smash{Policy prior}}}} \\
Log std. min.                & $-3$                  \\
Log std. max.                & $1$                   \\
Log std. min. (replanning)  & $-2$                  \\
Log std. max. (replanning)  & $1$                   \\
\\
\textbf{\textcolor{mygreen}{\underline{\smash{Architecture}}}} \\
Number of $V$-functions   & $2$                   \\ 
\\
\textbf{\textcolor{mygreen}{\underline{\smash{Optimization}}}} \\
Batch size                          & $256$                 \\
TD horizon ($N$)                    & $1$                   \\ 
Policy prior entropy coef.          & $1\times10^{-4}$  \\ \bottomrule
\end{tabular}%
}
\end{table}

\vspace{0.4in}
\begin{center}
    \textcolor{gray}{\textbf{\textemdash~Appendices continue on next page~\textemdash}}
\end{center}

\subsection{\method}
The hyperparameters specific to our method are listed in \cref{tab:dream-mpc-hparams}.

\begin{table}[h]
\centering
\parbox{\textwidth}{
\caption{\textbf{\method planning hyperparameters.} We use the same set of hyperparameters for all tasks. All other hyperparameters are the default TD-MPC2 and BMPC values respectively.}
\label{tab:dream-mpc-hparams}
\centering
\begin{tabular}{@{}ll@{}}
\toprule
\textbf{Hyperparameter}         & \textbf{Value} \\ \midrule
Iterations $I$                                  & $1$                   \\
Policy prior samples $N$                        & $5$                   \\
Optimization step size $\alpha$                 & $0.1$                  \\
Action reuse coefficient $\rho$                 & $0.1$                  \\ 
Uncertainty regularization coefficient $\lambda_{\text{unc}}$  & $0.01$                  \\ \bottomrule
\end{tabular}%
}
\end{table}

\section{Additional Results}
In this section, we provide results for \method when varying the computational budget, the learning curves for all baselines as well as detailed evaluation results for all environments.

\subsection{Learning Curves}
\begin{figure}[h!]
  \centering
  \includegraphics[width=0.75\textwidth, keepaspectratio]{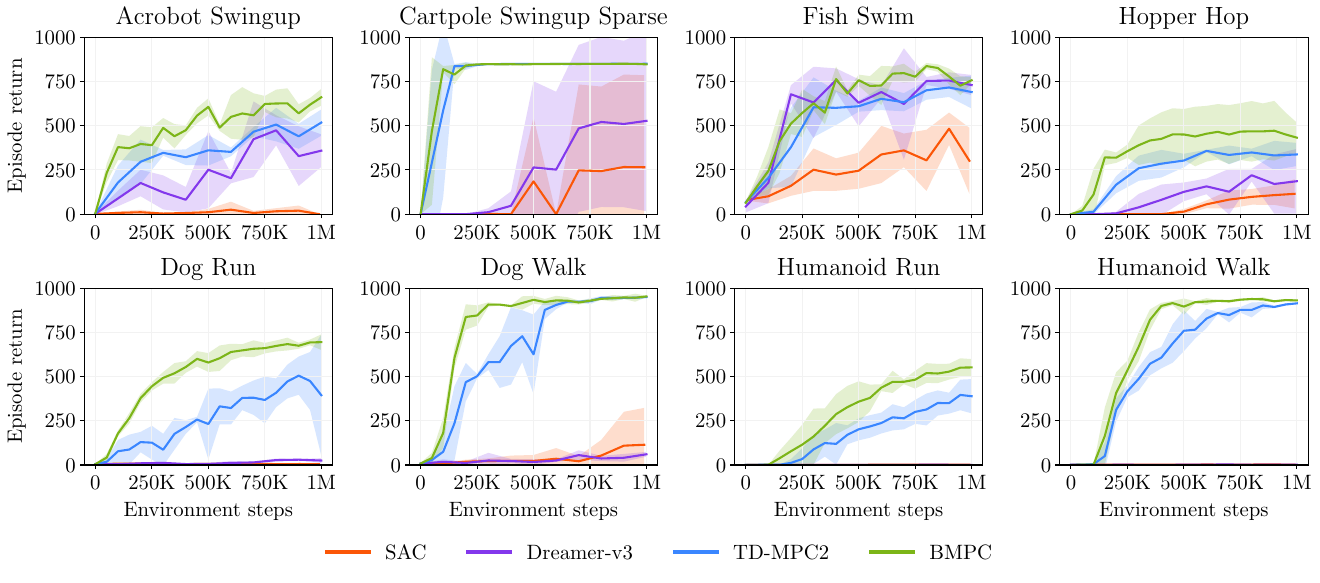}
  \caption{\textbf{Learning curves for the DeepMind Control Suite.} The line represents the mean episodic return and the shaded area the 95\% confidence interval across 3 seeds.}
  \label{fig:learning-curves-dmc}
\end{figure}

\begin{figure}[h!]
  \centering
  \includegraphics[width=0.75\textwidth, keepaspectratio]{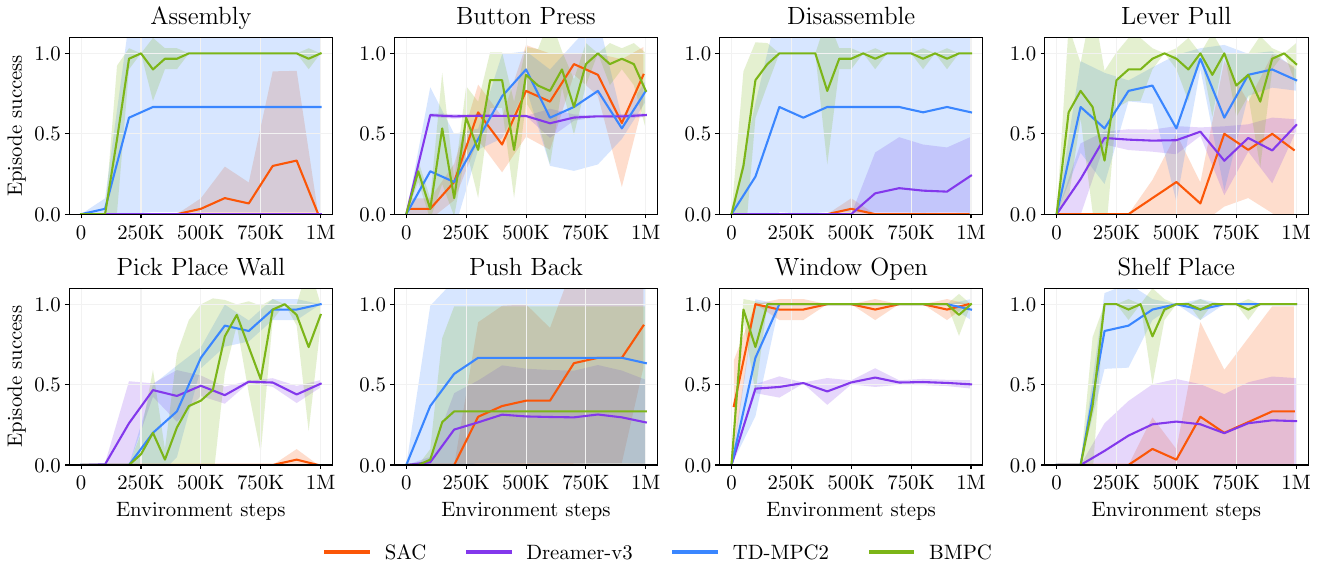}
  \caption{\textbf{Learning curves for Meta-World.} The line represents the mean episodic return and the shaded area the 95\% confidence interval across 3 seeds.}
  \label{fig:learning-curves-mw}
\end{figure}

\begin{figure}[h!]
  \centering
  \includegraphics[width=0.75\textwidth, keepaspectratio]{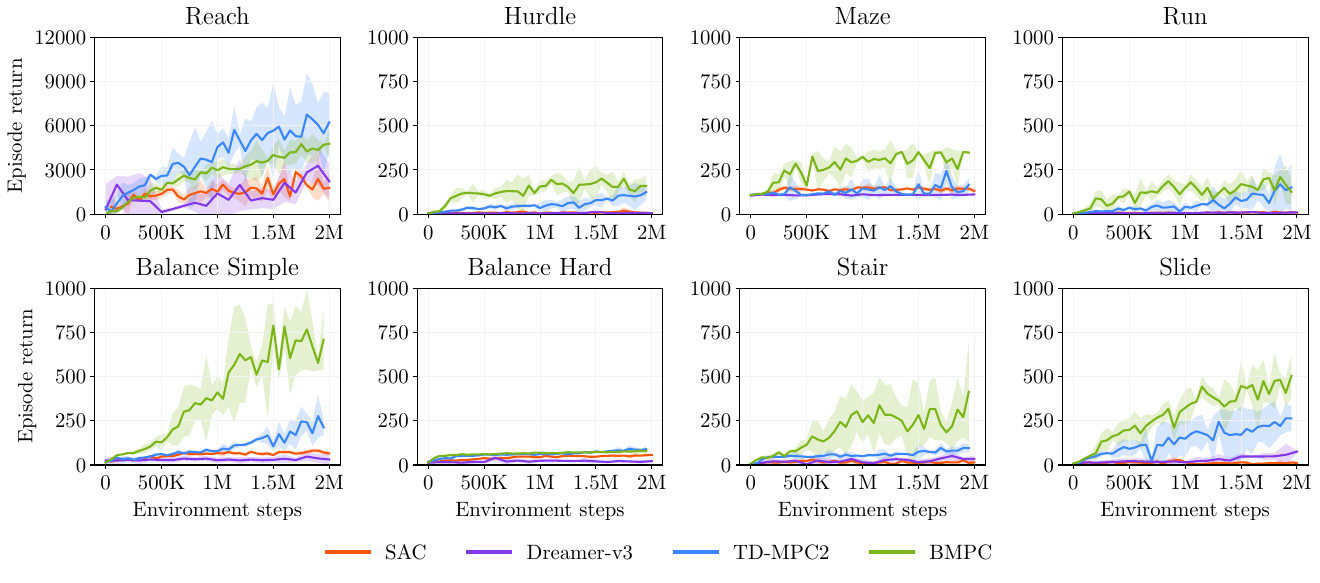}
  \caption{\textbf{Learning curves for HumanoidBench.} The line represents the mean episodic return and the shaded area the 95\% confidence interval across 3 seeds.}
  \label{fig:learning-curves-humanoidbench}
\end{figure}

\cref{fig:learning-curves-dmc,fig:learning-curves-mw,fig:learning-curves-humanoidbench} show the episode returns and the success rates as a function of environment steps, respectively.
\vspace{0.4in}
\begin{center}
    \textcolor{gray}{\textbf{\textemdash~Appendices continue on next page~\textemdash}}
\end{center}

\clearpage

\subsection{Detailed Evaluation Results}\label{sec:detailed-evaluation-results}
We provide detailed results for all environments in \cref{tab:eval-results-dmc,tab:eval-results-humanoidbench,tab:eval-results-mw}. We find that having a good policy is important because it leads to better value estimates, which are crucial for gradient-based MPC. While \method can improve the performance of the policy for TD-MPC2, it cannot consistently match the performance of MPPI. Since the performance of the policy is quite weak as shown in \cref{tab:tdmpc2-eval-results-dmc,tab:tdmpc2-eval-results-mw,tab:tdmpc2-eval-results-humanoidbench}, this fact favours MPPI, which has a higher diversity of the initial solutions due to the sampling procedure. While we can further improve the performance of \method with TD-MPC2 as a basis, for example by increasing the number of optimization iterations, this also increases computational costs. This highlights the importance of a good initial solution to warm-start the MPC optimization process, especially for high-dimensional problems.
\begin{scriptsize}
    \begin{table}[h!]
      \centering
    \begin{threeparttable}
    \setlength\tabcolsep{1.5pt}
      \caption{\textbf{DeepMind Control Suite evaluation results of different algorithms.}}
      \scriptsize %
      \renewcommand{\arraystretch}{1.15}
      \begin{tabular}{lN{1.75cm}N{1.75cm}N{1.75cm}N{1.75cm}N{1.75cm}N{1.75cm}}
        \toprule
        \textbf{Task} & \textbf{SAC} & \textbf{Dreamer-v3} & \textbf{TD-MPC2} & \textbf{BMPC} & \textbf{\method (TD-MPC2)} & \textbf{\method (BMPC)} \\
        \midrule
        Acrobot Swingup         & 176 $\pm$ 21  & 372 $\pm$ 141 & \underline{595 $\pm$ 34} & 587 $\pm$ 25             & 590 $\pm$ 40                                            & \textbf{596 $\pm$ 50}                                 \\
        Cartpole Swingup Sparse & 788 $\pm$ 10  & 538 $\pm$ 325 & \underline{848 $\pm$ 0}  & 837 $\pm$ 14             & 847 $\pm$ 3                                             & \textbf{849 $\pm$ 1}                                  \\
        Fish Swim               & 657 $\pm$ 110 & 729 $\pm$ 98  & 786 $\pm$ 8              & \underline{804 $\pm$ 17} & 764 $\pm$ 56                                            & \textbf{816 $\pm$ 11}                                 \\
        Hopper Hop              & 287 $\pm$ 15  & 198 $\pm$ 111 & \textbf{493 $\pm$ 47}    & 404 $\pm$ 39             & 307 $\pm$ 38                                            & \underline{423 $\pm$ 54}                              \\
        Dog Run                 & 15 $\pm$ 6    & 26 $\pm$ 7    & 358 $\pm$ 228            & \underline{678 $\pm$ 27} & 115 $\pm$ 72                                            & \textbf{703 $\pm$ 19}                                 \\
        Dog Walk                & 42 $\pm$ 33   & 47 $\pm$ 20   & 933 $\pm$ 10             & \underline{937 $\pm$ 4}  & 389 $\pm$ 22                                            & \textbf{946 $\pm$ 7}                                  \\
        Humanoid Run            & 83 $\pm$ 43   & 1 $\pm$ 1     & 344 $\pm$ 60             & \underline{528 $\pm$ 29} & 110 $\pm$ 10                                            & \textbf{531 $\pm$ 38}                                 \\
        Humanoid Walk           & 364 $\pm$ 95  & 2 $\pm$ 1     & 899 $\pm$ 10             & \underline{917 $\pm$ 6}  & 338 $\pm$ 63                                            & \textbf{937 $\pm$ 4}                                  \\
        \midrule
        \textbf{Mean}           & 302 $\pm$ 269 & 239 $\pm$ 261 & 657 $\pm$ 225            & \underline{711 $\pm$ 181}            & 433 $\pm$ 259                                           & \textbf{725 $\pm$ 181}                                         \\
        \bottomrule
      \end{tabular}
      \label{tab:eval-results-dmc}  
      \footnotesize{The results are the mean episode returns and standard deviations for three random seeds and ten test episodes. \textbf{Best} and \underline{second best} results are highlighted.}
      \end{threeparttable}
    \end{table}
\end{scriptsize}

\begin{scriptsize}
    \begin{table}[h!]
      \centering
    \begin{threeparttable}
    \setlength\tabcolsep{1.5pt}
      \caption{\textbf{Meta-World evaluation results of different algorithms.}}
      \scriptsize %
      \renewcommand{\arraystretch}{1.15}
      \begin{tabular}{lN{1.75cm}N{1.75cm}N{1.75cm}N{1.75cm}N{1.75cm}N{1.75cm}}
        \toprule
        \textbf{Task} & \textbf{SAC} & \textbf{Dreamer-v3} & \textbf{TD-MPC2} & \textbf{BMPC} & \textbf{\method (TD-MPC2)} & \textbf{\method (BMPC)} \\
        \midrule
        Assembly        & 0.0 $\pm$ 0.0   & 0.0 $\pm$ 0.0               & 1.0 $\pm$ 0.0               & 1.0 $\pm$ 0.0               & \underline{1.0 $\pm$ 0.0}                               & \textbf{1.0 $\pm$ 0.0}                                \\
        Button Press    & 0.27 $\pm$ 0.31 & \underline{0.61 $\pm$ 0.02} & 0.33 $\pm$ 0.47             & 0.33 $\pm$ 0.47             & 0.33 $\pm$ 0.47                                         & \textbf{0.67 $\pm$ 0.47}                              \\
        Disassemble     & 0.03 $\pm$ 0.05 & 0.27 $\pm$ 0.23             & 0.67 $\pm$ 0.47             & \underline{1.0 $\pm$ 0.0}   & 0.67 $\pm$ 0.47                                         & \textbf{1.0 $\pm$ 0.0}                                \\
        Lever Pull      & 0.03 $\pm$ 0.05 & 0.52 $\pm$ 0.1              & 0.0 $\pm$ 0.0               & \underline{0.67 $\pm$ 0.47} & 0.0 $\pm$ 0.0                                           & \textbf{0.67 $\pm$ 0.47}                              \\
        Pick Place Wall & 0.0 $\pm$ 0.0   & 0.21 $\pm$ 0.24             & \textbf{1.0 $\pm$ 0.0}      & 0.0 $\pm$ 0.0               & 0.67 $\pm$ 0.47                                         & \underline{0.67 $\pm$ 0.47}                           \\
        Push Back       & 0.67 $\pm$ 0.47 & 0.32 $\pm$ 0.23             & \underline{0.67 $\pm$ 0.47} & 0.33 $\pm$ 0.47             & \textbf{0.67 $\pm$ 0.47}                                & 0.33 $\pm$ 0.47                                       \\
        Shelf Place     & 0.0 $\pm$ 0.0   & 0.27 $\pm$ 0.21             & 0.67 $\pm$ 0.47             & 0.67 $\pm$ 0.47             & \underline{1.0 $\pm$ 0.0}                               & \textbf{1.0 $\pm$ 0.0}                                \\
        Window Open     & 1.0 $\pm$ 0.0   & 0.48 $\pm$ 0.09             & \underline{1.0 $\pm$ 0.0}   & 0.67 $\pm$ 0.47             & 0.67 $\pm$ 0.47                                         & \textbf{1.0 $\pm$ 0.0}                                \\
        \midrule
        \textbf{Mean}   & 0.25 $\pm$ 0.36 & 0.33 $\pm$ 0.18             & \underline{0.67 $\pm$ 0.33}             & 0.58 $\pm$ 0.32             & 0.62 $\pm$ 0.31                                         & \textbf{0.79 $\pm$ 0.23    }                                   \\
        \bottomrule
      \end{tabular}
      \label{tab:eval-results-mw}  
      \footnotesize{The results are the mean episode successes and standard deviations for three random seeds and ten test episodes. \textbf{Best} and \underline{second best} results are highlighted.}
      \end{threeparttable}
    \end{table}
\end{scriptsize}

\begin{scriptsize}
    \begin{table}[h!]
      \centering
    \begin{threeparttable}
    \setlength\tabcolsep{1.5pt}
      \caption{\textbf{HumanoidBench evaluation results of different algorithms.}}
      \scriptsize %
      \renewcommand{\arraystretch}{1.15}
      \begin{tabular}{lN{1.75cm}N{1.75cm}N{1.75cm}N{1.75cm}N{1.75cm}N{1.75cm}}
        \toprule
        \textbf{Task} & \textbf{SAC} & \textbf{Dreamer-v3} & \textbf{TD-MPC2} & \textbf{BMPC} & \textbf{\method (TD-MPC2)} & \textbf{\method (BMPC)} \\
        \midrule
        Balance Hard   & 55 $\pm$ 3     & 28 $\pm$ 12     & \textbf{92 $\pm$ 12}     & 81 $\pm$ 12               & 45 $\pm$ 10                                             & \underline{82 $\pm$ 12}                               \\
        Balance Simple & 70 $\pm$ 10    & 39 $\pm$ 14     & 240 $\pm$ 37             & \underline{489 $\pm$ 84}  & 47 $\pm$ 14                                             & \textbf{654 $\pm$ 89}                                 \\
        Hurdle         & 5 $\pm$ 3      & 13 $\pm$ 5      & 78 $\pm$ 24              & \underline{120 $\pm$ 43}  & 12 $\pm$ 1                                              & \textbf{249 $\pm$ 34}                                 \\
        Maze           & 140 $\pm$ 7    & 110 $\pm$ 4     & 169 $\pm$ 47             & \textbf{349 $\pm$ 2}      & 120 $\pm$ 8                                             & \underline{266 $\pm$ 33}                              \\
        Reach          & 2048 $\pm$ 212 & 2151 $\pm$ 1038 & \textbf{5037 $\pm$ 1436} & 4125 $\pm$ 324            & 2751 $\pm$ 444                                          & \underline{4348 $\pm$ 215}                            \\
        Run            & 8 $\pm$ 3      & 11 $\pm$ 5      & 136 $\pm$ 110            & \underline{139 $\pm$ 81}  & 10 $\pm$ 7                                              & \textbf{302 $\pm$ 11}                                 \\
        Slide          & 11 $\pm$ 5     & 56 $\pm$ 29     & 237 $\pm$ 54             & \underline{442 $\pm$ 36}  & 16 $\pm$ 3                                              & \textbf{632 $\pm$ 114}                                \\
        Stair          & 15 $\pm$ 15    & 35 $\pm$ 17     & 100 $\pm$ 18             & \underline{403 $\pm$ 145} & 30 $\pm$ 6                                              & \textbf{456 $\pm$ 145}                                \\
        \midrule
        \textbf{Mean}  & 294 $\pm$ 664  & 305 $\pm$ 698   & 761 $\pm$ 1617           & \underline{769 $\pm$ 1277}            & 379 $\pm$ 897                                           & \textbf{874 $\pm$ 1326}                                     \\
        \bottomrule
      \end{tabular}
      \label{tab:eval-results-humanoidbench}  
      \footnotesize{The results are the mean episode returns and standard deviations for three random seeds and ten test episodes. \textbf{Best} and \underline{second best} results are highlighted.}
      \end{threeparttable}
    \end{table}
\end{scriptsize}

\subsection{Detailed TD-MPC2 and BMPC Results}\label{sec:detailed-results}
We include full results of TD-MPC2 and BMPC for all environments in \cref{tab:tdmpc2-eval-results-dmc,tab:tdmpc2-eval-results-mw,tab:tdmpc2-eval-results-humanoidbench}, including the performance of using the underlying policy network only. We also conduct experiments in which we apply the test-time regularization defined in \cref{eq:uncertainty} with a regularization coefficient of $\lambda_{\text{unc}} = 0.01$ to TD-MPC2 and BMPC. While the regularization can improve the performance of BMPC in some cases, it causes a significant performance decrease for TD-MPC2, especially for high-dimensional problems.

\begin{scriptsize}
    \begin{table}[h!]
    \centering
    \begin{threeparttable}
    \setlength\tabcolsep{1.5pt}
      \caption{\textbf{DeepMind Control Suite evaluation results of different TD-MPC2 and BMPC variants.}}
      \scriptsize %
      \renewcommand{\arraystretch}{1.15}
      \begin{tabular}{lN{1.5cm}N{1.5cm}N{2.5cm}N{1.5cm}N{1.5cm}N{2.5cm}}
        \toprule
          \textbf{Environment}             & \textbf{TD-MPC2}                  & \textbf{TD-MPC2 (policy only)}   & \textbf{TD-MPC2 (w/ test-time regularization)} & \textbf{BMPC}      & \textbf{BMPC (policy only)}   & \textbf{BMPC (w/ test-time regularization)}   \\
          \hline
          Acrobot Swingup         & \textbf{595 $\pm$ 34}   & 551 $\pm$ 21         & \underline{594 $\pm$ 32} & 587 $\pm$ 25             & 564 $\pm$ 52            & 573 $\pm$ 11                       \\
          Cartpole Swingup Sparse & \underline{848 $\pm$ 0} & 760 $\pm$ 114        & 848 $\pm$ 0              & 837 $\pm$ 14             & \textbf{848 $\pm$ 1}    & 845 $\pm$ 3                        \\
          Fish Swim               & 786 $\pm$ 8             & 645 $\pm$ 83         & 783 $\pm$ 13             & \underline{804 $\pm$ 17} & \textbf{804 $\pm$ 14}   & 776 $\pm$ 9                        \\
          Hopper Hop              & \textbf{493 $\pm$ 47}   & 383 $\pm$ 154        & \underline{465 $\pm$ 79} & 404 $\pm$ 39             & 445 $\pm$ 106           & 440 $\pm$ 87                       \\
          Dog Run                 & 358 $\pm$ 228           & 89 $\pm$ 52          & 376 $\pm$ 231            & \underline{678 $\pm$ 27} & 670 $\pm$ 13            & \textbf{678 $\pm$ 23}              \\
          Dog Walk                & 933 $\pm$ 10            & 298 $\pm$ 20         & 926 $\pm$ 9              & \underline{937 $\pm$ 4}  & 930 $\pm$ 5             & \textbf{940 $\pm$ 4}               \\
          Humanoid Run            & 344 $\pm$ 60            & 65 $\pm$ 2           & 345 $\pm$ 55             & \textbf{528 $\pm$ 29}    & 458 $\pm$ 15            & \underline{514 $\pm$ 31}           \\
          Humanoid Walk           & 899 $\pm$ 10            & 142 $\pm$ 36         & 881 $\pm$ 9              & 917 $\pm$ 6              & \underline{930 $\pm$ 7} & \textbf{931 $\pm$ 3}               \\
          \midrule
          \textbf{Mean}           & 657 $\pm$ 225           & 367 $\pm$ 247        & 652 $\pm$ 221            & \underline{711 $\pm$ 181}            & 706 $\pm$ 187           & \textbf{712 $\pm$ 179}                      \\
          \bottomrule
      \end{tabular}
      \label{tab:tdmpc2-eval-results-dmc}  
      \footnotesize{The results are the mean episode returns and standard deviations for three random seeds and ten test episodes. \textbf{Best} and \underline{second best} results are highlighted.}
    \end{threeparttable}
    \end{table}
\end{scriptsize}

\begin{scriptsize}
    \begin{table}[h!]
    \centering
    \begin{threeparttable}
    \setlength\tabcolsep{1.5pt}
      \caption{\textbf{Meta-World evaluation results of different TD-MPC2 and BMPC variants.}}
      \scriptsize %
      \renewcommand{\arraystretch}{1.15}
      \begin{tabular}{lN{1.5cm}N{1.5cm}N{2.5cm}N{1.5cm}N{1.5cm}N{2.5cm}}
        \toprule
          \textbf{Environment}             & \textbf{TD-MPC2}                  & \textbf{TD-MPC2 (policy only)}   & \textbf{TD-MPC2 (w/ test-time regularization)} & \textbf{BMPC}      & \textbf{BMPC (policy only)}   & \textbf{BMPC (w/ test-time regularization)}   \\
          \hline
            Assembly        & 1.0 $\pm$ 0.0               & 1.0 $\pm$ 0.0        & 0.67 $\pm$ 0.47             & 1.0 $\pm$ 0.0             & \underline{1.0 $\pm$ 0.0}   & \textbf{1.0 $\pm$ 0.0}             \\
            Button Press    & 0.33 $\pm$ 0.47             & 0.0 $\pm$ 0.0        & \underline{0.67 $\pm$ 0.47} & 0.33 $\pm$ 0.47           & \textbf{1.0 $\pm$ 0.0}      & 0.33 $\pm$ 0.47                    \\
            Disassemble     & 0.67 $\pm$ 0.47             & 0.67 $\pm$ 0.47      & 0.67 $\pm$ 0.47             & \underline{1.0 $\pm$ 0.0} & 0.67 $\pm$ 0.47             & \textbf{1.0 $\pm$ 0.0}             \\
            Lever Pull      & 0.0 $\pm$ 0.0               & 0.0 $\pm$ 0.0        & 0.0 $\pm$ 0.0               & 0.67 $\pm$ 0.47           & \textbf{1.0 $\pm$ 0.0}      & \underline{0.67 $\pm$ 0.47}        \\
            Pick Place Wall & \textbf{1.0 $\pm$ 0.0}      & 0.0 $\pm$ 0.0        & 0.33 $\pm$ 0.47             & 0.0 $\pm$ 0.0             & \underline{0.67 $\pm$ 0.47} & 0.33 $\pm$ 0.47                    \\
            Push Back       & \underline{0.67 $\pm$ 0.47} & 0.33 $\pm$ 0.47      & \textbf{0.67 $\pm$ 0.47}    & 0.33 $\pm$ 0.47           & 0.33 $\pm$ 0.47             & 0.33 $\pm$ 0.47                    \\
            Shelf Place     & 0.67 $\pm$ 0.47             & 0.67 $\pm$ 0.47      & 1.0 $\pm$ 0.0               & 0.67 $\pm$ 0.47           & \underline{1.0 $\pm$ 0.0}   & \textbf{1.0 $\pm$ 0.0}             \\
            Window Open     & 1.0 $\pm$ 0.0               & 0.33 $\pm$ 0.47      & \underline{1.0 $\pm$ 0.0}   & 0.67 $\pm$ 0.47           & \textbf{1.0 $\pm$ 0.0}      & 0.67 $\pm$ 0.47                    \\
            \midrule
            \textbf{Mean}   & \underline{0.67 $\pm$ 0.33}             & 0.38 $\pm$ 0.35      & 0.62 $\pm$ 0.31             & 0.58 $\pm$ 0.32           & \textbf{0.83 $\pm$ 0.24}             & 0.67 $\pm$ 0.29                    \\
            \bottomrule
      \end{tabular}
      \label{tab:tdmpc2-eval-results-mw}  
      \footnotesize{The results are the mean episode returns and standard deviations for three random seeds and ten test episodes. \textbf{Best} and \underline{second best} results are highlighted.}
    \end{threeparttable}
    \end{table}
\end{scriptsize}

\begin{scriptsize}
    \begin{table}[h!]
    \centering
    \begin{threeparttable}
    \setlength\tabcolsep{1.5pt}
      \caption{\textbf{HumanoidBench evaluation results of different TD-MPC2 and BMPC variants.}}
      \scriptsize %
      \renewcommand{\arraystretch}{1.15}
      \begin{tabular}{lN{1.5cm}N{1.5cm}N{2.5cm}N{1.5cm}N{1.5cm}N{2.5cm}}
        \toprule
          \textbf{Environment}             & \textbf{TD-MPC2}                  & \textbf{TD-MPC2 (policy only)}   & \textbf{TD-MPC2 (w/ test-time regularization)} & \textbf{BMPC}      & \textbf{BMPC (policy only)}   & \textbf{BMPC (w/ test-time regularization)}   \\
          \hline
            Balance Hard   & \underline{92 $\pm$ 12}  & 34 $\pm$ 3           & \textbf{94 $\pm$ 22} & 81 $\pm$ 12                & 78 $\pm$ 8               & 80 $\pm$ 9                         \\
            Balance Simple & 240 $\pm$ 37             & 33 $\pm$ 16          & 208 $\pm$ 34         & \underline{489 $\pm$ 84}   & 414 $\pm$ 45             & \textbf{778 $\pm$ 77}              \\
            Hurdle         & 78 $\pm$ 24              & 14 $\pm$ 3           & 73 $\pm$ 27          & 120 $\pm$ 43               & \underline{147 $\pm$ 40} & \textbf{175 $\pm$ 51}              \\
            Maze           & 169 $\pm$ 47             & 111 $\pm$ 3          & 115 $\pm$ 4          & \textbf{349 $\pm$ 2}       & 121 $\pm$ 7              & \underline{347 $\pm$ 4}            \\
            Reach          & \textbf{5037 $\pm$ 1436} & 1558 $\pm$ 368       & 399 $\pm$ 208        & \underline{4125 $\pm$ 324} & 2117 $\pm$ 309           & 2279 $\pm$ 376                     \\
            Run            & 136 $\pm$ 110            & 8 $\pm$ 4            & 99 $\pm$ 72          & \underline{139 $\pm$ 81}   & 91 $\pm$ 25              & \textbf{222 $\pm$ 56}              \\
            Slide          & 237 $\pm$ 54             & 14 $\pm$ 2           & 248 $\pm$ 77         & \underline{442 $\pm$ 36}   & 250 $\pm$ 26             & \textbf{553 $\pm$ 100}             \\
            Stair          & 100 $\pm$ 18             & 24 $\pm$ 8           & 91 $\pm$ 23          & \underline{403 $\pm$ 145}  & 208 $\pm$ 46             & \textbf{432 $\pm$ 199}             \\
            \midrule
            \textbf{Mean}  & \underline{761 $\pm$ 1617}           & 224 $\pm$ 505        & 166 $\pm$ 106        & \textbf{769 $\pm$ 1277}             & 428 $\pm$ 646            & 608 $\pm$ 665                      \\
            \bottomrule
      \end{tabular}
      \label{tab:tdmpc2-eval-results-humanoidbench}  
      \footnotesize{The results are the mean episode returns and standard deviations for three random seeds and ten test episodes. \textbf{Best} and \underline{second best} results are highlighted.}
    \end{threeparttable}
    \end{table}
\end{scriptsize}

\vspace{0.4in}
\begin{center}
    \textcolor{gray}{\textbf{\textemdash~Appendices continue on next page~\textemdash}}
\end{center}

\clearpage
\subsection{Planner Sweep}
\begin{figure}[h!]
    \centering
    \includegraphics[width=0.5\linewidth]{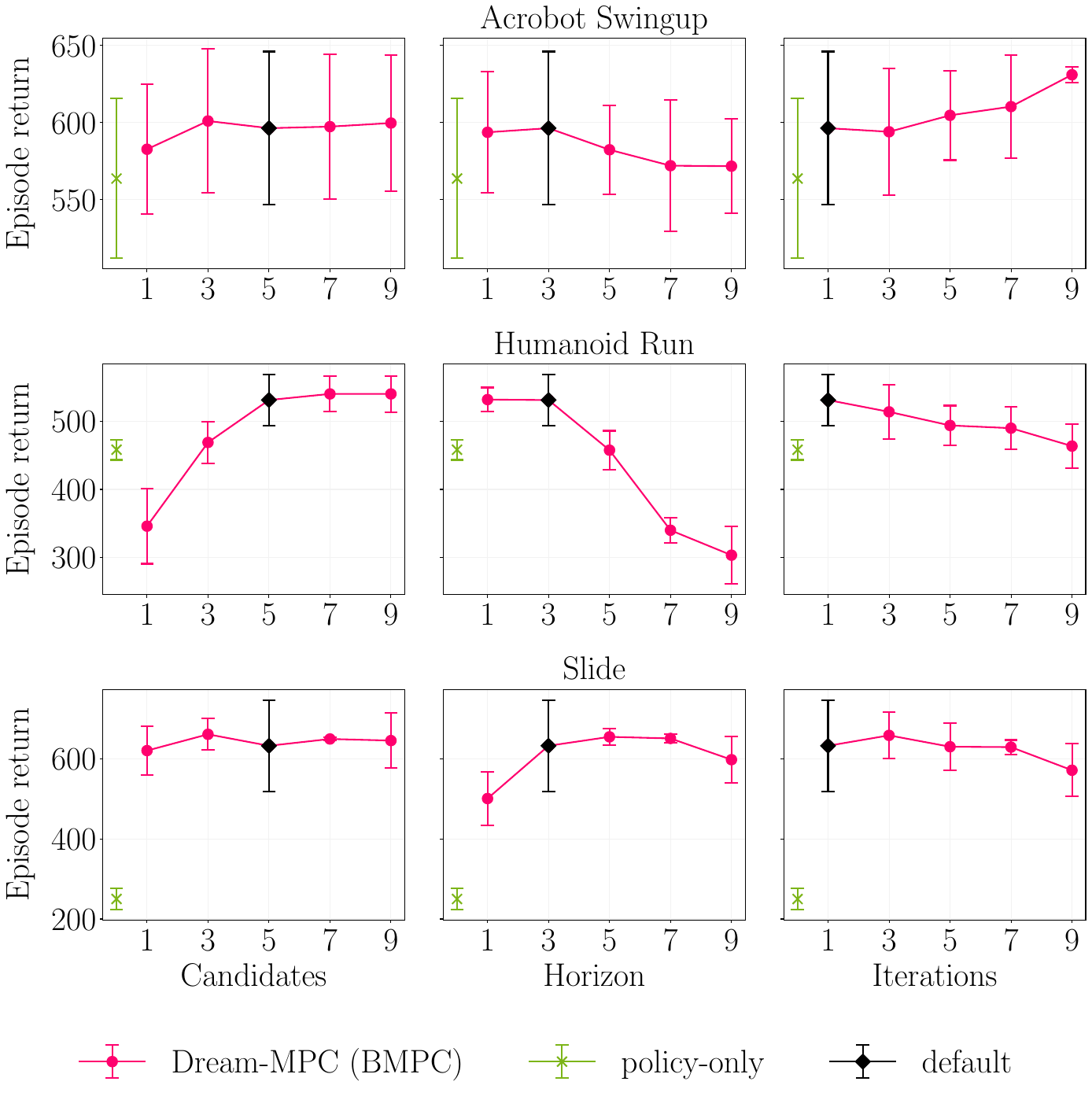}
    \caption{
        \textbf{Parameter sweep.} Performance of trained BMPC agents with \method at test time when varying the number of candidates, horizon and number of optimization iterations. When varying one hyperparameter, the others are fixed to their default value. We also include the performance of the learned policy $\pi_{\theta}$ and the default values of one iteration, a horizon of three and five candidate trajectories.
    }
    \label{fig:planning-params}
\end{figure}

\cref{fig:planning-params} shows the performance of trained BMPC agents with \method at test time when varying the number of candidates, horizon and number of optimization iterations. When varying one hyperparameter, the others are fixed to their default value.

\section{Integration into Dreamer}\label{sec:dreamer}
We further integrate our base method (without uncertainty regularization) into Dreamer \citep{hafner_dream_2020} to show that it also works with other model-based RL algorithms. Dreamer learns a latent dynamics model, often referred to as a world model, consisting of the following components:
\begin{itemize}
    \item Representation model: $p_{\theta}(s_t|s_{t-1},a_{t-1},o_t)$
    \item Transition model: $q_{\theta}(s_t|s_{t-1},a_{t-1})$
    \item Reward model: $q_{\theta}(r_t|s_t)$
    \item Observation model (only used as an additional learning signal): $q_{\theta}(o_t|s_t)$
\end{itemize}
All components are jointly optimized to increase the variational lower bound (ELBO), including reconstruction terms for observations and rewards as well as a KL regularizer:
\begin{equation}\label{eq:reconstruction-objective}
    \mathcal{L}_{\text{Rec}} = \mathbb{E} \left[\sum_{t} (\mathcal{L}_{O}^t + \mathcal{L}_{R}^t + \mathcal{L}_ {D}^t) \right] + \text{const},
\end{equation}
where
\begin{equation}
    \begin{aligned}
        \mathcal{L}_{O}^t &= \ln q(o_t|s_t), \\ \mathcal{L}_{R}^t &= \ln q(r_t|s_t),\\ \mathcal{L}_{D}^t &= - \beta \text{KL}(p(s_t|s_{t-1},a_{t-1},o_t) || q(s_t|s_{t-1},a_{t-1})).
    \end{aligned}
\end{equation}
The expected values are calculated based on the dataset and representation model. Please refer to \citet{hafner_dream_2020} for the derivation of the variational bound.

Following the original Dreamer implementation, we estimate state values using $V_{\lambda}$, an exponentially-weighted average of the reward estimates for a different number of steps beyond the horizon with the learned value model to balance bias and variance:
\begin{equation}
    V_R(s_{\tau}) = \mathbb{E}_{q_{\theta}, \pi_{\phi}}\left[\sum_{n=\tau}^{t+H} r_n\right],
\end{equation}
\begin{equation}
    V_N^k(s_{\tau}) = \mathbb{E}_{q_{\theta}, \pi_{\phi}}\left[\sum_{n=\tau}^{h-1} \gamma^{n-\tau} r_n + \gamma^{h-\tau} v_{\psi}(s_h)\right] \quad \text{with $h = \min(\tau + k, t + H)$},
\end{equation}
\begin{equation}\label{eq:vlambda}
    V_{\lambda}(s_{\tau}) = (1-\lambda) \sum_{n=1}^{H-1} \lambda^{n-1} V_N^n(s_{\tau}) + \lambda^{H-1} V_N^H(s_{\tau}).
\end{equation}

For each time step $t$, \method creates an initial sequence of actions by performing an imaginary rollout of the policy $\pi_{\phi}$ and generates $N$ candidate trajectories adding small perturbations to the initial action sequence: 
\begin{equation}\label{eq:trajectories-dreamer}
    \{\hat{a}^{(n)}\}_{n=1}^{N} = \{\pi_{\phi} (a_{\tau-1} | s_{\tau-1}) + \epsilon^{(n)}_{\tau} | \tau = t+1, ..., t+H+1\}_{n=1}^{N}, \quad \text{where}\ \epsilon^{(n)}_{\tau} \sim \mathcal{N}(0, \sigma_a^2).
\end{equation}
The imaginary rollout is done by encoding observations and actions into latent space using the representation model $p_{\theta}$ and repeatedly calling the one-step transition model $q_{\theta}$ to generate a sequence of predicted states $\{s_{\tau}\}_{\tau=t+1}^{t+H+1}$ for each candidate trajectory.
\begin{equation}\label{eq:rollout-states}
    s^{(n)}_{t} \sim p_{\theta}(s^{(n)}_t|s^{(n)}_{t-1},a^{(n)}_{t-1},o_t), \quad\quad s^{(n)}_{t+1:t+H+1} \sim \prod_{\tau=t+1}^{t+H+1} q_{\theta}(s^{(n)}_\tau|s^{(n)}_{\tau-1},a^{(n)}_{\tau-1})
\end{equation}
We integrate our gradient-based MPC method into Dreamer as shown in \cref{algo:agent}.
\begin{algorithm}[H]
    \small
    \begin{algorithmic}
    \STATE {\bfseries Input:} Representation model $p_{\theta}(s_t|s_{t-1},a_{t-1},o_t)$, transition model $q_{\theta}(s_t|s_{t-1},a_{t-1})$, reward model $q_{\theta}(r_t|s_t)$, value function model $v_{\psi}(s_t)$, policy model $\pi_{\phi}(a_t|s_t)$, exploration noise $p(\epsilon)$, action repeat $R$, seed episodes $S$, collect interval $C$, batch size $B$, chunk length $L$, learning rate $\eta$
    \newline
    \STATE Initialize dataset $\mathcal{D}$ with $S$ random seed episodes.
    \STATE Initialize model parameters $\theta, \phi, \psi$ randomly.
    \WHILE{not converged}
      \FOR{update step $s=1..C$}
        \STATE Draw sequences $\{(o_t,a_t,r_t)_{t=k}^{L+k}\}_{i=1}^B\sim\mathcal{D}$ uniformly at random from the dataset.
        \STATE Compute loss $\mathcal{L}(\theta)$ from \cref{eq:reconstruction-objective}.
        \STATE Update model parameters $\theta\leftarrow\theta-\eta\nabla_\theta\mathcal{L}(\theta)$.
        \STATE Imagine trajectories $\{(s_{\tau}, a_{\tau})\}_{\tau=t}^{t+H}$ from each $s_t$. 
        \STATE Predict rewards $\mathbb{E}\left[q_{\theta}(r_{\tau}|s_{\tau})\right]$ and values $v_{\psi}(s_{\tau})$.
        \STATE Compute value estimates $V_{\lambda}(s_{\tau})$ via \cref{eq:vlambda}.
        \STATE Update $\phi \leftarrow \phi + \eta \nabla_{\phi} \sum_{\tau=t}^{t+H} V_{\lambda}(s_{\tau})$.
        \STATE Update $\psi \leftarrow \psi - \eta \nabla_{\psi} \sum_{\tau=t}^{t+H} \frac{1}{2} ||v_{\psi}(s_{\tau}) - V_{\lambda}(s_{\tau})||^2$.
      \ENDFOR
      \newline
      \STATE $o_1\leftarrow\texttt{env.reset()}$
      \FOR{time step $t=1..[{\frac{T}{R}}]$}
        \STATE Infer current state $s_t \sim p_{\theta}(s_t|s_{t-1},a_{t-1},o_t)$ from the history.
        \STATE $a_t\leftarrow\texttt{planner(}s_t\texttt{)}$, see \cref{algo:dream-mpc-dreamer} for details.
        \STATE Add exploration noise $\epsilon\sim p(\epsilon)$ to the action. 
        \FOR{action repeat $k=1..R$}
          \STATE $r_t^k,o_{t+1}^k\leftarrow\texttt{env.step(}a_t\texttt{)}$
        \ENDFOR
        \STATE $r_t,o_{t+1} \leftarrow \sum_{k=1}^R r_t^k, o_{t+1}^R$ 
      \ENDFOR
      \STATE $\mathcal{D}\leftarrow\mathcal{D}\cup\{(o_t,a_t,r_t)_{t=1}^T\}$ 
      \newline
    \ENDWHILE
    \end{algorithmic}
    \caption{{\bf \small \method integration into Dreamer}}
    \label{algo:agent}
\end{algorithm}

{\scriptsize
\vspace*{-2ex}
\begin{algorithm}[!htp]
    \small
    \caption{{\bf \small \method planner for Dreamer}
    \label{algo:dream-mpc-dreamer}}
    \begin{algorithmic}
        \STATE {\bfseries Input:} Representation model $p_{\theta}(s_t | s_{t-1}, a_{t-1}, o_t)$, transition model $q_{\theta}(s_t|s_{t-1},a_{t-1})$, reward model $q_{\theta}(r_t|s_t)$, value function model $v_{\psi}(s_t)$, policy model $\pi_{\phi}(a_t|s_t)$, planning horizon $H$, optimization iterations $I$, candidates per iteration $J$, action noise $\sigma_a^2$, action optimization rate $\alpha$
        \newline
        \STATE Initialize proposal by rolling out the policy $\pi_{\phi}$ with the transition model $\hat{a}_{t:t+H} \sim \pi_{\phi}(s_{t:t+H})$.
        \STATE Generate $N$ candidates by adding noise $\mathcal{N}(0,\sigma_a^2)$ to the proposal via \cref{eq:trajectories-dreamer}.
        \STATE Initialize candidate action sequences $a_{t:t+H}$ via \cref{eq:action-reuse}. 
        \FOR{optimization iteration $i=1, 2, \dots I$}
            \FOR{candidate action sequence $n=1, 2, \dots N$}
                \STATE Predict imagined states $s_{\tau}^{(n)} = s^{(n)}_{t:t+H+1}$ via \cref{eq:rollout-states} 
                \STATE Predict rewards $\mathbb{E}\left[q_{\theta}(r_{\tau}^{(n)}|s_{\tau}^{(n)})\right]$ and values $v_{\psi}(s_{\tau}^{(n)})$ 
                \STATE Compute value estimates $V_{\lambda}(s_{\tau}^{(n)})$ via \cref{eq:vlambda}
                \STATE Optimize action sequence via $a^{(n)}_{\tau} \leftarrow \{a^{(n)}_{\tau} + \alpha \nabla_{a^{(n)}_{\tau}} V_{\lambda}^{(n)}(s_{\tau}^{(n)}) | \tau = t, ..., t+H\}$ 
            \ENDFOR
        \ENDFOR
        \STATE \textbf{return} First optimized action $a_t^{(k)}$ with $k = \argmax_{n} \{V_{\lambda}^{(n)}\}_{n=1}^N$.
    \end{algorithmic}
\end{algorithm}
}

\subsection{Experiments}\label{sec:dreamer-experiments}
We evaluate our method on four different environments from the DeepMind Control Suite and compare our method with PlaNet \citep{hafner_learning_2019}, Dreamer \citep{hafner_dream_2020}, SAC+AE \citep{yarats_2021}, a variant of the model-free Soft Actor Critic (SAC) \citep{haarnoja_soft_2018} algorithm for image-based observations and the Policy+Grad-MPC method proposed in \citep{s_v_gradient-based_2023}. 
\begin{figure}[htpb]
    \centering
    \includegraphics[width=\textwidth, keepaspectratio]{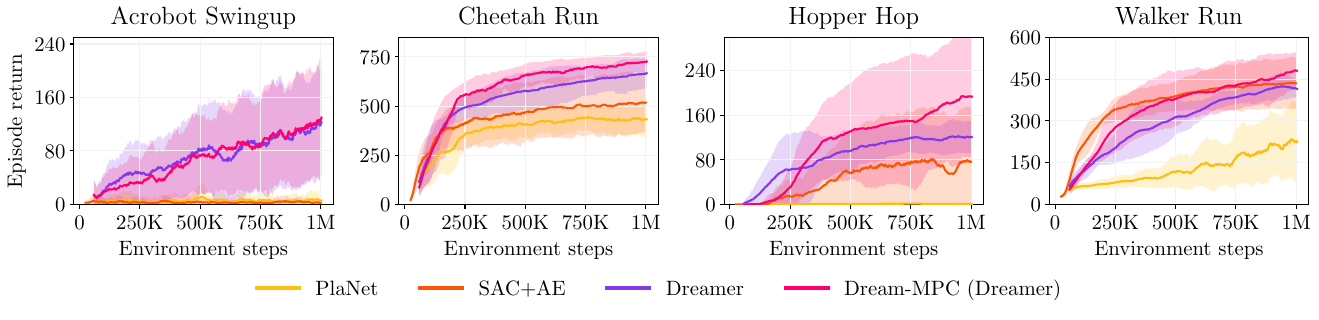}
    \caption{\textbf{Learning curves for four tasks from the DeepMind Control Suite.} The line represents the mean episodic return and the shaded area the 95\% confidence interval across 3 seeds.}
    \label{fig:learning-curves-dreamer}
\end{figure}
Note that Policy+Grad-MPC and \method both share the general idea of using a policy network to warm-start gradient-based MPC. We provide a summary of the main differences in \cref{sec:differences-grad-mpc}. All experiments are performed with only RGB visual observations with a resolution of 64 $\times$ 64.

We evaluate the performance of our method when enabling planning already during training. The learning curves are shown in \cref{fig:learning-curves-dreamer} and the evaluation results are presented in \cref{tab:eval-results-dreamer}. We find that our method can not only outperform the baselines, but also that planning during training can improve the sample efficiency without leading to premature convergence. In contrast to PlaNet (CEM) and Grad-MPC, which both use $1000 \times 10 \times 12 = \num{120000}$ evaluations of the world model at each time step, our method only requires $5 \times 1 \times 15 = 75$ evaluations. These results are not only promising since Dreamer uses a recurrent dynamics model and a relatively long planning horizon, but also in particular for Acrobot Swingup, which is a non-linear system with chaotic dynamics. All aspects usually affect gradient quality negatively, especially since first order gradient estimators can accumulate significant variance over long-horizon rollouts, which makes them in particular ineffective in chaotic systems \citep{suh_differentiable_2022}.

We benchmark inference times of the different methods on a single Nvidia GeForce RTX 4090 GPU. The results in \cref{tab:dreamer-times} show that \method is significantly faster as Grad-MPC, which uses a much higher number of candidate trajectories. While Policy+Grad-MPC is faster than \method due to using a horizon of one, the overall performance is worse compared to using the policy only because such a myopic optimization is most likely unsuitable for many problems. Note that at the moment a batched version of one operation in the recurrent world model is missing in PyTorch, which slows the parallelized gradient computation down. While this can potentially be further improved, it affects all gradient-based MPC methods in the same way, thus leading to a fair comparison.

\begin{scriptsize}
    \begin{table}[H]
    \setlength\tabcolsep{20pt}
    \centering
      \caption{\textbf{Inference times of different methods for Acrobot Swingup.} Mean and standard deviation for three random seeds and ten test episodes per seed.}
      \renewcommand{\arraystretch}{1.15}
      \begin{tabular}{lr}
        \toprule
        \textbf{Method} & \textbf{Inference time} \\
        \midrule
        PlaNet            &    31.10 $\pm$ 0.65 \unit{\milli\second} \\
        Grad-MPC          &   195.75 $\pm$ 1.33 \unit{\milli\second} \\
        Policy+Grad-MPC   &    23.16 $\pm$ 0.55 \unit{\milli\second} \\
        \method (Dreamer) &    44.86 $\pm$ 0.60 \unit{\milli\second} \\
        \bottomrule
      \end{tabular}
      \label{tab:dreamer-times}  
    \end{table}
\end{scriptsize}

\subsection{Gradient Analysis}\label{sec:gradient-analysis}
\begin{figure}[htpb]
    \centering
    \includegraphics[width=\textwidth, keepaspectratio]{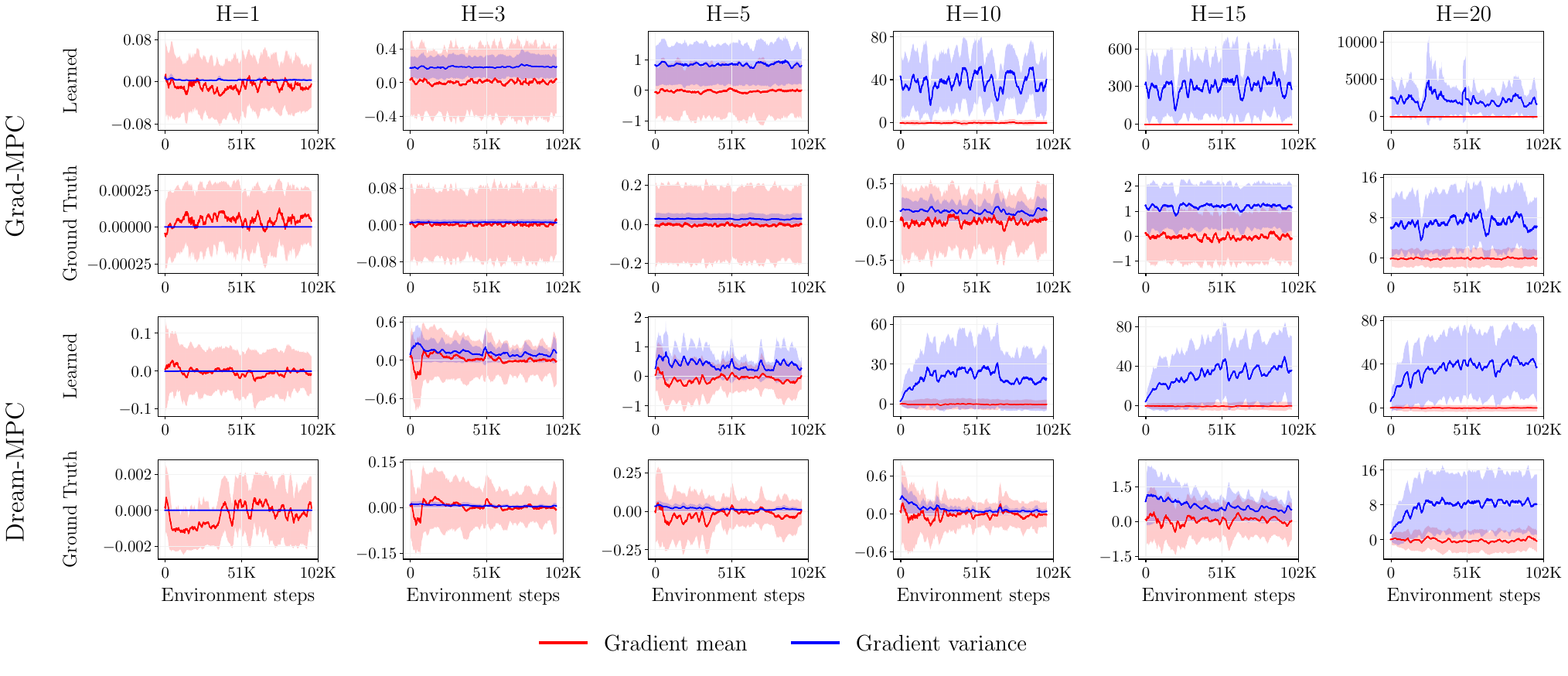}
    \caption{\textbf{Planner gradients of Grad-MPC and \method.} For different planning horizons on the Pendulum-v1 environment using the ground truth (simulator) and learned dynamics model respectively and state observations. The values are represented by their mean and standard deviation for three different random seeds. The default hyperparameters provided in \cref{tab:hyperparams-dreamer} are used unless otherwise specified.}
    \label{fig:grads}
\end{figure}

We evaluate the planner gradients of Grad-MPC and of our method for the ground truth dynamics (simulator) and the learned dynamics model for different planning horizons on the Pendulum-v1 environment with state observations. As \cref{fig:grads} shows, the magnitudes of the gradients are in reasonable orders when using the ground truth dynamics. While the variance increases for longer horizons and might also do for more complex problems, the gradients do not explode or vanish in this case. However, the variance increases significantly for longer planning horizons when using the learned dynamics model. In contrast to Grad-MPC, the variance increases much less for \method and although relatively large remains bounded, suggesting that the performance issues of gradient-based planning should not solely be attributed to issues with the gradients caused by the architecture of the world model. Our work shows that there are more aspects that need to be considered such as the quality of the initial proposal for MPC and the learned world model, advocating that further research on gradient-based planning is needed. 
\begin{figure}[htpb]
    \centering
    \includegraphics[width=\textwidth, keepaspectratio]{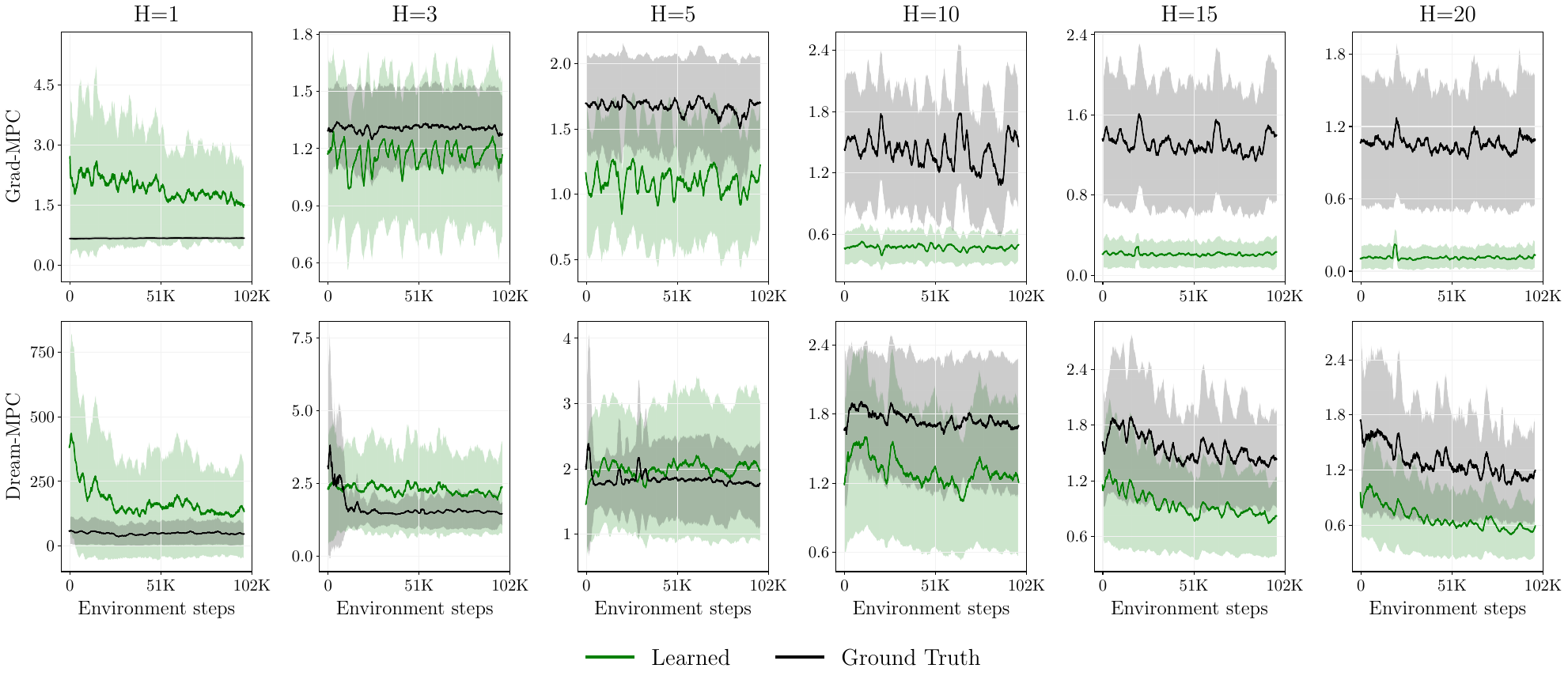}
    \caption{\textbf{Expected Signal-to-Noise Ratio (ESNR) of the planner gradients of Grad-MPC and \method.} Calculated via \cref{eq:esnr} for different planning horizons on the Pendulum-v1 environment using the ground truth (simulator) and learned dynamics model respectively and state observations. The values are represented by their mean and standard deviation for three different random seeds. The default hyperparameters provided in \cref{tab:hyperparams-dreamer} are used unless otherwise specified.}
    \label{fig:grads_esnr}
\end{figure}

As pointed out in \citet{parmas_2023}, simply evaluating the gradient quality based on variance alone is insufficient. Thus, we follow the proposal of the authors and analyze the gradients using their Expected Signal-to-Noise Ratio (ESNR), which is defined as
\begin{equation}\label{eq:esnr}
    \text{ESNR}(\nabla R) = \mathbb{E} \left[\frac{\sum \mathbb{E} [\nabla R]^2}{\sum \text{Var}[\nabla R]}\right],
\end{equation}
where $R = \sum_{\tau=t+1}^{t+H+1} r_{\tau}$ is the return, i.e., the undiscounted sum of rewards.

\cref{fig:grads_esnr} shows the ESNRs of Grad-MPC and \method using the ground truth dynamics or learned dynamics model. While the ESNR remains stable when using the ground truth dynamics, especially for longer horizons the ESNR drops when using the learned model. Recent findings \citep{georgiev_pwm_2024} suggest that learned models can improve ESNR compared to using the ground truth dynamics for some problems, indicating the possibility of further improvement. While the ESNR significantly suffers for horizons greater than ten for Grad-MPC using the learned dynamics model, the ESNR for \method remains much more stable for increasing horizons. Together with the variance which increases but does not explode, this suggests that our method is more robust compared to Grad-MPC.

\subsection{Model Exploitation}
\begin{figure}[htpb]
    \centering
    \begin{subfigure}[t]{.45\textwidth}\centering
        \centering
        \includegraphics[width=\textwidth]{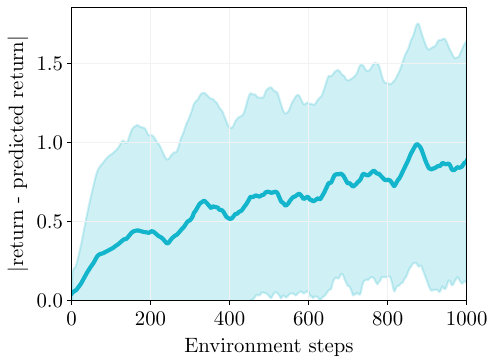}
        \caption{Mean difference between actual and predicted returns and standard deviation for three different seeds and ten test episodes per seed.}
        \label{fig:me-avg}
    \end{subfigure}\hfill%
    \begin{subfigure}[t]{.45\textwidth}\centering
        \centering
        \includegraphics[width=\textwidth]{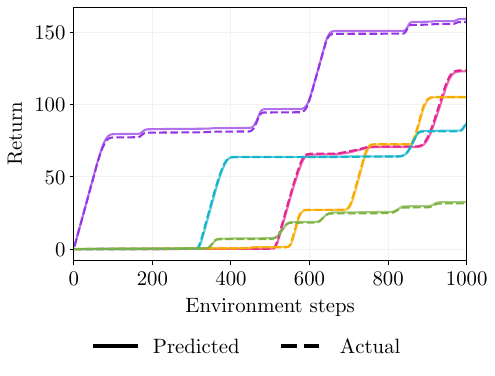}
        \caption{Actual and predicted return for five exemplary evaluation episodes.}
        \label{fig:me-subset}
    \end{subfigure}
    \caption{\textbf{Analysis of predicted returns over the number of environment steps for Acrobot Swingup.}}
    \label{fig:model-exploitation}
\end{figure}

We further analyze the problem of model exploitation, a general challenge in model-based reinforcement learning, where policies tend to exploit inaccuracies in high-capacity dynamics models, potentially leading to poor real-world performance despite high predicted returns \citep{clavera_2018}. Since our method optimizes actions to maximize expected returns, we rely on accurate predictions. \cref{fig:model-exploitation} shows the mean difference between the actual returns and the predicted returns of a trained policy on the Acrobot Swingup task in for three different seeds and ten test episodes per seed. 

We find that the differences are quite small, which indicates that the policy may not exploit the learned model. This is probably because the prediction horizon is sufficiently short and MPC may also help to compensate for model inaccuracies by replanning at each step. While the models for other environments might not necessarily be as accurate as for Acrobot Swingup, we empirically find that the learned model tends to estimate the reward quite accurately. Using an ensemble of models to consider uncertainty as for TD-MPC2 can further help to reduce model exploitation.
\vspace{0.4in}
\begin{center}
    \textcolor{gray}{\textbf{\textemdash~Appendices continue on next page~\textemdash}}
\end{center}

\clearpage

\subsection{Implementation Details}
We use PyTorch \citep{pytorch_2019} implementations of SAC+AE\footnote{\url{https://github.com/denisyarats/pytorch_sac_ae}}, PlaNet and Dreamer\footnote{\url{https://github.com/yusukeurakami/dreamer-pytorch}} that are distributed under MIT license and also base the implementations of Grad-MPC and of our method on the latter. The hyperparameters are listed in \cref{tab:hyperparams-dreamer}. 

We use the default hyperparameters for SAC+AE as described in \citet{yarats_2021}, except for the action repeat, which we set to two for a fair comparison.

\vspace{0.4in}
\begin{center}
    \textcolor{gray}{\textbf{\textemdash~Appendices continue on next page~\textemdash}}
\end{center}

\clearpage

\begin{scriptsize}
\begin{table}[H]
  \centering
  \caption{\textbf{Hyperparameters and their values used for the experiments.}}
  \begin{tabular}{llc}
    \toprule
    \textbf{Algorithm} & \textbf{Hyperparameter} & \textbf{Value} \\
    \midrule
    \multirow{18}{*}{All} & Optimizer & Adam \citep{kingma_adam_2017} \\
                          & Max. episode length & 1000 \\
                          & Action repeat & 2 \\
                          & Experience size & 1000000 \\
                          & Embedding size & 1024 \\
                          & Hidden size & 200 \\
                          & Belief size & 200 \\
                          & State size & 30 \\
                          & Exploration noise & 0.3 \\
                          & Seed episodes & 5 \\
                          & Collect interval & 100 \\
                          & Batch size & 50 \\
                          & Overshooting distance & 0 \\
                          & Overshooting KL beta & 0 \\
                          & Overshooting reward scale & 0 \\
                          & Global KL beta & 0 \\
                          & Free nats & 3 \\
                          & Bit depth & 5 \\
    \midrule
    Dreamer \& \method & Planning horizon & 15 \\
    \midrule
    \multirow{9}{*}{\shortstack[l]{Dreamer, \method\\ \& Policy+Grad-MPC}} & Activation function & ReLU / ELU \\
                                                           & Model learning rate & 6e-4 \\
                                                           & Actor learning rate & 8e-5 \\
                                                           & Critic learning rate & 8e-5 \\
                                                           & Adam epsilon & 1e-7 \\
                                                           & Grad clip norm & 100 \\
                                                           & Discount factor & 0.99 \\
                                                           & Horizon discount factor & 0.95 \\
    \midrule
    \multirow{5}{*}{\method}    & Action optimization rate & 0.1 \\
                                & Action noise & 0.2 \\
                                & Action reuse coefficient & 0.1 \\
                                & Candidates & 5 \\
                                & Optimization iterations & 1 \\
    \midrule
    \multirow{8}{*}{Grad-MPC \& PlaNet} & Activation function & ReLU \\
                            & Candidates & 1000 \\
                            & Elite candidates & 100 \\
                            & Optimization iterations & 10 \\
                            & Grad clip norm & 1000 \\
                            & Model learning rate & 1e-3 \\
                            & Adam epsilon & 1e-4 \\
                            & Planning horizon & 12 \\
    \midrule
    Grad-MPC & Action optimization rate & 0.05 \\
    \midrule
    \multirow{3}{*}{Policy+Grad-MPC} & Action optimization rate & 0.05 \\
                                    & Optimization iterations & 10 \\
                                    & Planning horizon  & 1 \\
    \bottomrule
  \end{tabular}
  \label{tab:hyperparams-dreamer}  
  \vspace*{-2ex}
\end{table}
\end{scriptsize}

\section{Summary of Differences to Policy+Grad-MPC}\label{sec:differences-grad-mpc}
We summarize the main differences between \method and the Policy+Grad-MPC method proposed in \citet{s_v_gradient-based_2023} as follows:
\begin{itemize}
  \item \textbf{Trajectory optimization.} While the general idea of using a policy to initialize gradient-based MPC is shared by both methods, there are important differences. \method uses not just a single trajectory but samples few trajectories from the policy and optimizes each trajectory independently. Additionally, rollout and optimization is performed using longer horizons than just a horizon of one, which is used by Policy+Grad-MPC. While these values can be parameterized, they have a significant impact on the behavior and performance of the optimization. For example, using a horizon of one time step leads to a myopic optimization, which is unsuitable for most problems as outlined in \cref{sec:dreamer}. Longer rollouts with learned world models are also more challenging due to imperfect models as shown in \cref{sec:gradient-analysis}. 
  \item \textbf{Uncertainty regularization.} We propose to incorporate uncertainty regularization into the MPC objective, which we find to be particularly important for high-dimensional problems.
  \item \textbf{Action reuse.} We further propose to reuse previously optimized actions instead of completely discarding them to reduce the number of optimization iterations and improve computational efficiency.
  \item \textbf{Extensive experiments and thorough ablations.} Grad-MPC \citep{s_v_gradient-based_2023} provides only limited experimental results and lacks in-depth implementation details. While it shows that gradient-based MPC with a policy network is promising for two sparse-reward tasks from the DeepMind Control Suite, it does not provide a full evaluation of the method in diverse settings such as different benchmarks, different world models or types of observations, nor does it address high-dimensional problems, efficiency of gradient-based MPC or analyzes why the performance of gradient-based MPC is usually worse, compared to gradient-free methods. In contrast, \method offers a comprehensive set of experiments that systematically analyze the performance of our method across a wide range of conditions, providing new insights into its applicability and efficiency to enable further research. 
  \item \textbf{Training with gradient-based MPC.} We also evaluate \method when enabling gradient-based MPC already during training and not just during inference. In contrast, Policy+Grad-MPC is only evaluated using pretrained Dreamer models. Our results show that our method is also competitive to gradient-free MPC methods such as MPPI in this setting. In contrast, our experiments with Policy+Grad-MPC showed that it prematurely converges due to the horizon of just one time step.
  \item \textbf{Different world models.} We integrate our method into different types of world models, i.e., Dreamer (generative) and TD-MPC2 (implicit, control-centric) to show that our method is not targeted to a specific world model architecture while Policy+Grad-MPC only evaluates their method using Dreamer.
  \item \textbf{Implementation.} Furthermore, we were not able to reproduce the results shown in \citet{s_v_gradient-based_2023} with the given information because it lacks in-depth implementation details and there is no official implementation available. In contrast, we provide implementation details and open-source our implementation so that future work can replicate and build upon. 
\end{itemize}

\vspace{0.4in}
\begin{center}
    \textcolor{gray}{\textbf{\textemdash~Appendices continue on next page~\textemdash}}
\end{center}

\clearpage
\section{Analysis of \method's Performance on Dog Run}\label{sec:analysis-dog-run}
As shown in \cref{fig:learning-curves-dream-mpc-tdmpc2}, \method (TD-MPC2) achieves a lower asymptotic performance when using gradient-based MPC during training compared to BMPC or TD-MPC2 for high-dimensional problems, such as Dog Run. To understand why \method performs worse than MPPI for Dog Run when being used during training, we empirically analyze the quality of the used TD-MPC2/BMPC world model for Dog Run and how it affects overall performance of MPC.

\subsection{Value Estimation}\label{sec:value-estimation}
\begin{figure}[htpb]
  \centering
  \includegraphics[width=\textwidth, keepaspectratio]{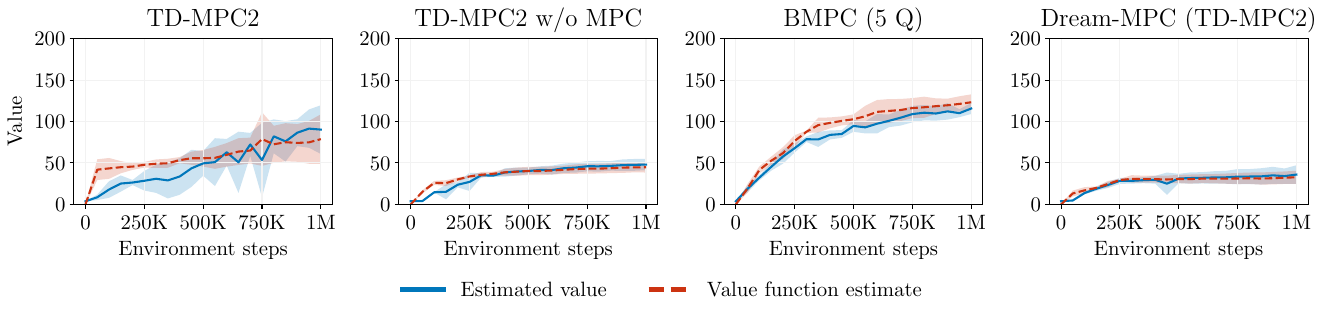}
  \caption{\textbf{Value estimations for Dog Run.} The lines represent the mean values and the shaded areas the 95\% confidence interval across 3 seeds.}
  \label{fig:value-estimation-dog-run}
\end{figure}

Recent work \citep{lin2025tdmpc2improvingtemporaldifference} has shown that TD-MPC2 suffers from value overestimation, particularly for high-dimensional tasks. The reasons are attributed to a policy mismatch: since the actions optimized by MPC are used to interact with the environment, the resulting training distribution corresponds to the MPC policy rather than that of the policy prior network. We follow the approach proposed by the authors and estimate values using the average discounted return for 100 evaluation episodes. The value approximation errors for BMPC, TD-MPC2 and \method (TD-MPC2) are shown in \cref{fig:value-estimation-dog-run}. For BMPC, we use five Q-networks in all following experiments instead of two V-networks to isolate potential confounding effects arising from using V-networks instead.

Similar to \citet{lin2025tdmpc2improvingtemporaldifference}, we find that for high-dimensional problems, such as Dog Run, TD-MPC2 tends to overestimate the values, although the gap is in our case smaller and also changes from an overestimation in the beginning to an underestimation at the end of the training. While this suggests that the underlying mechanisms might be more complex than a systematic overestimation and that further research on this is required, it emphasizes that in general calibrated value estimates are needed for policy learning and MPC. 

The bootstrapping mechanism of BMPC mitigates the distribution mismatch, which is indicated by lower value estimation errors and improved training stability. While the value estimation errors for TD-MPC2 without MPC are also small, the overall performance is worse and suggests that the value function has not seen state-action pairs, which would lead to a high reward. We discuss this in more detail in \cref{sec:policy-mismatch}. While gradient-based MPC constrains the policy mismatch by staying closer to the nominal policy distribution, it is prone to local optima, which can impair asymptotic performance, especially for high-dimensional problems.

\begin{figure}[htpb]
    \centering
    \begin{subfigure}[ht]{0.32\textwidth}
        \centering
        \includegraphics[width=\textwidth]{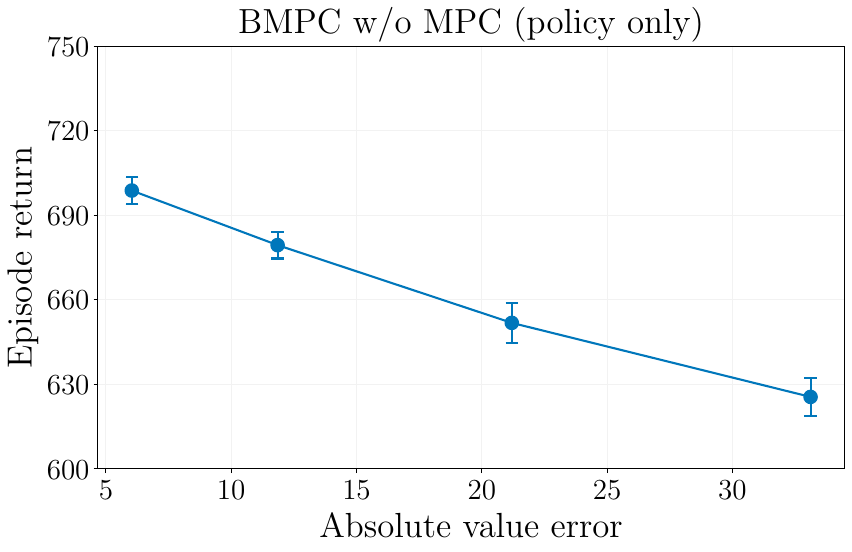}
        \vspace{-1.5em}
        \caption*{}
    \end{subfigure}
    \vspace{0.6em}
    \hfill
    \begin{subfigure}[ht]{0.32\textwidth}
        \raggedleft
        \includegraphics[width=\textwidth]{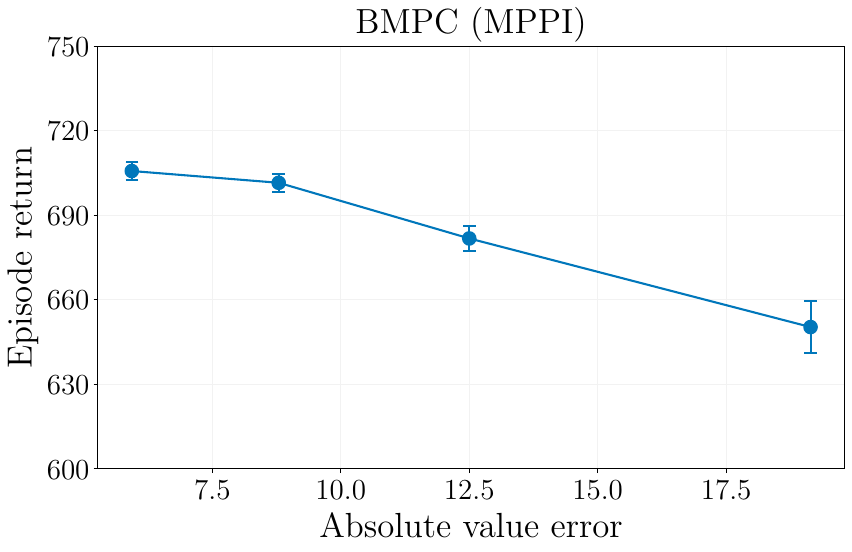}
        \vspace{-1.5em}
        \caption{}\label{subfig:value-estimates-a}
    \end{subfigure}
    \hfill
    \begin{subfigure}[ht]{0.32\textwidth}
        \centering
        \includegraphics[width=\textwidth]{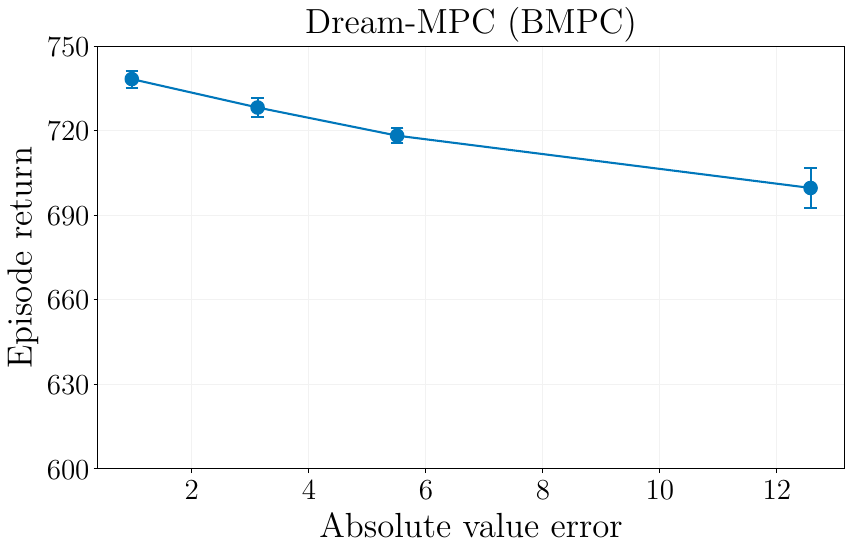}
        \vspace{-1.5em}
        \caption*{}
    \end{subfigure}
    \\
    \begin{subfigure}[ht]{0.32\textwidth}
        \centering
        \includegraphics[width=\textwidth]{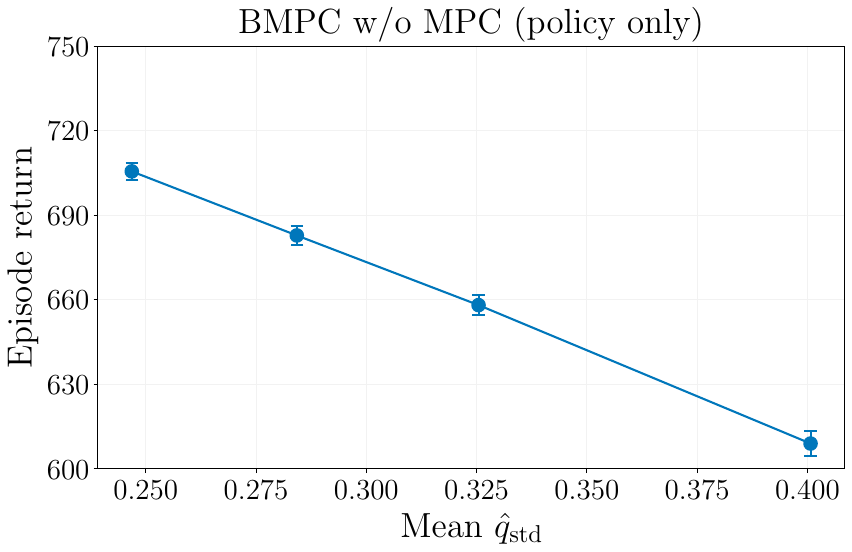}
        \vspace{-1.5em}
        \caption*{}
    \end{subfigure}
    \vspace{0.6em}
    \hfill
    \begin{subfigure}[ht]{0.32\textwidth}
        \raggedleft
        \includegraphics[width=\textwidth]{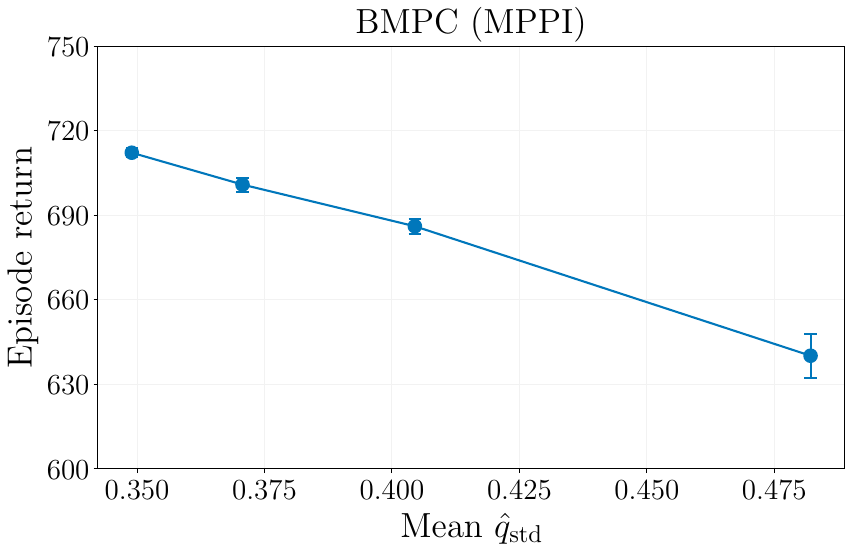}
        \vspace{-1.5em}
        \caption{}\label{subfig:value-estimates-b}
    \end{subfigure}
    \hfill
    \begin{subfigure}[ht]{0.32\textwidth}
        \centering
        \includegraphics[width=\textwidth]{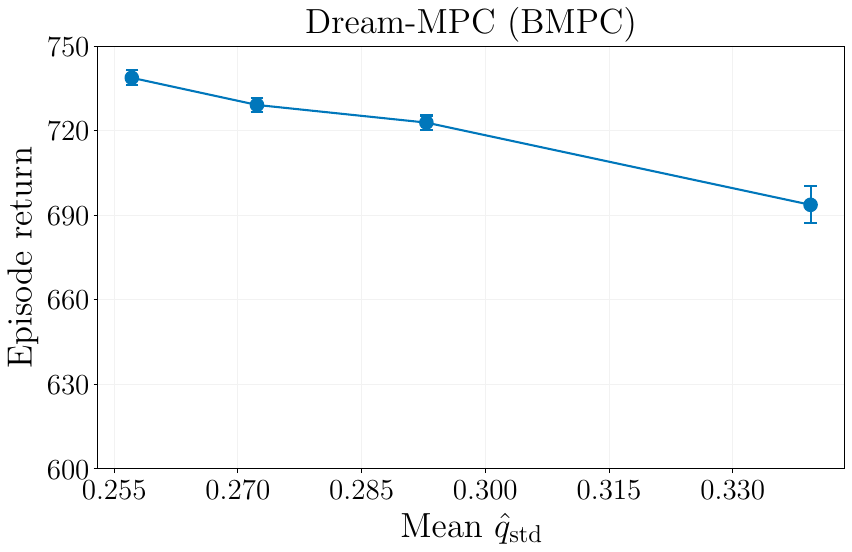}
        \vspace{-1.5em}
        \caption*{}
    \end{subfigure}
    \\
    \begin{subfigure}[ht]{0.32\textwidth}
        \centering
        \includegraphics[width=\textwidth]{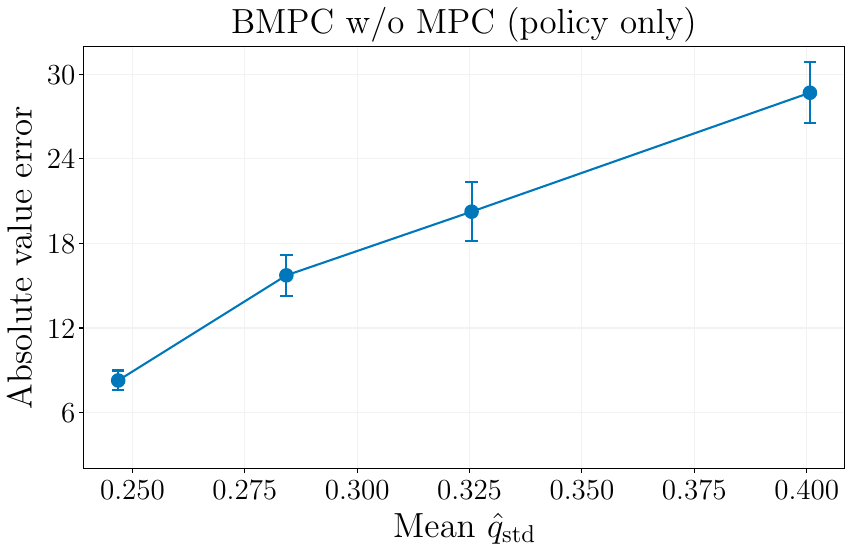}
        \vspace{-1.5em}
        \caption*{}
    \end{subfigure}
    \vspace{0.6em}
    \hfill
    \begin{subfigure}[ht]{0.32\textwidth}
        \raggedleft
        \includegraphics[width=\textwidth]{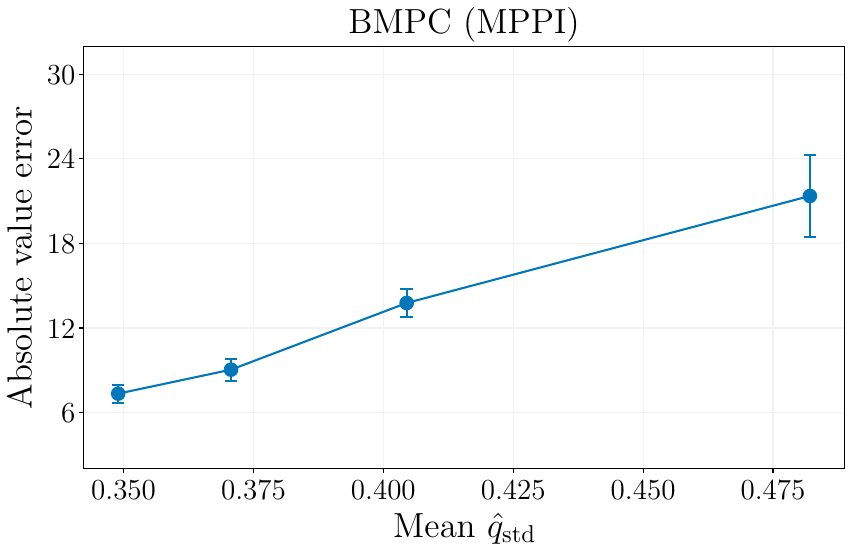}
        \vspace{-1.5em}
        \caption{}\label{subfig:value-estimates-c}
    \end{subfigure}
    \hfill
    \begin{subfigure}[ht]{0.32\textwidth}
        \centering
        \includegraphics[width=\textwidth]{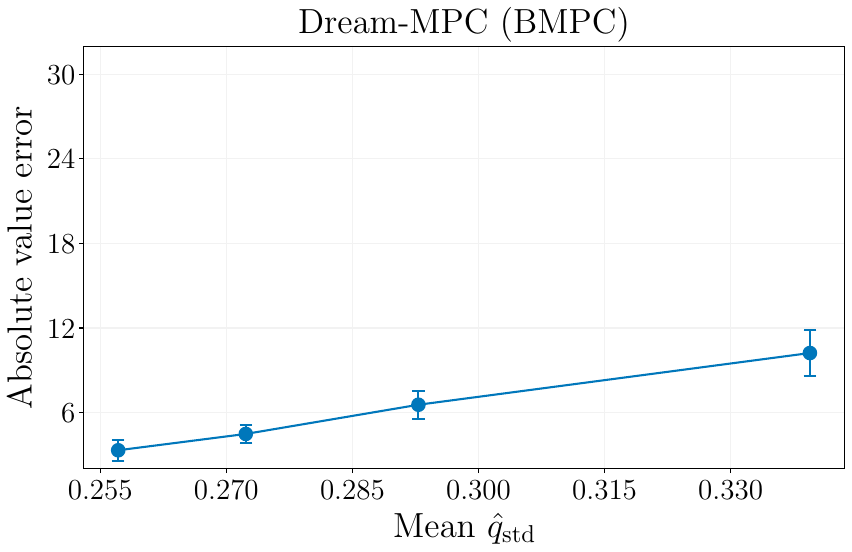}
        \vspace{-1.5em}
        \caption*{}
    \end{subfigure}
    \caption{\textbf{Effects of value approximation errors.} Analysis for BMPC on Dog Run. (a) Absolute value error vs. episode return. (b) Mean variance of predicted Q-values vs. episode return. (c) Mean variance of predicted Q-values vs. absolute value error.}
    \label{fig:value-estimation}
\end{figure}

Next, we want to evaluate how much value estimation errors affect the overall performance during inference. \cref{fig:value-estimation} shows the connections between the episode return $G$ and the estimated absolute value approximation error $e_q$ (\cref{subfig:value-estimates-a}), between $G$ and the episode mean variance of the predicted Q-values $\hat{q}_{\text{std}}$ (\cref{subfig:value-estimates-b}) as well as between $\hat{q}_{\text{std}}$ and $e_q$ (\cref{subfig:value-estimates-c}), respectively. Errors are binned into quartiles. We find that regardless of whether the policy, MPPI or gradient-based MPC is used, a lower episode return corresponds to a higher absolute value error. This is also reflected by the Spearman correlation coefficients reported in \cref{tab:spearman-correlations} and emphasizes that accurate value estimates are in general important for all considered methods. 

We further find that there is a connection between the episode return and the variance of the predicted Q-values $\hat{q}_{\text{std}}$ as shown in \cref{subfig:value-estimates-b} and by the Spearman coefficients. This indicates that episodes with a higher critic ensemble disagreement have much lower returns, suggesting that uncertainty is a good indicator for poor performance. 
Furthermore, there is a weaker correlation between $\hat{q}_{\text{std}}$ and $e_q$, which suggests that higher uncertainty also tends to be accompanied by larger value estimation errors. Although the correlation is not as strong as for the return, this indicates that uncertainty is meaningfully aligned with value function errors, and also confirms empirically the importance of taking uncertainty into account. 
\begin{table}[htpb]
\centering
\parbox{\textwidth}{
\caption{\textbf{Spearman correlation coefficients for Dog Run using BMPC with 5 Q-networks.} We omit p-values since they are all much smaller than 0.05.}
\label{tab:spearman-correlations}
\centering
\begin{tabular}{@{}lccc@{}}
\toprule
\textbf{Method}         & \textbf{Metric} & \textbf{Outcome} & \textbf{Spearman Coefficient} \\ \midrule
\multirow{3}{*}{Policy}                             & $\hat{q}_{\text{std}}$ & $G$ &  -0.93              \\
& $e_q$ & $G$ &   -0.74             \\
& $\hat{q}_{\text{std}}$ & $e_q$ & 0.68               \\ \midrule
\multirow{3}{*}{MPPI}                             & $\hat{q}_{\text{std}}$ & $G$ &  -0.88             \\
& $e_q$ & $G$ & -0.66               \\
& $\hat{q}_{\text{std}}$ & $e_q$ &      0.68          \\ \midrule
\multirow{3}{*}{\method} & $\hat{q}_{\text{std}}$ & $G$ & -0.80                \\
& $e_q$ & $G$ &       -0.59         \\         
& $\hat{q}_{\text{std}}$ & $e_q$ & 0.46 \\ \bottomrule
\end{tabular}%
}
\end{table}

While correlations are consistent when using the policy only, with MPPI and with gradient-based MPC, their strength varies. There is a trend suggesting that the correlations are slightly weaker with MPC, and gradient-based MPC showing even weaker correlations than MPPI. We hypothesize that the replanning of MPC allows for mitigating errors and that gradient-based MPC is less affected from model errors since the actions are closer to the nominal policy actions, which reduces the chance of out-of-distribution queries of the world model. This property makes gradient-based MPC potentially promising for offline RL, where the state-action coverage is limited.

\subsection{Policy Mismatch}\label{sec:policy-mismatch}
We further perform an experiment, where we analyze and compare the estimated values of BMPC with TD-MPC2. First, we use the much stronger policy of BMPC to evaluate whether the Q-networks of TD-MPC2, which was trained without MPC, can accurately estimate the values for good actions, although they have potentially not been seen or explored during training.

\begin{figure}[htpb]
  \centering
  \includegraphics[width=0.65\textwidth, keepaspectratio]{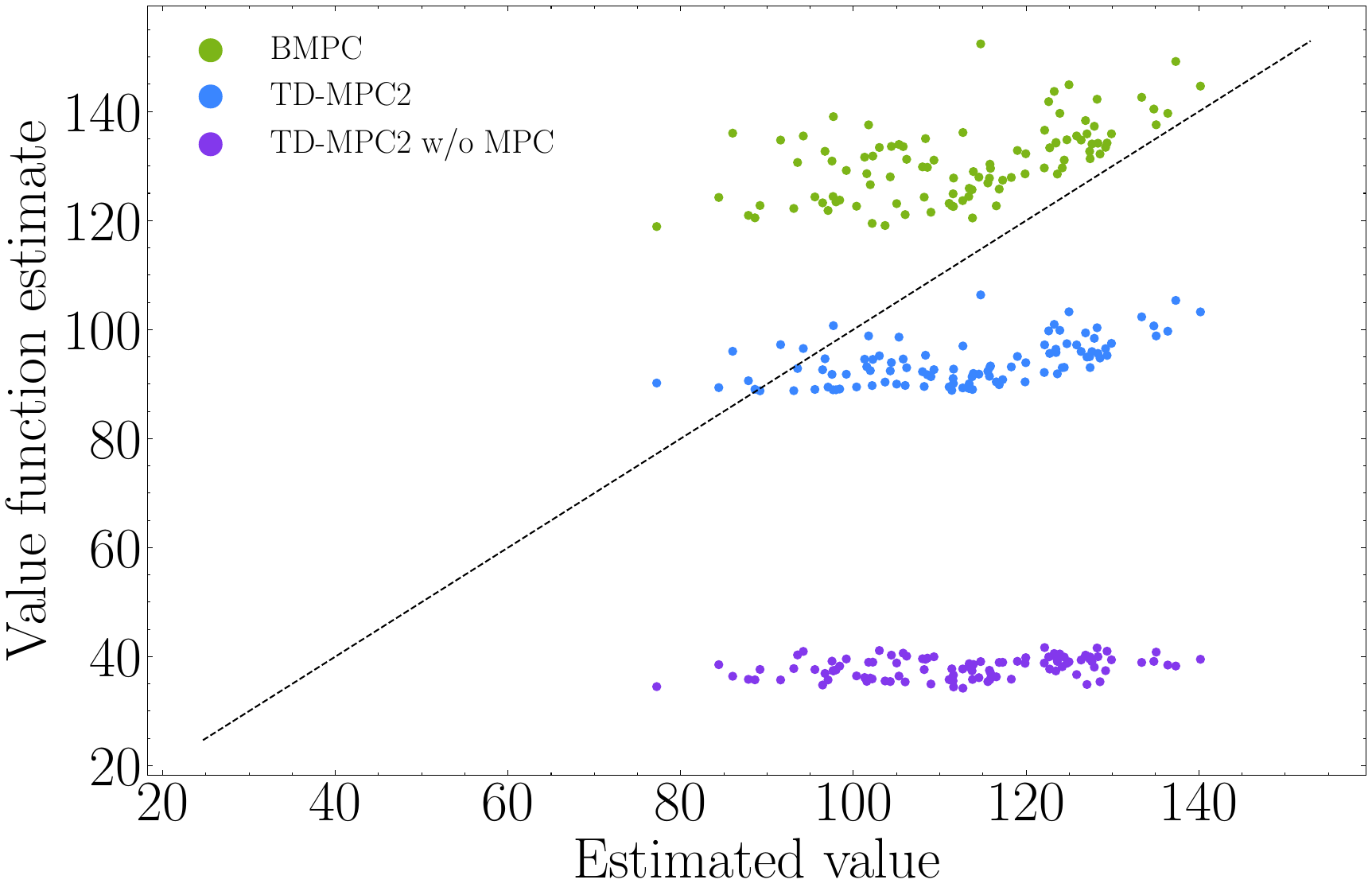}
  \caption{\textbf{Value estimation comparison for BMPC policy on Dog Run.}}
  \label{fig:value-estimation-comparison-dog-run}
\end{figure}

We find that BMPC slightly overestimates the values as shown in \cref{fig:value-estimation-comparison-dog-run}, but is quite accurate for trajectories corresponding to higher returns. TD-MPC2 trained without MPC in contrast significantly underestimates the values and basically predicts the same value for all episodes, indicating that the model has not seen state-action pairs leading to higher returns during training.

Next, we repeat this experiment, but use TD-MPC2, which was trained with MPPI to analyze whether the sampling-based MPC leads to improved exploration that helps to correct wrong value estimates. While TD-MPC2 still underestimates the values, the errors are much smaller, and it is specifically quite accurate for regions corresponding to lower returns. This suggests that sampling-based MPC can help to partially mitigate value approximation errors due to improved exploration during training and further highlights the importance of the policy and value functions being aligned to each other. Even if we have a well-performing policy, the world model needs to provide accurate predictions for the actions from the policy.

MPPI as a sampling-based MPC method benefits from a higher diversity across the candidate solution, which encourages exploration and thus improves training stability. The MPPI planner used in TD-MPC2 and BMPC uses a large number of random samples in addition to the policy prior samples, which help to explore underestimated or uncovered regions. Gradient-based MPC in contrast is prone to converging to local optima. Nevertheless, Dream-MPC performs surprisingly well and at least slightly improves compared to using only the policy prior, but is not able to match the performance of MPPI for high-dimensional problems when used during training. BMPC achieves the best performance, potentially because of a correction of the policy mismatch due to the bootstrapping mechanism in combination with carefully made design decisions to avoid premature convergence to local optima. BMPC adds an entropy loss to the policy learning objective and adds exploration in its lazy reanalyze mechanism to avoid premature convergence to local optima. We believe that this can potentially also help to further improve gradient-based MPC, but leave this for future work.

\end{document}